\documentclass[lettersize,journal]{IEEEtran}
\usepackage{amsmath,amsfonts}
\usepackage[ruled,vlined]{algorithm2e}
\usepackage{array}
\usepackage[caption=true,font=normalsize,labelfont=sf,textfont=sf]{subfig}
\usepackage{textcomp}
\usepackage{stfloats}
\usepackage{url}
\usepackage{verbatim}
\usepackage{graphicx}
\usepackage{cite}
\usepackage{hyperref}
\usepackage{multirow}
\usepackage{cleveref}
\usepackage{threeparttable}
\usepackage{xcolor}
\usepackage{amsmath}
\usepackage{amssymb}
\usepackage{tabularx}
\usepackage{caption}

\hypersetup{
    colorlinks=true,
    linkcolor=blue,
    filecolor=blue,
    urlcolor=black,
    citecolor=blue,
    pdfborder={0 0 0}
}

\Crefname{figure}{Fig.}{Figs.}
\Crefname{equation}{Eq.}{Eqs.}

\def\eg{e.g.}

\def\ie{{i.e.}}

\begin{document}

\title{pFedLVM: A Large Vision Model (LVM)-Driven and Latent Feature-Based Personalized Federated Learning Framework in Autonomous Driving}

\author{Wei-Bin Kou, Qingfeng Lin, Ming Tang, Sheng Xu, Rongguang Ye, Yang Leng, \\ ~~~~~~~Shuai Wang, Guofa Li, Zhenyu Chen, Guangxu Zhu*, Yik-Chung Wu*
\thanks{Wei-Bin Kou, Qingfeng Lin, Yang Leng and Yik-Chung Wu are with the Department of Electrical and Electronic Engineering, The University of Hong Kong, Hong Kong, China.}
\thanks{Guangxu Zhu is with Shenzhen Research Institute of Big Data, Shenzhen, China.}
\thanks{Ming Tang and Rongguang Ye are with the Department of Computer Science and Engineering, Southern University of Science and Technology, Shenzhen, China.}
\thanks{Sheng Xu is with Research Institute of Electronic Science and Technology, University of Electronic Science and technology, Chengdu, China.}
\thanks{Shuai Wang is with Shenzhen Institute of Advanced Technology, Chinese Academy of Sciences, Shenzhen, China.}
\thanks{Zhenyu Chen is with the State Key Laboratory for Novel Software Technology, Nanjing University, Nanjing, China.}
\thanks{Guofa Li is with the College of Mechanical and Vehicle Engineering, Chongqing University, Chongqing, China}
\thanks{\textit{(Corresponding author: Guangxu Zhu and Yik-Chung Wu.)}}
}

\markboth{IEEE TRANSACTIONS ON INTELLIGENT TRANSPORTATION SYSTEMS,~Vol.~xx, No.~x, April~2024}%
{Shell \MakeLowercase{\textit{et al.}}: A Sample Article Using IEEEtran.cls for IEEE Journals}

\maketitle

\begin{abstract}
Deep learning-based Autonomous Driving (AD) semantic segmentation (SSeg) models often exhibit poor generalization due to data heterogeneity in an ever domain-shifting environment. While Federated Learning (FL) could improve the generalization of an AD SSeg model (known as FedAD system), conventional models often struggle with under-fitting as the amount of accumulated training data progressively increases. To address this issue, instead of conventional small models, employing Large Vision Models (LVMs) in FedAD is a viable option for better learning of representations from a vast volume of data. However, implementing LVMs in FedAD introduces three challenges: \textbf{(I)} the extremely high communication overheads associated with transmitting LVMs between participating vehicles and a central server; \textbf{(II)} lack of computing resource to deploy LVMs on each vehicle; \textbf{(III)} the performance drop due to LVM focusing on shared features but overlooking local vehicle characteristics. To overcome these challenges, we propose pFedLVM, a LVM-Driven, Latent Feature-Based Personalized Federated Learning framework. In this approach, the LVM is deployed only on central server, which effectively alleviates the computational burden on individual vehicles. Furthermore, the exchange between central server and vehicles are the learned features rather than the LVM parameters, which significantly reduces communication overhead. In addition, we utilize both shared features from all participating vehicles and individual characteristics from each vehicle to establish a personalized learning mechanism. This enables each vehicle's model to learn features from others while preserving its personalized characteristics, thereby outperforming globally shared models trained in general FL. Extensive experiments demonstrate that pFedLVM outperforms the existing state-of-the-art approach by 18.47\%, 25.60\%, 51.03\% and 14.19\% in terms of mIoU, mF1, mPrecision and mRecall, respectively.
\end{abstract}


\section{Introduction}
\label{sec:intro} 
Autonomous Driving (AD) perception is essential \cite{natan2022towards} for AD vehicles. Typically, AD perception involves various tasks, such as semantic segmentation (SSeg) \cite{kou2024fast}, object detection \cite{song2024robustness}, and object tracking \cite{karle2023multi}. This research primarily concentrates on SSeg, which targets at categorizing each pixel into specific predefined categories. Contemporary SSeg models are predominantly developed using Deep Learning (DL) \cite{xiao2020multimodal,10494721,10372140}, particularly through Convolutional Neural Networks (CNNs) \cite{yu2021bisenet,10414408} and Transformers \cite{hoyer2023domain}. These models are usually trained on extensive datasets with pixel-level annotations. Once these models are trained, they are generally capable of predicting the category of each pixel in provided images. 

In general, AD SSeg is a highly complex task, and one major challenge in developing an AD SSeg model is its poor generalization due to significant data heterogeneity \cite{9917556}, which results from frequent domain shifting. For example, an AD vehicle transitioning into an unfamiliar environment may experience a notable decline in model performance compared to operations in usual and known settings. In practice, it is difficult to collect a  centralized and super-large scale dataset that covers almost all AD scenarios and conditions to train a SSeg model. Therefore, the AD SSeg model typically needs to update continuously in a driving-while-training manner.

\begin{figure}[!t]
\vspace{-0.4cm}
\centering
\includegraphics[width=\linewidth]{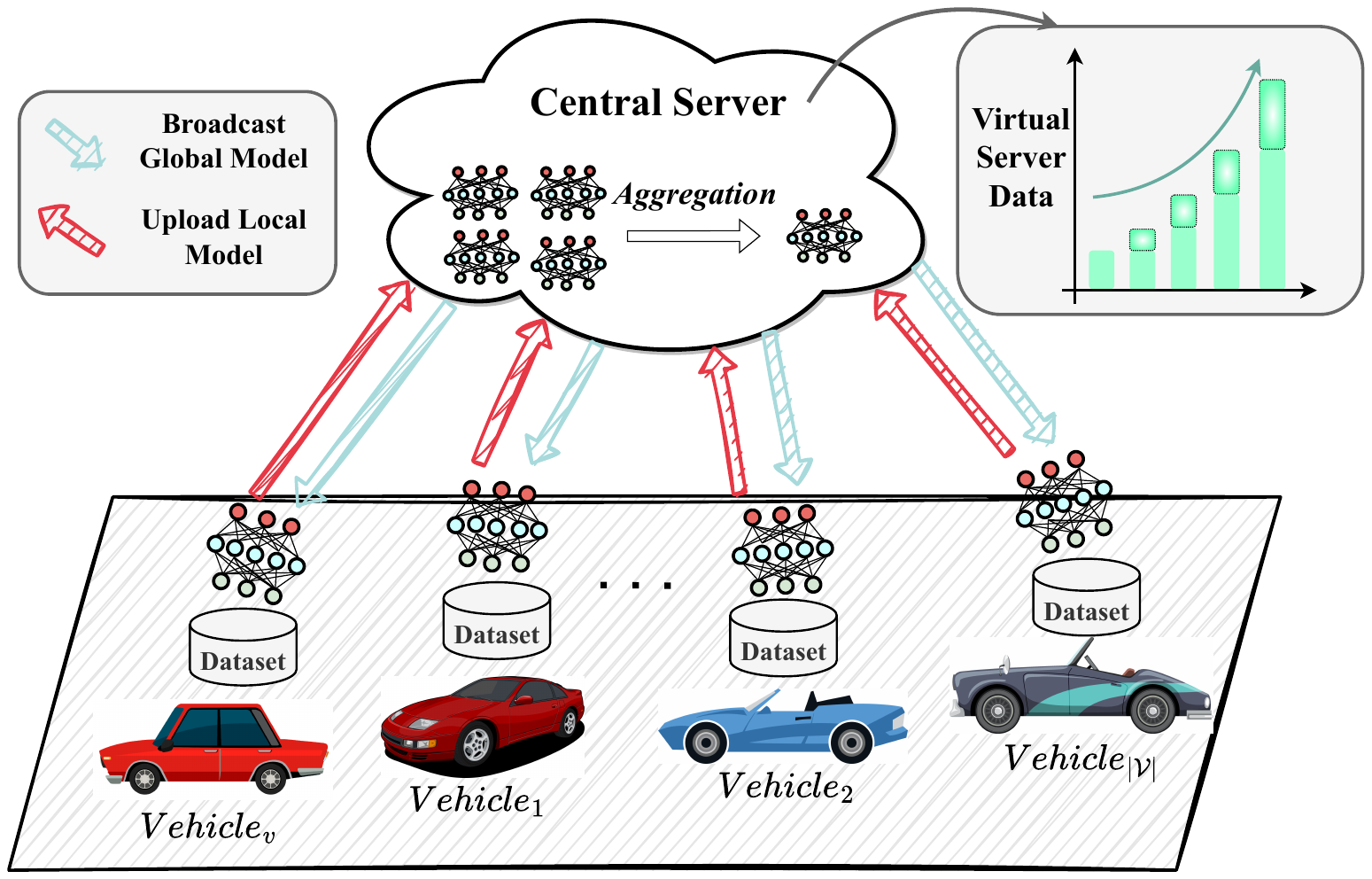}
\vspace{-1.5cm}
\caption{The Federated AD (FedAD) system. As time progresses, the virtual dataset that needs to be fitted by FedAD model expands continuously.}
\label{fig:short-a}
\vspace{-0.7cm}
\end{figure}

Federated Learning (FL) \cite{mcmahan2017communication,9272666} is an effective solution to make use of diverse data \cite{8970161,zhu2023pushing} from locationally distributed vehicles for improving the AD model generalization while preserving data privacy. Generally known as the FedAD \cite{kou2023communication,wu2024hierarchical,10346209,9831009}, the FL-based AD system typically composes of one \textbf{Central Server} and multiple \textbf{Vehicles}. The FedAD training procedure involves the following steps: \textbf{(I) Vehicle Updates:} each vehicle trains its local model using continuously collected data. \textbf{(II) Server Aggregation:} after every several updates at vehicles, the central server receives all participating vehicles' model and aggregates them as a weighted average, and then redistributes the aggregated model to all participating vehicles. \textbf{(III) Life-Long Learning:} steps \textbf{(I)} and \textbf{(II)} are iterated to learn from dynamically changing data. This FedAD system is pictorially illustrated in \Cref{fig:short-a}. 

As time progresses, the amount of data incorporated in the FedAD system continually expands, which helps to achieve substantial improvement in AD model generalization compared to deep learning model trained on data from each individual vehicle. However, such ever-expanding data is a double-edged sword. In general, the \textbf{Bias-Variance Tradeoff} \cite{nakkiran2021deep} in classical regime states that when the volume of data is less than what the model can comfortably accommodate, the model tends to overfit. Conversely, if the data volume exceeds the model's capacity, the model is susceptible to under-fitting. In the context of the FedAD system, as more and more data is used to train the FL model, the amount of information would eventually exceeds the model capacity, increasing the risk of under-fitting and leading to poor generalization performance.

\begin{figure}[t]
\centering
\includegraphics[width=\linewidth]{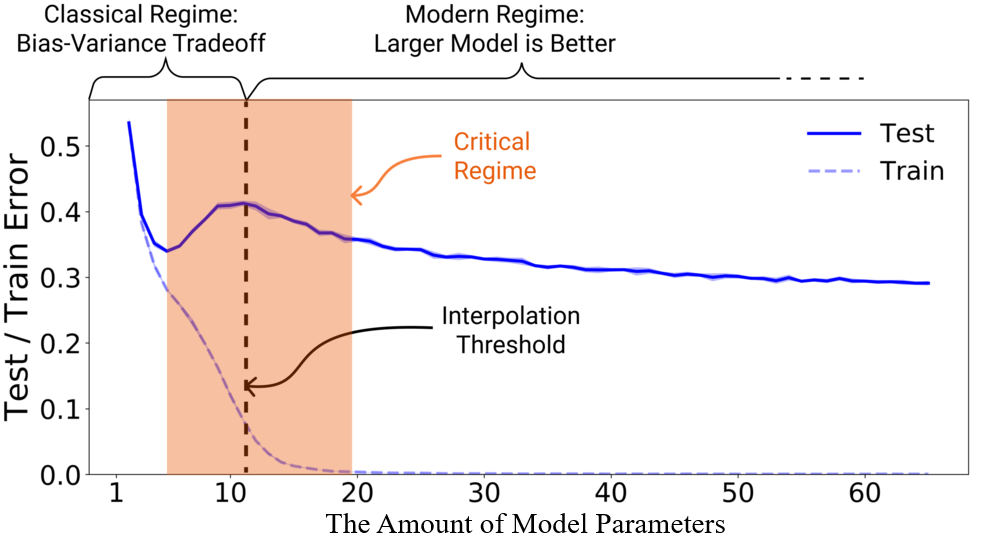}
\vspace{-0.5cm}
\caption{Illustration of the strength of larger model \cite{nakkiran2021deep}.}
\label{fig:largeisbetter}
\vspace{-0.5cm}
\end{figure}

To tackle such under-fitting problem, recent research enters the era of \textbf{Large Models (LMs)}. This is attributed to following reasons:
\begin{enumerate}
    \item  \textbf{Model Architecture:} On one hand, LMs have more layers (depth) and more neurons per layer (width). This allows them to form a hierarchy of features, from simple to complex, and to capture more intricate patterns in the data. On the other hand, LMs generally incorporate attention mechanisms \cite{vaswani2017attention}, which allow the model to focus on different parts of the input when generating each part of the output. This leads to a more context-aware representation of the data and results in more meaningful feature extraction. 
    \item  \textbf{Modern Regime \cite{nakkiran2021deep}:} As illustrated in \Cref{fig:largeisbetter}, larger model is better. Specifically, smaller models generally locate at Classical Regime, where these models face the problem of bias-variance trade-off. In contrast, LMs have a large number of parameters and they generally enter Modern Regime. This enables LMs to effectively capture and model the underlying distribution of the data, especially for complex AD SSeg task. 
    \item  \textbf{Scaling Law:} \cite{kaplan2020scaling} found that the model performance depends most strongly on the scale of the number of model parameters, the size of the training dataset, and the amount of compute used for training. The pretrained LMs contain billions of parameters, and are generally trained on large-scale datasets by using a massive amount of computing resources. Therefore, the pretrained LMs generally hold quite good performance to process complex and highly heterogeneous driving environments. 
    \item  \textbf{Pioneering Success:} Large Language Models (LLMs) have marked a significant milestone in the advancement of natural language processing (NLP), exhibiting proficiency across a variety of applications, such as language comprehension and generation \cite{Zhu2023mini}, interpretation of user intentions \cite{ouyang2022training}, solving question answering tasks based on structured data \cite{jiang2023structgpt}, and complex reasoning \cite{wei2022chain}. Similarly, Large Vision Models (LVMs) demonstrate the capability to interpret visual input and extract comprehensive semantic insights from image data \cite{wang2023videomae}.
\end{enumerate}

However, deploying LMs within FedAD system presents its own set of challenges. Firstly, due to the extremely large number of parameters in LMs, the communication between vehicles and the central server can lead to unbearable overheads. Secondly, the computing resource at vehicles would not be sufficient to train a large model locally. Thirdly, the inherent focus of LMs on extracting general features often leads to the inadvertent neglect of unique, vehicle-specific local characteristics. This issue primarily stems from the intrinsic weakness of \textit{generic overgeneralization} in LMs \cite{ralethe2022generic,collacciani2023interpretation}. Such overgeneralization issue arises due to the LMs' training on widely varying data aimed at capturing universally applicable patterns. Consequently, the nuanced details that are critical for distinguishing specific local attributes may be overlooked.

\begin{figure*}[!t]
\hspace{-0.5cm}
\includegraphics[width=1.05\linewidth,height=0.56\linewidth]{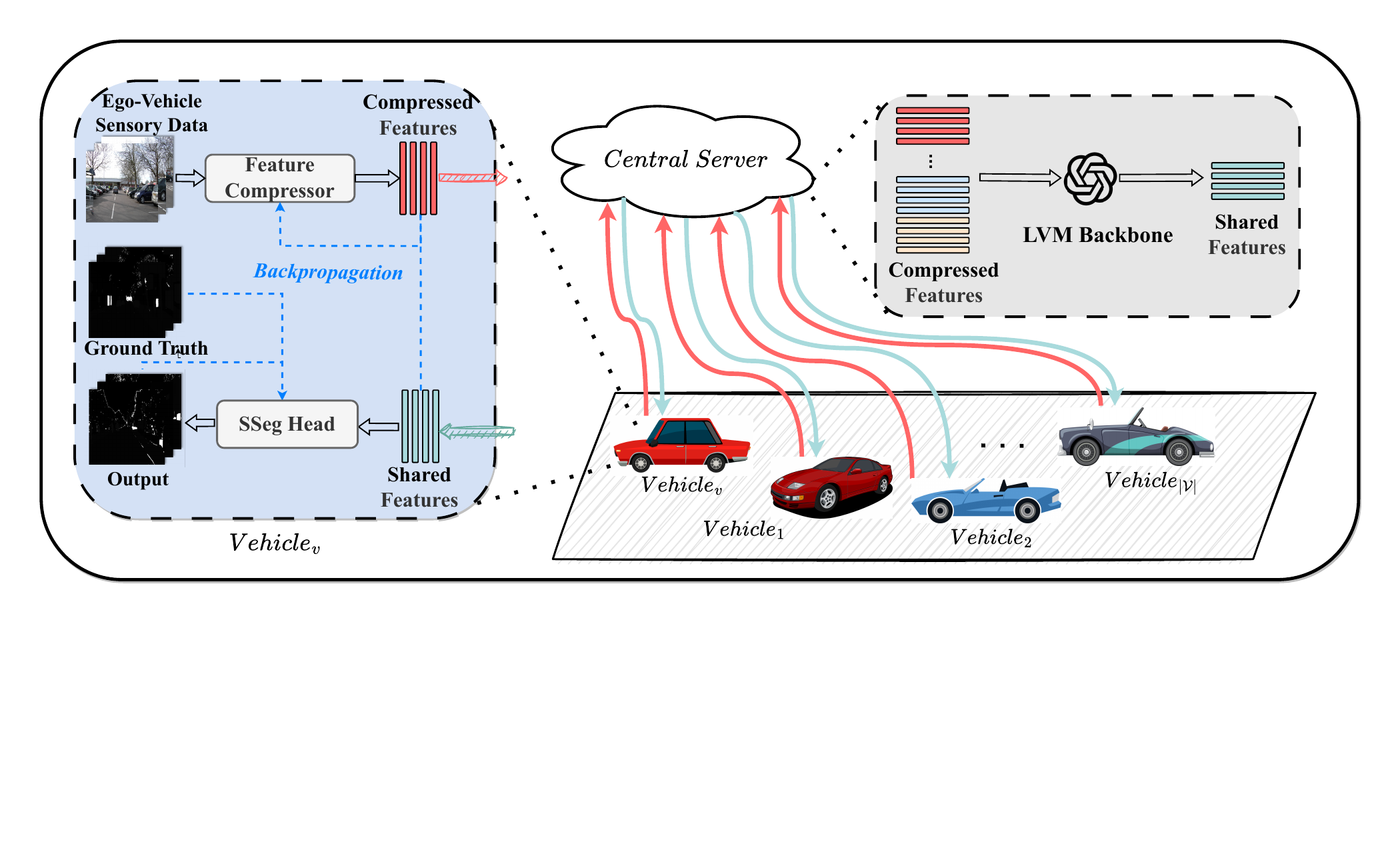}
\vspace{-3.6cm}
\caption{The illustration of the proposed pFedLVM framework. The proposed pFedLVM is composed of one central server and $|\mathcal{V}|$ vehicles. 
For each vehicle (\eg, $Vehicle_v$), the feature compressor extracts the compressed features which are transmitted to central server. The central server then uses LVM as a backbone to extract the shared features of all participating vehicles, and returns the extracted shared features to all involved vehicles. Once each vehicle received the shared features from the central server, the downstream SSeg head, taking  such shared features as input, is optimized by using the loss between the output of the SSeg head and ground truth via back propagation, while the feature compressor is updated by using the distance between the shared features and the compressed features via back propagation.}
\label{Fig.pFedLVM}
\vspace{-0.4cm}
\end{figure*}

To overcome these challenges in the context of FedAD system, we propose the \textbf{pFedLVM} framework which involves a two-fold strategy. On the one hand, since a major part of the AD system is vision-based, we propose deploying Large Vision Models (LVMs) on the central server but not at vehicles. To alleviate the computational burden on individual vehicles and communication overhead, each vehicle would train on a small model. Furthermore, the FL exchange between vehicles and the central server would be on features rather than the LVM parameters. On the other hand, to ensure that each vehicle's unique characteristics are not neglected, the proposed framework incorporates a feature-based personalized FL (pFL) mechanism. This mechanism leverages both shared features of all involved vehicles and each vehicle's individual characteristics, allowing for models to be tailored to each vehicle. Such pFL mechanism is particularly useful in FedAD because vehicles' behavior or driving pattern varies significantly for different vehicles. The proposed pFedLVM framework is summarized in \Cref{Fig.pFedLVM}.

With the proposed pFedLVM aiming to address the challenges posed by using LVMs in FedAD and pave the way for more effective and better autonomous driving systems, the main contributions of this paper are summarized as follow:
\begin{itemize}
    \item This work leverages LVMs in FedAD system. LVMs can overcome under-fitting problem in the ever-increasing training dataset in FedAD.
    \item To effectively alleviate the computational burden of each vehicle in FedAD system, we propose using LVMs as backbone on the central server only but not at the vehicle level. In addition, we exchange only the extracted features (instead of the parameters of LVMs), reducing communication overhead while sharing diverse knowledge learned at each vehicle.
    \item To avoid LVMs' overlooking local vehicle characteristics, we propose utilizing both shared features from all participating vehicles and individual characteristics from each vehicle to establish a personalized learning mechanism. This enables each vehicle's model to learn features from others while preserving its personalized characteristics, improving the inference performance in non-independent and identically distributed (non-i.i.d.) AD scenarios.
    \item Extensive experiments show that our proposed methods outperforms existing state-of-the-art (SOTA) benchmarks by 18.47\%, 25.60\%, 51.03\% and 14.19\% in terms of mIoU, mF1, mPrecision and mRecall, respectively. Additional empirical analyses are also conducted to explain the superiority of the proposed method.
\end{itemize}

The remainder of this paper is organized as follows. \Cref{related_work} provides an overview of the related work. \Cref{methodology} elaborates on the proposed pFedLVM framework, while Section \Cref{overhead_complexity} analyzes the communication overhead and complexity of  pFedLVM. Subsequently, \Cref{experiments} presents a comprehensive set of experiments along with an analysis of the empirical results. The paper concludes in \Cref{conclusion}.

\section{Related Works} \label{related_work}
\subsection{Semantic Segmentation (SSeg)}
SSeg is a field within computer vision and robotics focusing on enabling machines to interpret and understand the semantic information of vehicles' surroundings, typically through various forms of sensory data such as images and lidars. This capability is crucial for AD \cite{10388394,10161421} to understand the layout of the street scene, including the road, pedestrian, sidewalks, buildings, and other static and dynamic elements. Modern semantic understanding heavily relies on DL, with Fully Convolutional Networks (FCNs)-based models as early representatives \cite{yang2022deaot,zhou2022rethinking}. In recent years, Transformer-based approaches \cite{xie2021segformer} have also been proposed for semantic segmentation. Recently, Bird's Eye View (BEV) \cite{9697426} technique is widely adopted for road scene understanding. Moreover, federated learning \cite{10342134} has been adopted to improve the AD model generalization. In this paper, we propose to use LVMs backbone coupled with downstream SSeg head to understand semantic information of vehicles' surroundings.

\subsection{Federated Autonomous Driving (FedAD) System}
Current AD systems are commonly grouped into two categories in various studies: modular-based \cite{sun2016high, jiang2019multi} and learning-based \cite{xiao2020multimodal,nguyen2022deep,10494721,10372140}. Modular-based methods, although structured, are plagued by error propagation due to potential inaccuracies in both the modeling and problem-solving phases. In contrast, typical learning-based end-to-end approaches \cite{9165167} offer a promising alternative that mitigates error propagation. These methods directly transform sensory inputs, such as LiDAR point clouds and camera imagery, into vehicular control actions, encompassing throttle, brake, and steering commands. On the other hand, learning-based models can also be used within a modular pipeline, for example, using a learning model for SSeg task within the SSeg module \cite{10414408}. Nevertheless, the intrinsic challenge for learning-based paradigms lies in their generalization capabilities, often resulting in performance limitations to specific scenarios.

FL presents itself as a novel approach aimed at enhancing the generalization ability of learning-based systems \cite{10324362,9244132}. It achieves this through the aggregation of model parameters. Within the realm of AD, FL capitalizes on vehicular networks to combine insights from diverse vehicles operating across varied environments. Consequently, when an AD system encounters a new data sample or edge case, it can disseminate newfound knowledge to the centralized server and subsequently other vehicles, all while safeguarding data privacy \cite{10033088}. A notable instance is the cloud federated robotic system proposed in \cite{liu2020federated}, which augments the behavior cloning technique to yield precise control commands by leveraging RGB imagery, depth perception, and semantic segmentation. Recently, communication-efficient researches  \cite{10101681,10279509,9505307,10529194} develop rapidly so that FedAD could be more feasible.

\subsection{Personalized Federated Learning (pFL)}
To surmount the challenge of discrepancies in local data distribution in FL, pFL has been proposed as a solution \cite{pillutla2022federated}. This approach customizes the model in each client to account for the unique characteristics of local data. One of the most popular and effective methods for achieving pFL is the architecture-based approach \cite{tan2022towards}. This method decouples the model's parameters, allowing only a subset of parameters to be shared and aggregated among clients, while the remaining private parameters are selected based on model architecture~\cite{zhang2021parameterized} or data similarities~\cite{huang2021personalized, bui2019federated} to learn solely on local data. Another option enables each client to fine-tune global model locally \cite{NEURIPS2020_24389bfe,collins2021exploiting}. For example, ~\cite{NEURIPS2020_24389bfe} considers the global model as an initial shared model. By performing a few additional training steps locally, all the clients can easily fine-tune the initial shared model. \cite{collins2021exploiting} implements the above strategy by splitting the backbone into a global model (representation) and a client-specific head and fine-tunes the head locally to achieve personalization. In contrast, unlike the methods mentioned earlier, this paper introduces a novel strategy for pFL that is centered on feature maps, allowing each vehicle to learn concurrently from others while also maintaining its unique characteristics.

\subsection{Large Vision Models (LVMs)}
Recently, LLMs have achieved great success in the NLP field in various scenarios, such as user intent understanding \cite{ouyang2022training}, knowledge utilization \cite{jiang2023structgpt} and complex reasoning \cite{fu2022complexity} in a zero-shot/few-shot setting. Inspired by the achievements of pre-trained LLMs in NLP field, researchers have turned their attention to exploring pre-trained LVMs. These models, pre-trained on extensive image datasets, hold the ability to decipher image content and distill rich semantic information \cite{ren2023rejuvenating}. By learning representations and features from a significant volume of data, these models enhance the ability of computers to comprehend and analyze images, facilitating a range of diverse downstream applications \cite{wang2023all}. In this paper, by leveraging on their exceptional capabilities of semantic understanding, we propose the use of LVMs to extract and integrate the shared representations and features from all participating vehicles.

\section{Methodology}
\label{methodology}

\subsection{pFedLVM Overview}
The key notations in pFedLVM formulation are summarized in \Cref{tab:HFRS}. We consider a FedAD system, which includes a cloud server and $\mathcal{|V|}$ vehicles. $Vehicle_v$ denotes the $v$-th vehicle connected to the cloud server, where $v = 1, 2, \cdots, |\mathcal{V}|$. $Vehicle_v$ has a local dataset $\mathcal{D}_{v}$ with size $|\mathcal{D}_{v}|$. The Central Server virtually covers dataset $\mathcal{D} \triangleq \cup_{v=1}^{|\mathcal{V}|} \mathcal{D}_{v}$ with size $|\mathcal{D}|=\sum_{v=1}^{|\mathcal{V}|} |\mathcal{D}_v|$.
The proposed pFedLVM consists of two key elements.

Firstly, we propose deploying LVMs exclusively on the central server, and the central server and vehicles share collective knowledge via exchanging features. This strategy aims not only to significantly reduce the computational load at the vehicle level, but also allows knowledge sharing while alleviating communication overheads. Specifically, on the vehicle side, we propose a feature compressor (with parameters $\omega_{v, c}$) on each vehicle to extract compressed features (denoted as $\mathcal{F}_{v, c}$). Then such features are transmitted to the central server. Once the central server receives the compressed features $\mathcal{F}_{v, c}$ from $Vehicle_v$, where $v = 1, 2, \cdots, |\mathcal{V}|$, it concatenates such features to form $\mathcal{F}_{cat}$ which is then fed into the LVM backbone (with parameters $\omega_{lvm}$) to produce the shared feature maps $\mathcal{F}_{shd}$ of all participating vehicles. After that, the central server redistributes $\mathcal{F}_{shd}$ to all participating vehicles.

Secondly, to acknowledge and incorporate the unique characteristics of each vehicle, we introduce a personalized learning mechanism. This mechanism utilizes both the shared features (denoted as $\mathcal{F}_{shd}$) of all participating vehicles and the local characteristics (denoted as $\mathcal{F}_{v, c}$) of each vehicle. As aforementioned, we utilizes a feature compressor (with parameters $\omega_{v, c}$) to extract compressed features $\mathcal{F}_{v, c}$ in order to reduce communication overheads. The feature compressor parameters $\omega_{v,c}$ is optimized by the loss depending on the discrepancy between $\mathcal{F}_{v, c}$ and $\mathcal{F}_{shd}$ via back propagation. In addition, we also design downstream personalized control module to enhance the inference performance. The control module in general consists of two blocks: modular heads (including sequentially connected perception head, planning head and control head) and end-to-end head. Such heads are updated by the loss taking into account these heads' output and corresponding heads' ground truth. By introducing such personalized mechanism, it allows for the models (including feature compressor and various downstream heads) to be customized for each vehicle. This mechanism, with its superior ability to capture unique patterns and preferences, enables each vehicle to perform better than a global model trained through a general FL. This is particularly advantageous in FedAD, where the behavior and patterns among vehicles can vary substantially. 

To facilitate the understanding of how vehicles' personalized models (including the personalized Feature Compressor and the personalized SSeg head) and server's LVM backbone work, we take one vehicle as an example. Specifically, this vehicle's Feature Compressor, server's LVM backone, and this vehicle's SSeg Head, process the inputs sequentially. The data flow starts at the vehicle's Feature Compressor and then proceeds through the server's LVM. It then concludes at the vehicle's SSeg head, where it outputs the predicted semantic masks. This process ensures a streamlined SSeg task.

\begin{table}[tp]
    \centering
    \footnotesize 
    \renewcommand{\arraystretch}{1.0}
    \caption{Key Notations in pFedLVM}
    \begin{tabularx}{\linewidth}{ll}
    \hline
        \textbf{Symbols} & \textbf{Definitions} \\ \hline
        $\mathop{v}$ & Vehicle ID \\
        $\mathcal{V}$ & Participating vehicle set \\
        $\mathcal{D}_{v}$ & Training dataset on Vehicle $\mathop{v}$ \\
        $\mathcal{D}_{v}^{(j)}$ & The $j$-th batch of data out of $\mathcal{D}_{v}$ \\
        $\mathcal{\omega}_{v, c}$ & Model parameters of Feature Compressor on Vehicle $\mathop{v}$ \\
        $\mathcal{\omega}_{v, p}$ & Model parameters of SSeg Head on Vehicle $\mathop{v}$ \\
        $\mathcal{\omega}_{lvm}$ & Model parameters of the LVM backbone  \\
        $\mathcal{F}_{v, c}$ & The compressed feature maps of $Vehicle_v$ \\
        $\mathcal{F}_{cat}$ & The concatenated feature maps of all vehicles \\
       \multirow{2}{*}{$\mathcal{F}_{shd}$} & The cross-vehicle shared feature maps extracted\\ ~ &~~~~by server-side LVM backbone \\
       $\mathcal{G}_{v, p}$ & The ground truth of SSeg Head of $Vehicle_v$ \\
       $\mathcal{O}_{v, p}$ & The output of SSeg Head of $Vehicle_v$ \\
       \multirow{2}{*}{$\mathcal{*}^{(j)}$} & The counterpart of $\mathcal{D}_{v}^{(j)}$ for symbol ``$*$''. For example, \\~&~~~~$\mathcal{F}_{v, c}^{(j)}$ means the $\mathcal{F}_{v, c}$ due to $\mathcal{D}_v^{(j)}$  \\
       \hline
    \end{tabularx}
\label{tab:HFRS}
\vspace{-0.5cm}
\end{table}

\subsection{LVM-driven Shared Feature Extraction}
\subsubsection{LVM Backbone}
Pretrained ImageGPT \cite{chen2020generative}, often abbreviated as iGPT, is an outstanding representative of LVMs and selected as the backbone. By pretraining on a large-scale image dataset, iGPT learns rich representations of images that can be leveraged for a variety of image-processing tasks. In our proposed framework, the iGPT is utilized to fuse and extract shared features in a zero-shot fashion. These shared features are high-dimensional vectors that capture the model's understanding of the image content, and they can be used as input for a variety of downstream tasks. 

The strength of a LVM backbone over a traditional CNN-based backbone stems from its model complexity and scale. While CNN-based backbone is effective for many applications, their performance generally plateaus on complex and large-scale datasets. In contrast, the LVM backbone operates with larger-scale parameters and enhanced learning capabilities. This makes LVMs particularly suitable for complex SSeg task in dynamic AD environments.

\subsubsection{Shared Feature Extraction and Fusion}
Recall that all the participating vehicles extract their compressed features (termed as $\mathcal{F}_{v, c}^{(j)}$ where $v=1, 2, \cdots, |\mathcal{V}|$) and send them to the central server. The central server concatenates such features together to form $\mathcal{F}_{cat}^{(j)}$ which is then fed into the LVM backbone to produce the shared feature maps $\mathcal{F}_{shd}^{(j)}$. This process is given by 
\begin{align}
    \mathcal{F}_{cat}^{(j)} &= \bigoplus_{v=1}^{|\mathcal{V}|} \mathcal{F}_{v, c}^{(j)},
    \label{Eq:f_cat}
    \\
    \mathcal{F}_{shd}^{(j)} &= \omega_{lvm}(\mathcal{F}_{cat}^{(j)}),
    \label{Eq:w_lvm}
\end{align}
where symbol $\bigoplus$ is defined as concatenation operation of all involved items.

From \Cref{Eq:f_cat,Eq:w_lvm}, it is obvious that $\mathcal{F}_{shd}^{(j)}$ contains the fused features of all participating vehicles. It is worth noting that due to the powerful capabilities of representation of LVMs, the generated fusion features $\mathcal{F}_{shd}^{(j)}$ are generally stable and effective instead of under-fitting for ever-increasing vast amount of data. Once the shared feature maps $\mathcal{F}_{shd}^{(j)}$ have been extracted, the central server sends them back to all participating vehicles. The shared features serves two purposes: 1) It enables each vehicle to benefit from the shared knowledge of all the other vehicles, improving all participating vehicles' inference performance; 2) The shared feature can be used for personalized learning, which is covered next. 

\subsection{Latent Feature-based Personalized FL (pFL)}
In this section, we present the details of the proposed innovative mechanism of pFL that hinges on feature maps. This strategy enables each vehicle to simultaneously learn from the collective knowledge of others while preserving its distinct characteristics. Specifically, in the context of AD, the driving environments of different vehicles vary significantly. For example, some vehicles may drive in countryside, where they are surrounded by trees, animals, etc., while other vehicles may drive in the urban areas, where the dominant scenes involve mostly pedestrians and buildings. The shared features of these two cases, such as road appearance, play crucial role in AD SSeg, whereas the personalized information, such as color, texture, contributes more for individual cases. 

\begin{figure}[t]
\vspace{-0.2cm}
\hspace{-0.3cm}
\includegraphics[width=1.1\linewidth,height=1.05\linewidth]{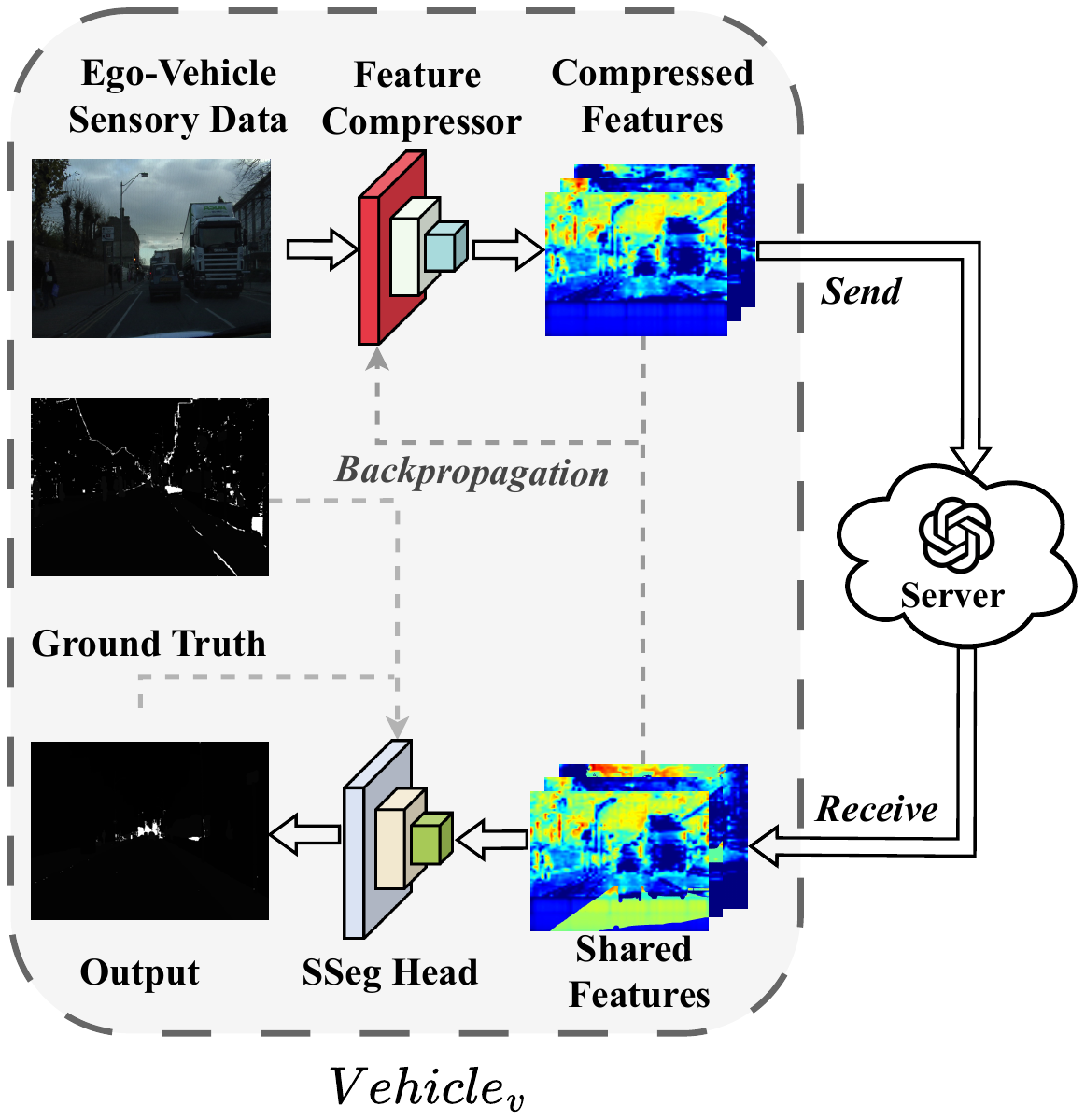}
\vspace{-2.8cm}
\caption{This figure illustrates the proposed latent feature-based pFL. We utilize the compressed features and the shared features to compute the loss for the feature compressor update. Meanwhile, the SSeg head, taking the shared feature as input, is trained based on the output of the SSeg head and the ground truth. As a result, the feature compressor and the SSeg head both learn shared features while maintaining local uniqueness.}
\label{Fig.pFedLVM_personalized}
\vspace{-0.5cm}
\end{figure}

To retain these personalized information, each vehicle in the proposed pFedLVM adopts two personalized models: \textbf{(I) Personalized Feature Compressor:} Each vehicle's Feature Compressor is a CNN-based model and is optimized by back propagation based on the loss between the local compressed features and the received shared features. \textbf{(II) Personalized SSeg Head:} Each vehicle's SSeg head takes the received shared features as input to predict the semantic masks, and then is optimized by back propagation based on the loss between the predicted semantic masks and the ground truth. These two personalized models guarantee to learn both personalized characteristics and the shared features for each vehicle. \Cref{Fig.pFedLVM_personalized} illustrates such pFL in details.

\subsubsection{Personalized Feature Compressor}
For the feature compressor, each vehicle undertakes the training based on the onboard dataset (termed as $\mathcal{D}_{v}$) as well as the shared feature maps (termed as $\mathcal{F}_{shd}$). The loss function for optimizing $\omega_{v,c}$ is defined as $\mathcal{E}_{c}(\mathbf{\omega}_{v, c},\mathcal{F}_{v, c}^{(j)}, \mathcal{F}_{shd}^{(j)})$, as shown in  \Cref{Eq:comp_forward,Eq:comp_loss}:
\begin{align}
\mathcal{F}_{v, c}^{(j)} &= \omega_{v, c}(\mathcal{D}_v^{(j)}),
\label{Eq:comp_forward}
\\
\mathop{\mathrm{min}}_{\mathbf{\omega}_{v, c}}\mathcal{L}_{c}(\mathcal{\omega}_{v, c}) &=
\frac{1}{|\mathcal{D}_{v}|} \sum_{\mathcal{D}_{v}^{(j)}\in\mathcal{D}_{v}}\mathcal{E}_{c}(\mathbf{\omega}_{v, c}, \mathcal{F}_{v, c}^{(j)}, \mathcal{F}_{shd}^{(j)}).
\label{Eq:comp_loss}
\vspace{-0.2cm}
\end{align}

As detailed in \Cref{Eq:comp_forward}, $\mathcal{D}_v^{(j)}$ is fed into the feature compressor $\omega_{v, c}$ to extract the compressed features (termed as $\mathcal{F}_{v, c}^{(j)}$). Once each vehicle has extracted the compressed features, such features are sent to the central server, which consumes much smaller communication overheads compared to transferring raw images and preserves privacy as well. Then the server performs a critical role in merging these features and distributing the shared features (termed as $\mathcal{F}_{shd}^{(j)}$) back to all vehicles.

Upon receipt of these shared features $\mathcal{F}_{shd}^{(j)}$, each vehicle calculates the loss in \Cref{Eq:comp_loss} which represents the discrepancy between the compressed features $\mathcal{F}_{v, c}^{(j)}$ and the shared features $\mathcal{F}_{shd}^{(j)}$. Subsequently, this loss is optimized to update  $\omega_{v, c}$ through back propagation. \Cref{Eq:comp_loss} is a general expression, and the specific loss function depends on the applications involved. In the implementation in \Cref{experiments}, we employ the mean-square error loss (see \Cref{Tab:train} for details). 

With respect to the Feature Compressor architecture, it can be designated according to practical needs and conditions. In our design, we adopt a CNN-based model. Notably, the compression level of the Feature Compressor should strike a balance between the communication overhead and vehicles' model performance (\eg, mIoU). In general, higher compression level (corresponding to smaller compressed feature size) leads to smaller communication overhead, but results in poor performance of vehicles' models. In contrast, lower compression level (corresponding to larger compressed feature size) enhances each vehicle's model performance, but it increases the communication overhead.

\subsubsection{Personalized Downstream Heads}
Besides for learning the individual feature compressor, the shared features $F_{shd}^{(j)}$ from the central server are also used to train the downstream heads at each vehicle. In general, the heads are commonly divided into two categories: modular heads (including perception head, planning head and control head) and end-to-end head. As all such heads are trained in a similar manner, in this paper, we just focus on the SSeg head to demonstrate the proposed feature-based personalized FL mechanism. The function of SSeg head is to classify each pixel of an image into one of the predefined categories that represent specific semantic classes. Training of other heads can be easily added as an extension. 

For the SSeg head (with parameters $\omega_{v, p}$), the loss is the distance between the output (termed as $\mathcal{O}_{v, p}^{(j)}$) and ground truth (termed as $\mathcal{G}_{v, p}^{(j)}$) of SSeg head which is given by \Cref{Eq:perc_forward,Eq:perc_loss}: 
\begin{align}
   \mathcal{O}_{v, p}^{(j)} &= \omega_{v, p}(\mathcal{F}_{shd}^{(j)}), 
   \label{Eq:perc_forward}
   \\
   \mathop{\mathrm{min}}_{\mathbf{\omega}_{v, p}}~\mathcal{L}_{p}(\mathcal{\omega}_{v, p}) &=
\frac{1}{|\mathcal{D}_{v}|}
\sum_{\mathcal{D}_{v}^{(j)}\in\mathcal{D}_{v}
} \mathcal{E}_{p}(\mathbf{\omega}_{v, p},\mathcal{G}_{v, p}^{(j)}, \mathcal{O}_{v, p}^{(j)}).
\label{Eq:perc_loss}
\end{align}

As outlined in \Cref{Eq:perc_forward}, once $Vehicle_v$ receives the shared features $\mathcal{F}_{shd}^{(j)}$ from the central server, these features are fed into the SSeg head with parameters $\omega_{v, p}$ to generate the output $\mathcal{O}_{v, p}^{(j)}$. Subsequently, as detailed in \Cref{Eq:perc_loss}, this output along with the ground truth $\mathcal{G}_{v, p}^{(j)}$ is employed to calculate the loss $\mathcal{E}_{p}$. In the implementation, we employ the cross entropy loss for $\mathcal{E}_{p}$, but the loss function can take other forms to suit the specific application. Based on this loss, the SSeg head model $\omega_{v, p}$ can be updated via back propagation. It is clear that the trained $\omega_{v, p}$ exhibits properties associated with personalized FL thanks to its integration of both shared features and unique characteristics specific to each vehicle due to proprietary SSeg task data. 

With respect to the SSeg head architecture, it can be designated accordingly based on practical conditions and needs. In our proposal, we adopt ASSP architecture \cite{chen2017deeplab}, which is designed to segment objects at multiple scales and is characterized by its parallel modules with dilated convolution at different dilation rates, enabling the model to capture multi-scale context by aggregating features from various receptive field sizes.

Trading off the generalization and the personalization is a critical consideration in the proposed pFedLVM. In general, a FL model with high generalization is capable of performing well across a broad range of driving scenarios, which is crucial for ensuring the robustness and scalability of SSeg model in AD. In contrast, incorporating personalization allows FL model to adapt to individual vehicles, potentially enhancing each vehicle's satisfaction and effectiveness in AD contexts. The proposed pFedLVM achieves a trade-off between the generalization and the personalization by leveraging the following elements: \textbf{(I) Server-side LVM backbone:} The LVM backbone is pretrained on a large-scale dataset, resulting in a powerful generalization capabilities for a large range of AD scenarios. \textbf{(II) Vehicle-side personalized Feature Compressor and SSeg Head:} Personalized Feature Compressor and SSeg Head are optimized according to the personalized inputs and the received shared features, guaranteeing to learn both personalized characteristics and shared features.

\subsection{Summary of the pFedLVM Algorithm}
The proposed pFedLVM framework in AD is summarized as \Cref{alg:pFLVM}. The algorithm is designed to facilitate the training of personalized models, which includes a personalized Feature Compressor and a personalized downstream SSeg head, based on mini-batches of data in an iterative fashion. The training process and a toy example of pFedLVM are summarized as follows:

\subsubsection{Training of pFedLVM}
The training of pFedLVM builds on a mini-batch basis. For each mini-batch, vehicles compress the inputs in parallel by respective Feature Compressor and output the compressed features. Then, send the compressed features are sent to the server simultaneously. After that, the server extracts the shared features of all vehicles' compressed features by using the LVM backbone, and then sends the shared features back to all vehicles. Subsequently, all vehicles operate in parallel. For each vehicle, on the one hand, the Feature Compressor is optimized via back propagation based on the loss between the local compressed features and received shared features; on the other hand, the SSeg head takes the received shared features as inputs to predict the semantic masks, then is optimized via back propagation based on the loss between the predicted semantic masks and the ground truth.

\begin{figure}[t]
\vspace{-0.2cm}
\hspace{-0.3cm}
\includegraphics[width=1.05\linewidth,height=0.55\linewidth]{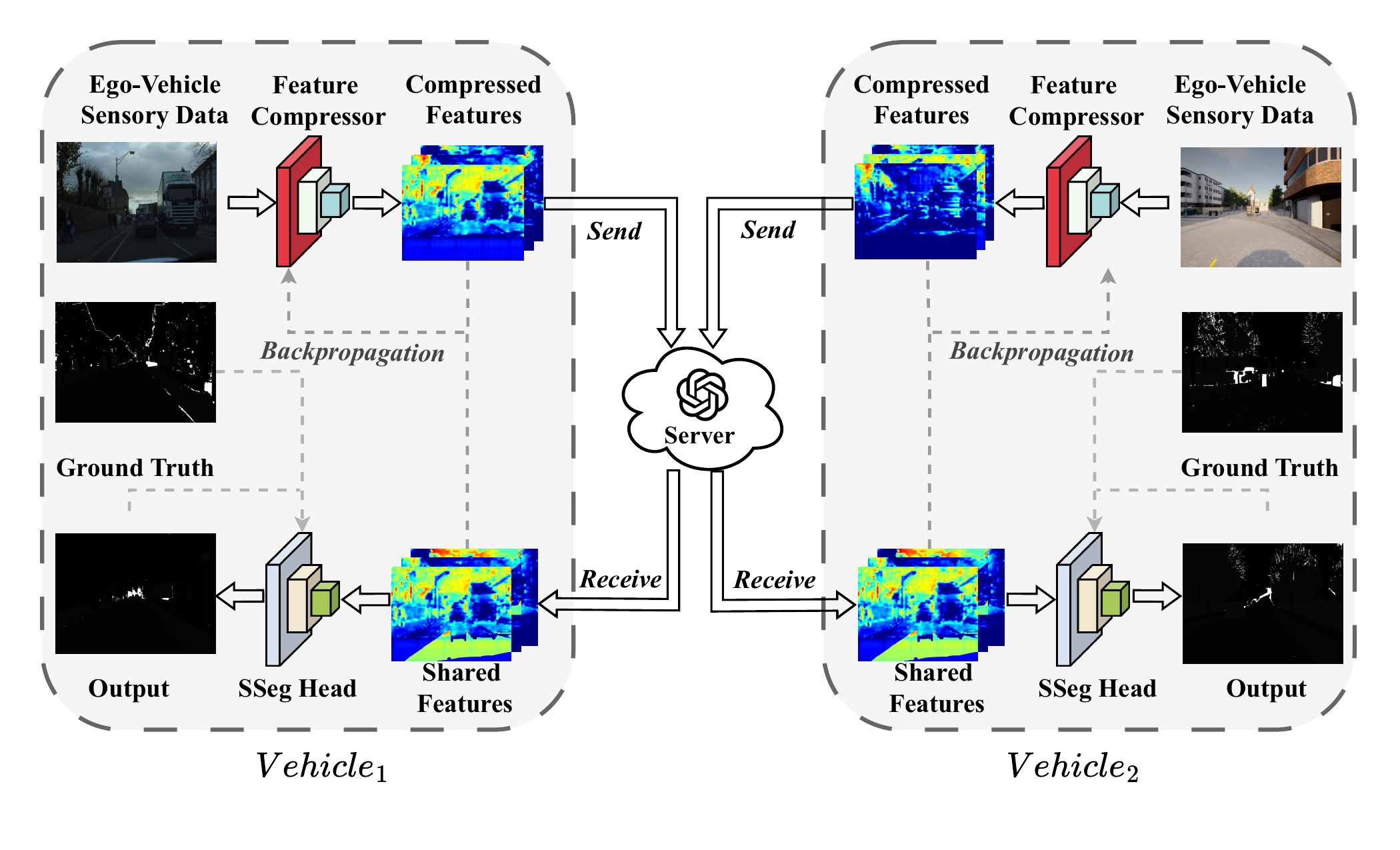}
\vspace{-0.8cm}
\caption{Illustration of the toy example of pFedLVM.}
\label{Fig.pFedLVM_toy_example}
\vspace{0.2cm}
\end{figure}

\subsubsection{Toy Example of pFedLVM}
To facilitate the understanding of the working procedure of pFedLVM, a toy example is illustrated in \Cref{Fig.pFedLVM_toy_example}. Specifically, this toy example includes $Vehicle_1$ and $Vehicle_2$, and the datasets that locate on both vehicles are heterogeneous. As discussed earlier, pFedLVM operates in a mini-batch basis and it should repeat multiple mini-batches until convergence. For each mini-batch, as $Vehicle_1$ and $Vehicle_2$ drive at their respective environments, they compress the latent features of inputs by their respective Feature Compressors in parallel, and send the compressed features to the server simultaneously. Then, the server uses the LVM backbone to extract the shared features of the compressed features of $Vehicle_1$ and $Vehicle_2$, and then broadcasts the shared features back to $Vehicle_1$ and $Vehicle_2$. Subsequently, $Vehicle_1$ and $Vehicle_2$ optimize their Feature Compressors simultaneously via back propagation based on the loss between their compressed features and the received shared features. Furthermore, $Vehicle_1$ and $Vehicle_2$ predict the semantic masks using SSeg heads by taking the received shared features as input, and then optimize their respective SSeg head via back propagation based on the loss between the predicted semantic masks and the ground truth.

\setlength{\textfloatsep}{0pt}
\begin{algorithm}[t]
\caption{pFedLVM}
\label{alg:pFLVM}
\KwIn{$\omega_{lvm}$, $\mathcal{D}_{v}$, where $v$ = $1, 2, \cdots, |\mathcal{V}|$}
\KwOut{$\omega_{v, c}, \omega_{v, p}$}
   \nl  \For {Batch $j=1,\cdots$}{
   \nl \textbf{Vehicle Side:} \\
   \nl \For {Vehicle $v=1,\cdots,|\mathcal{V}|$ \textbf{in parallel}}{
            \nl Feature Compression: $\mathcal{F}_{v, c}^{(j)}=\omega_{v, c}(\mathcal{D}_v^{(j)})$ \\
            \nl Send $\mathcal{F}_{v, c}^{(j)}$ to Server \\
            \nl Wait until receive $\mathcal{F}_{shd}^{(j)}$ from server \\
            \nl Update $\omega_{v, c}$ via \Cref{Eq:comp_loss} \\
            \nl Update $\omega_{v, p}$ via \Cref{Eq:perc_loss} \\
	}
    \nl \\
    \nl \textbf{Server Side:} \\
    \nl Receive $\mathcal{F}_{1, c}^{(j)},\cdots,\mathcal{F}_{|\mathcal{V}|, c}^{(j)}$ \\
    \nl Concatenate $\mathcal{F}_{1, c}^{(j)},\cdots,\mathcal{F}_{|\mathcal{V}|, c}^{(j)}$: 
    $\mathcal{F}_{cat}^{(j)} = \bigoplus_{v=1}^{|\mathcal{V}|} \mathcal{F}_{v, c}^{(j)}$ \\
    \nl Extract shared features: $\mathcal{F}_{shd}^{(j)}=\omega_{lvm}(\mathcal{F}_{cat}^{(j)})$ \\
    \nl Return $\mathcal{F}_{shd}^{(j)}$ to all vehicles \\
    }
\end{algorithm}

\section{Communication overhead and complexity analyses of pFedLVM}
\label{overhead_complexity}

\subsection{Communication Efficiency Analysis of pFedLVM}
\begin{figure}[tp]
\centering
\vspace{-2.0cm}
\subfloat[The protocol of exchanging features in pFedLVM]{\includegraphics[width=\linewidth]{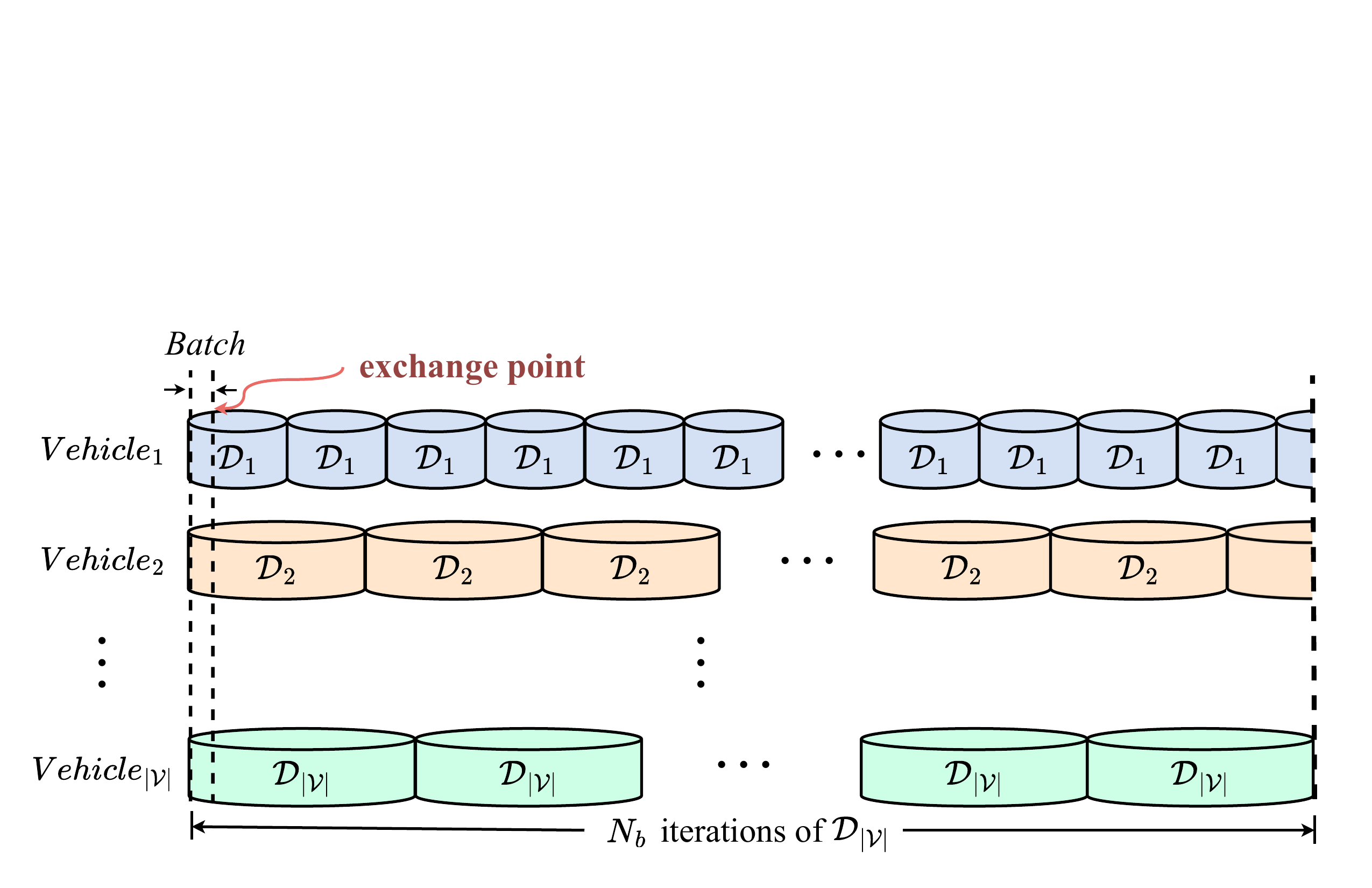}%
\label{Fig.comm_effi_analysis_pfedlvm}}

\vspace{-2.0cm}
\subfloat[The protocol of exchanging LVMs in typical FL]{\includegraphics[width=\linewidth]{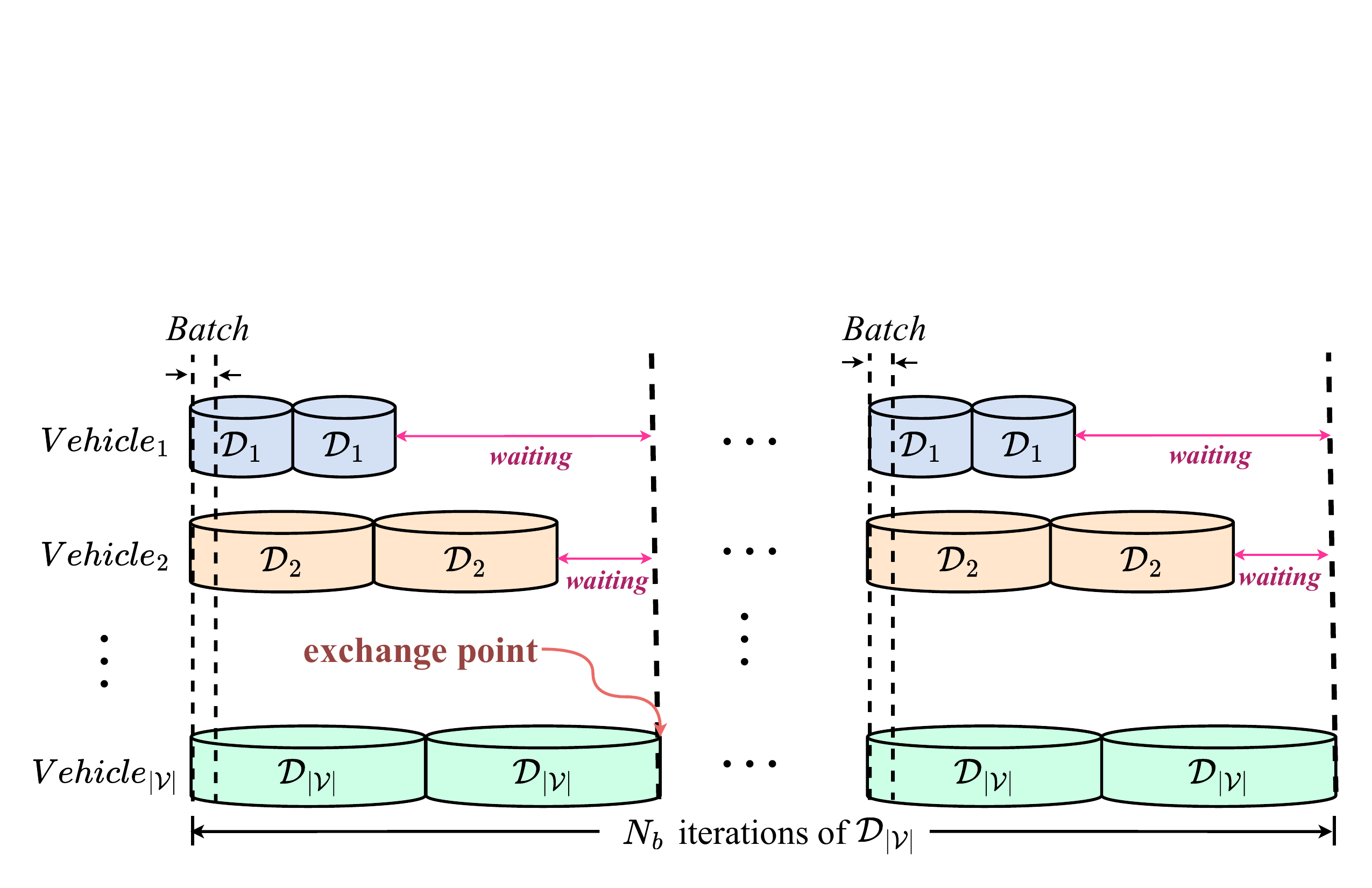}%
\label{Fig.comm_effi_analysis_fl}}
\caption{The illustration for communication overhead analyses. In this figure, we assume the $Vehicle_{|\mathcal{V}|}$ has the largest dataset among all involved vehicles, \ie, $S_{max}$ = $|\mathcal{D}_{|\mathcal{V}|}|$. \textbf{(a)} In the proposed pFedLVM, the feature exchanges occur at the end of each mini-batch (denoted by the ``exchange point'' in the figure). The vehicles with smaller dataset may repeat more than one iteration to align with the $Vehicle_{|\mathcal{V}|}$. \textbf{(b)} In the typical FL, the LVM exchanges occur per certain local iterations (here is 2)}. The vehicles with smaller dataset need to wait to align with the $Vehicle_{|\mathcal{V}|}$.
\label{Fig.comm_effi_analysis}
\vspace{0.3cm}
\end{figure}

\Cref{Fig.comm_effi_analysis} shows the communication paradigms of two cases: 
I) the proposed pFedLVM which trains personalized models of each vehicle via feature sharing with the central server; 
II) deploying the LVMs in the FedAD system and train the global LVM in a typical FL way, where LVM parameters are exchanged between vehicles and the central server. 

In order to compare the communication overheads of the two schemes, we introduce some symbols. Firstly, let $S_{max}$ represents the size of the largest dataset (dubbed as $\mathcal{D}_{max}$) among all involved vehicles included in the FedAD system, \ie, $S_{max}$ = $|\mathcal{D}_{max}|$ = $\max_{v \in \mathcal{V}}\{|\mathcal{D}_{v}|\}$. To align with the two considered cases, we assume they both execute a total of $N_b$ iterations of $\mathcal{D}_{max}$. Furthermore, we denote $B_s$, $M_b$ and $F_b$ as the min-batch size, the size of the LVMs, and feature size of one mini-batch of input images, respectively. Generally, $M_b$ is determined by the architecture of LVMs, while $F_b$ is dependant on the architecture of the feature compressor in pFedLVM. At last, let $\sigma$ represents the number of local iterations of each vehicle between two adjacent aggregations in the typical FL.

Based on the above defined symbols, we can derive the communication overhead of the proposed pFedLVM. As illustrated in \Cref{Fig.comm_effi_analysis_pfedlvm}, in each round of local update, there are $\lfloor\frac{S_{max}}{B_s}\rfloor$ mini-batches of features to exchange between each vehicle and the central server, where the symbol ``$\lfloor\cdot\rfloor$'' is a the floor function, which returns the largest integer that is less than or equal to the value between ``$\lfloor$'' and ``$\rfloor$''. Therefore, for each vehicle, the communication overhead in each round of local update for both upload and download is $F_b \times \lfloor\frac{S_{max}}{B_s}\rfloor \times 2$. As there are $|\mathcal{V}|$ vehicles in the system and we consider $N_b$ iterations of local updates, the total communication overheads is 
\begin{align}
   M_{pFL} &= N_b \times (F_b \times \lfloor\frac{S_{max}}{B_s}\rfloor \times 2) \times |\mathcal{V}|. 
   \label{Eq:overhead_pfedlvm}
\end{align}

On the other hand, for the typical FL, as showcased in \Cref{Fig.comm_effi_analysis_fl}, it exchanges the parameters of the LVM once after $\sigma$ local updates at the vehicle level. Therefore, the number of rounds for parameter exchange is $\lfloor\frac{N_b}{\sigma}\rfloor$. Since each round of parameter exchange involves both uplink and downlink, the total communication overhead for the typical FL system is
\begin{align}
   M_{FL} &= \lfloor\frac{N_b}{\sigma}\rfloor \times M_b \times 2 \times |\mathcal{V}|. 
\label{Eq:overhead_fl_lvm}
\end{align}

To gain more insight into the communication overheads reduction for exchanging features in pFedLVM compared to exchanging LVMs in typical FL, we can define the saving as 
\begin{align}
   \eta &= (1 - \frac{M_{pFL}}{M_{FL}}) \times 100\%
   \nonumber 
   \\
   &=(1 - \frac{F_b}{M_b} \times \frac{N_b \times \lfloor\frac{S_{max}}{B_s}\rfloor}{\lfloor\frac{N_b}{\sigma}\rfloor}) \times 100\%
   \nonumber
   \\
   &\stackrel{N_b \gg \sigma}{\approx} (1 - \frac{F_b}{M_b} \times \lfloor\frac{S_{max}}{B_s}\rfloor \times \sigma) \times 100\%.
\label{Eq:overhead_reduction}
\end{align}
From \Cref{Eq:overhead_reduction}, we can observe that communication resource reduction of the proposed pFedLVM depends on multiple factors, including the mini-batch size $B_s$, the size of the largest dataset among all involved vehicles $S_{max}$, the feature size of each mini-batch $F_b$, the size of the adopted LVMs $M_b$, and the aggregation interval $\sigma$ in typical FL. 

\subsection{Complexity Analysis of pFedLVM}
In this section, we will conduct the space and time complexity analyses of the proposed pFedLVM.

\subsubsection{Space Complexity Analysis}
Recall that in the proposed pFedLVM, there are one central server and $|\mathcal{V}|$ AD vehicles. Assume that the compressed features $\mathcal{F}_{v, c}^{(\cdot)}$ in each vechicle and the shared feature $\mathcal{F}_{shd}^{(\cdot)}$ of each mini-batch share the same size (denoted as $F_b$ in the previous subsection). Therefore, for each vehicle, it needs extra $F_b$ space units to store the compressed feature, and in total it needs $|\mathcal{V}| \times F_b$ space units for all involved vehicles. For the central server, it also needs $|\mathcal{V}| \times F_b$ space units to store the received features from involved vehicles. Therefore, the space complexity of the proposed pFedLVM is O($|\mathcal{V}|$).

\begin{figure}[!t]
\centering
\vspace{-0.25cm}
\includegraphics[width=\linewidth,height=\linewidth]{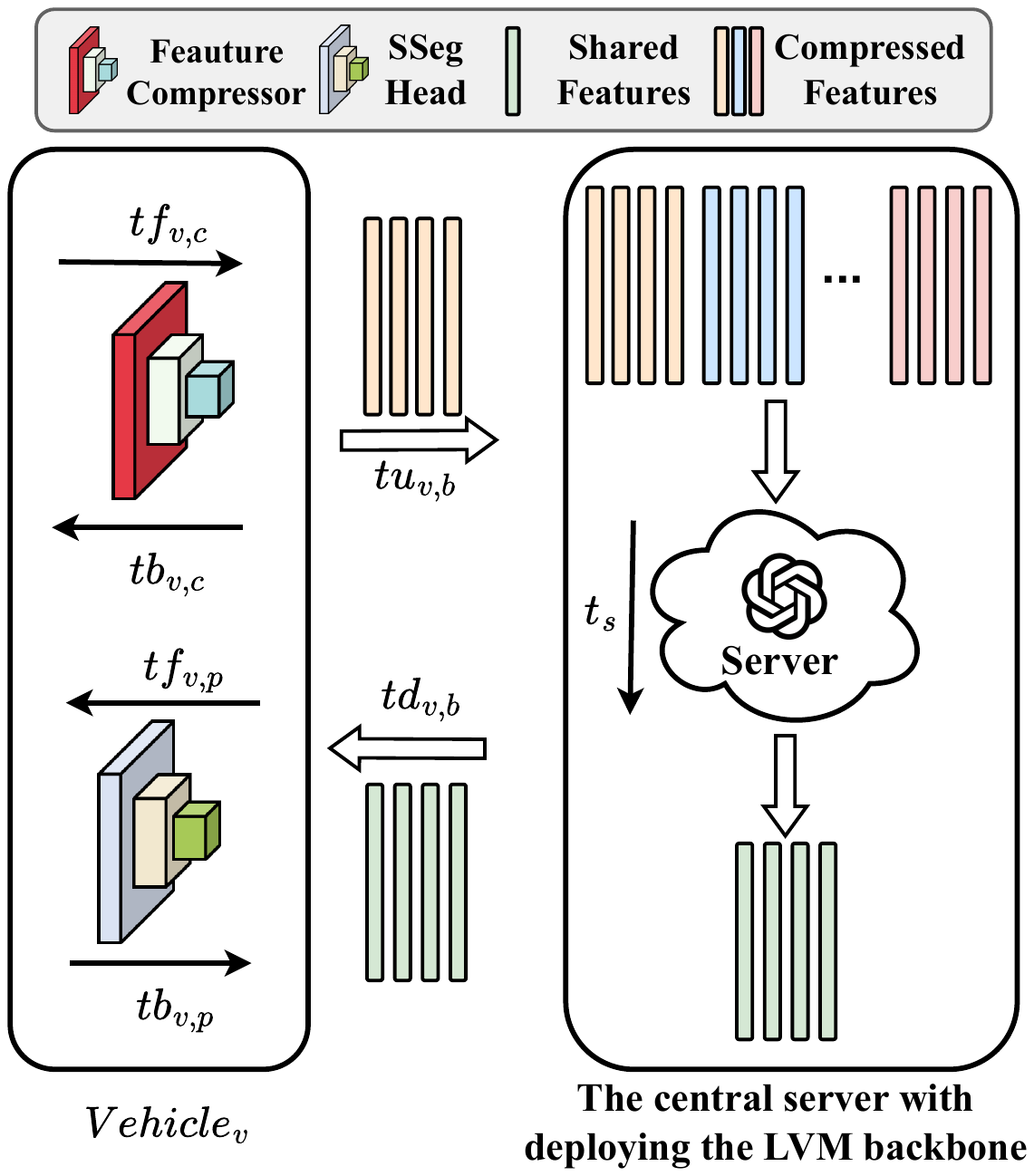}
\vspace{-2.2cm}
\caption{Illustration of the involved times of the $b$-th feature exchange for $Vehicle_v$ in pFedLVM. Specifically, $Vehicle_v$ takes $tf_{v, c}$ to compress the latent features and $tu_{v, b}$ to send the compressed features to the server. Then the server takes $t_s$ to extract the shared features by using LVM backbone and $td_{v, b}$ to send the shared features back to $Vehicle_v$. Subsequently, on the one hand, $Vehicle_v$ takes $tb_{v, c}$ to optimize the Feature Compressor via back propagation; on the other hand, $Vehicle_v$ takes $tf_{v, p}$ to predict the semantic masks and $tb_{v, p}$ to optimize the SSeg Head via back propagation.}
\label{Fig.time_complexity}
\vspace{0.2cm}
\end{figure}

\subsubsection{Time Complexity Analysis}
As illustated in \Cref{Fig.time_complexity}, for the $b$-th min-batch, the forward time and backward time of feature compressor and SSeg head for $Vehicle_{v}$ are denoted as $tf_{v, c}$, $tb_{v, c}$, $tf_{v, p}$, and $tb_{v, p}$, respectively; the computing time in the central server to extract the shared feature is denoted as $t_s$; the communication time for uploading and downloading features are denoted as $tu_{v, b}$ and $td_{v, b}$, respectively. 

As demonstrated in \Cref{Fig.time_complexity}, the total time for the $b$-th mini-batch can be divided into three stages: I) for the first stage, the central server needs to take $\max_{v \in \mathcal{V}}\{tf_{v, c}+tu_{v, b}\}$ to receive the compressed features from all involved vehicles. II) for the second stage, once the central server received all features, it takes the time of $t_s$ to extract the shared features. III) for the third stage, $Vehicle_v$ takes $td_{v, b}$ to download the shared features, and takes $\max(tb_{v, c}, tf_{v, p}+tb_{v, p})$ to optimize its feature compressor and SSeg head.
Therefore, the total time in third stage is $\max(tb_{v, c}, tf_{v, p}+tb_{v, p})+td_{v,b}$. For all involved vehicles, the total time of the third stage is $\max_{v \in \mathcal{V}}\{\max(tb_{v, c}, tf_{v, p}+tb_{v, p})+td_{v,b}\}$. Therefore, we can obtain the time of pFedLVM in the $b$-th mini-batch is
\begin{align}
   t_b &=\max_{v \in \mathcal{V}}\{tf_{v, c}+tu_{v, b}\} + t_s + \nonumber \\ 
   &~~~\max_{v \in \mathcal{V}}\{\max(tb_{v, c}, tf_{v, p}+tb_{v, p})+td_{v,b}\}.
\label{Eq:batc_time}
\end{align}
We further assume that all personalized models in pFedLVM finally converge after the training of $N_b$ iterations of $\mathcal{D}_{max}$. Therefore, the total time $t$ is
\begin{align}
   t = \sum_{b=1}^{N_b} t_b
   \leq N_b \times \max_{b \in \{1, 2, \ldots, N_b\}}\{t_b\},
\label{Eq:all_time}
\end{align}
and the time complexity of pFedLVM is O($N_b$).

\section{Experiments}
\label{experiments}

In this section, we present experimental results to verify the proposed pFedLVM framework for the street scene semantic understanding task in the context of AD. Our experiments assess the performance of LVM backbone and the proposed personalized learning mechanism compared with some existing FL benchmarks, including FedAvg \cite{mcmahan2017communication}, FedProx \cite{li2020federated} and FedDyn \cite{acar2021federated}. 
Given that the control, planning, and end-to-end heads could employ update mechanism analogous to that used in the SSeg head, we focus on the SSeg head as a representative demonstration on the effectiveness of our proposed pFedLVM framework.

\subsection{Datasets, Metrics and Implementation}
\subsubsection{Datasets}
We employ two public datasets in our experiments: Cityscapes dataset \cite{Cordts2016Cityscapes} and CamVid dataset \cite{brostow2008segmentation}. Cityscapes dataset, captured across multiple cities, comprises 2,975 training images and 500 validation images. The training dataset features pixel-level labels for 19 classes, such as vehicles, pedestrians, etc. To simulate the practical scenario where different vehicles might have different amount of data, we partition this training dataset randomly for multiple vehicles. On the other hand, the CamVid dataset includes a total of 701 samples, each with a pixel-level label for 11 classes. In the experiments, we divide randomly selected 600 samples into different vehicles. The remaining samples serve as test dataset. 

\begin{table}[tp]
\setlength{\tabcolsep}{25.0pt}
\caption{The number of RGB images on each vehicle}
\begin{tabularx}{\linewidth}{ccc}
\hline
\multirow{2}{*}{Vehicle ID} & \multicolumn{2}{c}{The Number of RGB Images} \\ \cline{2-3} 
                            & Cityscapes         & CamVid        \\ \hline
Vehicle \#1                 & 848                & 128            \\
Vehicle \#2                 & 1046                & 167           \\
Vehicle \#3                 & 1081                & 305            \\ \hline
Total                       & 2975               & 600           \\ \hline
\end{tabularx}
\label{Tab:number_for_vehicles}
\end{table}

\begin{table}[tp]
\vspace{0.1cm}
    \centering
    \renewcommand{\arraystretch}{1.0}
    \setlength{\tabcolsep}{15.0pt}
    \caption{Hardware/Software configurations}
    \begin{tabularx}{\linewidth}{ll}
    \hline
        \textbf{Items} & \textbf{Configurations} \\ \hline
        CPU  & AMD Ryzen 9 3900X 12-Core \\ 
        GPU  & NVIDIA GeForce 4090 $\times$ 2 \\ 
        RAM  & DDR4 32G \\ 
        DL Framework  & PyTorch @ 2.1.1+cu121 \\ 
        GPU Driver  & 530.30.02 \\ 
        CUDA  & 12.1 \\ 
        cuDNN  & 8902 \\ \hline
    \end{tabularx}
\label{Tab:configs}
\end{table}

\begin{table}[tp]
    \centering
    \renewcommand{\arraystretch}{1.0}
    \setlength{\tabcolsep}{15.0pt}
    \caption{Training configurations}
    \begin{tabularx}{\linewidth}{ll}
    \hline
        \textbf{Items} & \textbf{Configurations} \\ \hline
        Loss $\mathcal{E}_{c}$  & nn.MSELoss  \\ 
        Loss $\mathcal{E}_{p}$  & nn.CrossEntropyLoss \\ 
        Feature Compressor Architecture   & CNN \\ 
        SSeg Head Architecture  & ASSP \cite{chen2017deeplab} \\ 
        Optimizer  & nn.Adam \\ 
        Adam Betas  & (0.9, 0.999) \\ 
        Weight Decay  & 1e-4 \\ 
        Batch Size  & 8 \\ 
        Learning Rate  & 3e-4 \\ 
        \hline
    \end{tabularx}
\label{Tab:train}
\vspace{0.2cm}
\end{table}

\subsubsection{Evaluation Metrics}
\label{eval_metrics}
We assess the proposed pFedLVM using four metrics: \textbf{mIoU}: the mean of intersection over union; \textbf{mPrecision (mPre for short)}: the mean ratio of true positive pixels to the total predicted positive pixels; \textbf{mRecall (mRec for short)}: the mean ratio of true positive pixels to the total positive ground truth pixels; \textbf{mF1}: the mean of harmonic mean of precision and recall, providing a balanced measure of these two metrics. Such metrics are evaluated across all semantic classes, offering a comprehensive view of pFedLVM's performance. These metrics are formally listed in \Cref{Eq:mF1}:
\begin{align}
    &mIoU = \frac{1}{\mathcal{C}}\sum_{c=1}^{\mathcal{C}}IoU_c = \frac{1}{\mathcal{C} \mathcal{N}}\sum_{c=1}^{\mathcal{C}} \sum_{n=1}^{\mathcal{N}} \frac{TP_{n, c}}{FP_{n, c}\hspace{-0.1cm}+\hspace{-0.1cm} TP_{n, c}\hspace{-0.1cm} +\hspace{-0.1cm} FN_{n, c}},
    \nonumber
    \\
    \vspace{-0.2cm}
    &mPre = \frac{1}{\mathcal{C}}\sum_{c=1}^{\mathcal{C}}Pre_c = \frac{1}{\mathcal{C} \mathcal{N}}\sum_{c=1}^{\mathcal{C}} \sum_{n=1}^{\mathcal{N}} \frac{TP_{n, c}}{FP_{n, c} + TP_{n, c}},
    \nonumber
\end{align}
\begin{align}
    &mRec = \frac{1}{\mathcal{C}}\sum_{c=1}^{\mathcal{C}}Rec_c = \frac{1}{\mathcal{C} \mathcal{N}}\sum_{c=1}^{\mathcal{C}} \sum_{n=1}^{\mathcal{N}} \frac{TP_{n, c}}{TP_{n, c} + FN_{n, c}},
    \nonumber
    \\
    \vspace{-0.2cm}
    &mF1 = \frac{1}{\mathcal{C}}\sum_{c=1}^{\mathcal{C}}F1_c = \frac{1}{\mathcal{C}}\sum_{c=1}^{\mathcal{C}} \frac{2 * Pre_c * Rec_c}{Pre_c + Rec_c},
    \label{Eq:mF1}
\end{align}
where $TP$, $FP$, $TN$ and $FN$ stand for True Positive, False Positive, True Negative and False Negative, respectively. $\mathcal{C}$ denotes the number of semantic classes within the test dataset, with values set to 19 for the Cityscapes dataset and 11 for the CamVid dataset. Similarly, $\mathcal{N}$ signifies the size of the test dataset, which amounts to 500 for Cityscapes and 101 for CamVid.

\subsubsection{Implementation Details}
The main hardware and software configurations are summarized in \Cref{Tab:configs}, and the major training details are given in \Cref{Tab:train}.

\subsection{Evaluation of LVM backbone and SSeg Head}
The publicly available pre-trained iGPT outputs multiple layers of hidden features. These features are essentially representations of the input data as understood by the model at various levels of abstraction, complexity and capability. In this section, we will present comprehensive experimental results to evaluate the performance of iGPT for the street scene semantic understanding task within the context of AD. 
Furthermore, we also conduct additional experiments to evaluate the performance of the LVM+Head framework. Our objective is to benchmark its efficacy against current SOTA models: BiSeNetV2 \cite{yu2021bisenet} and SegNet \cite{badrinarayanan2017segnet}, with all the metrics mentioned in subsection \ref{eval_metrics}.

\begin{figure*}[!t]
\centering
\subfloat[mIoU on Cityscapes]{\includegraphics[width=0.25\linewidth,height=0.15\linewidth]{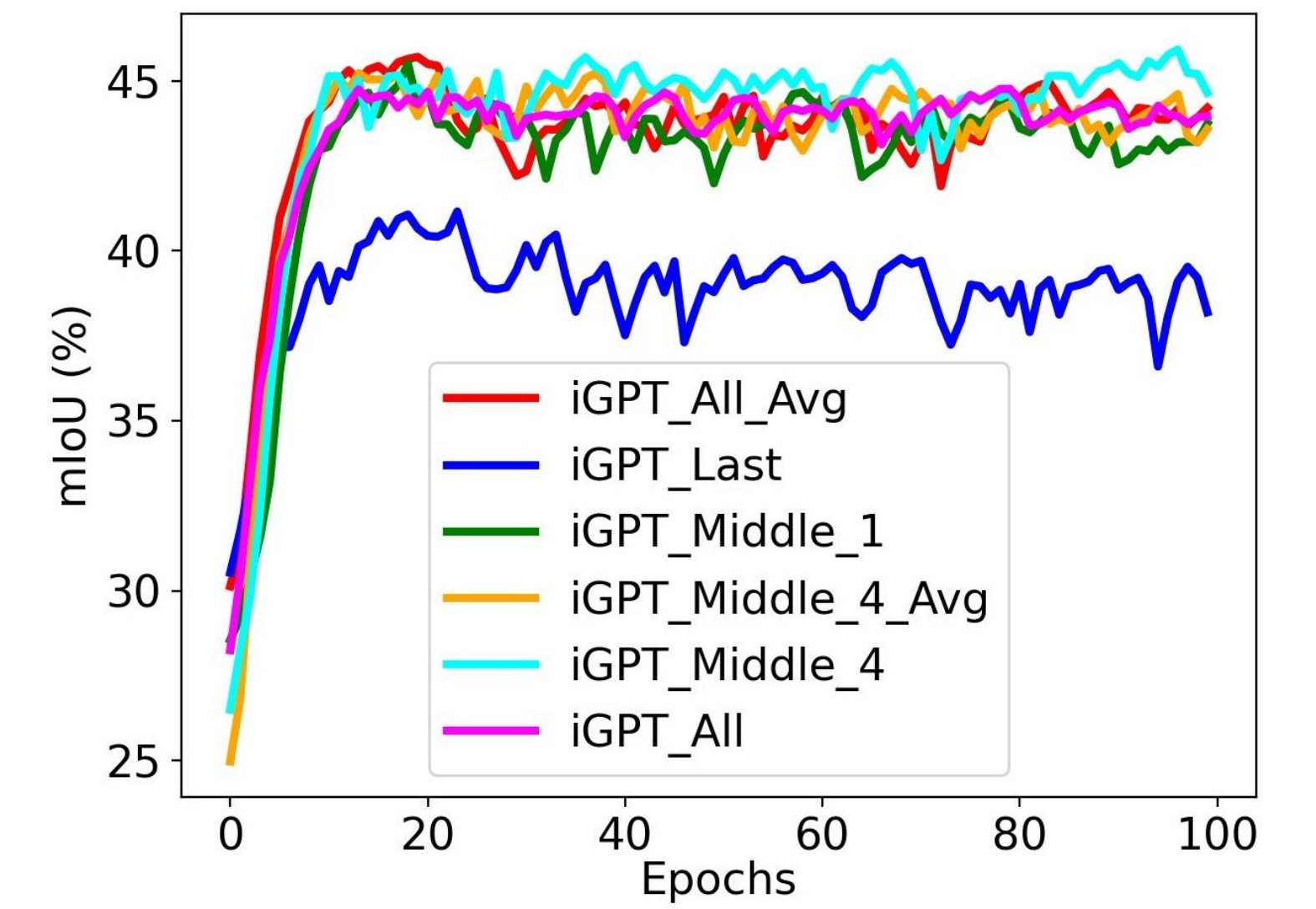}%
\label{Fig:hidden_feats_e}}
\subfloat[mPrecision on Cityscapes]{\includegraphics[width=0.25\linewidth,height=0.15\linewidth]{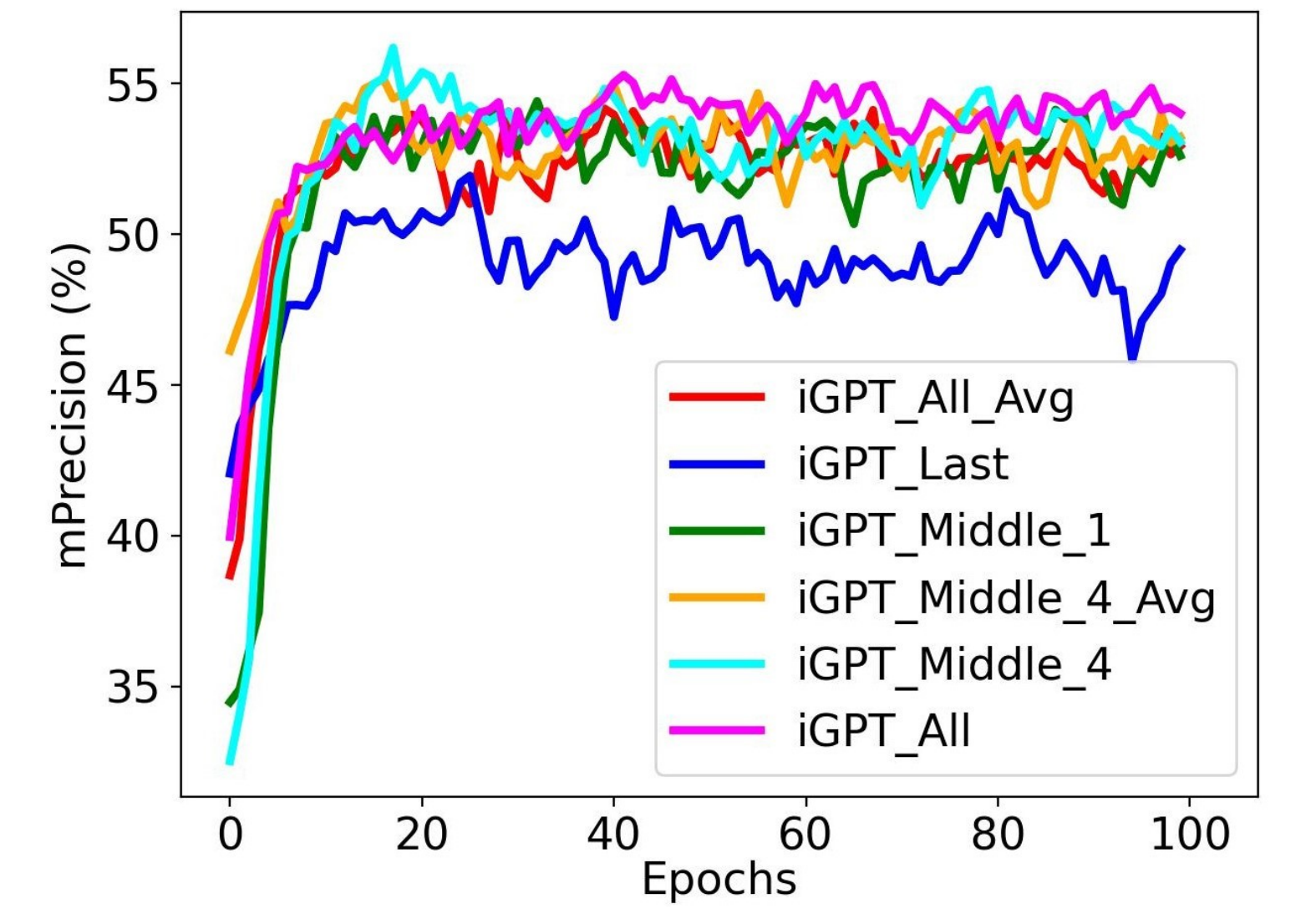}%
\label{Fig:hidden_feats_f}}
\subfloat[mRecall on Cityscapes]{\includegraphics[width=0.25\linewidth,height=0.15\linewidth]{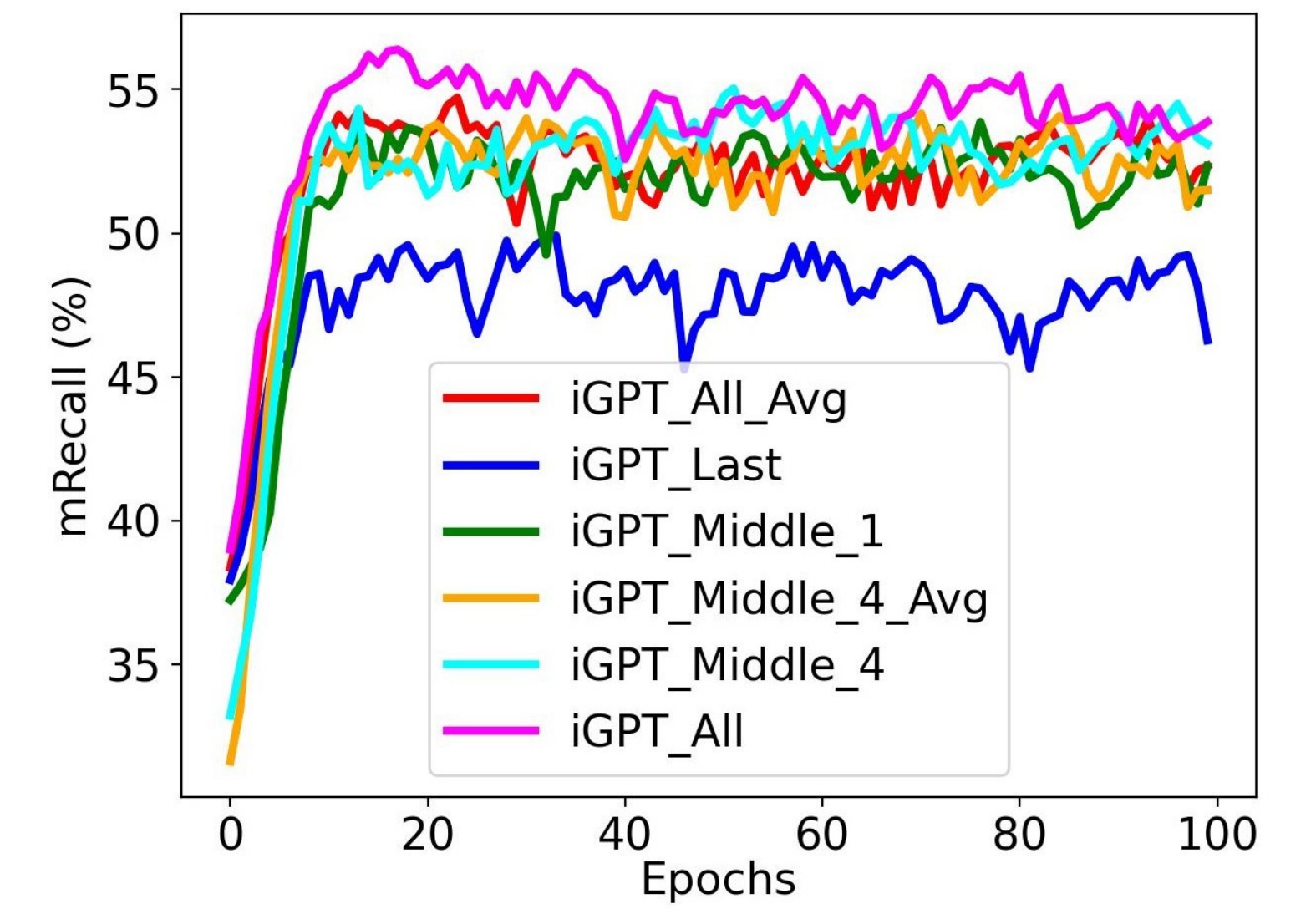}%
\label{Fig:hidden_feats_g}}
\subfloat[mF1 on Cityscapes]{\includegraphics[width=0.25\linewidth,height=0.15\linewidth]{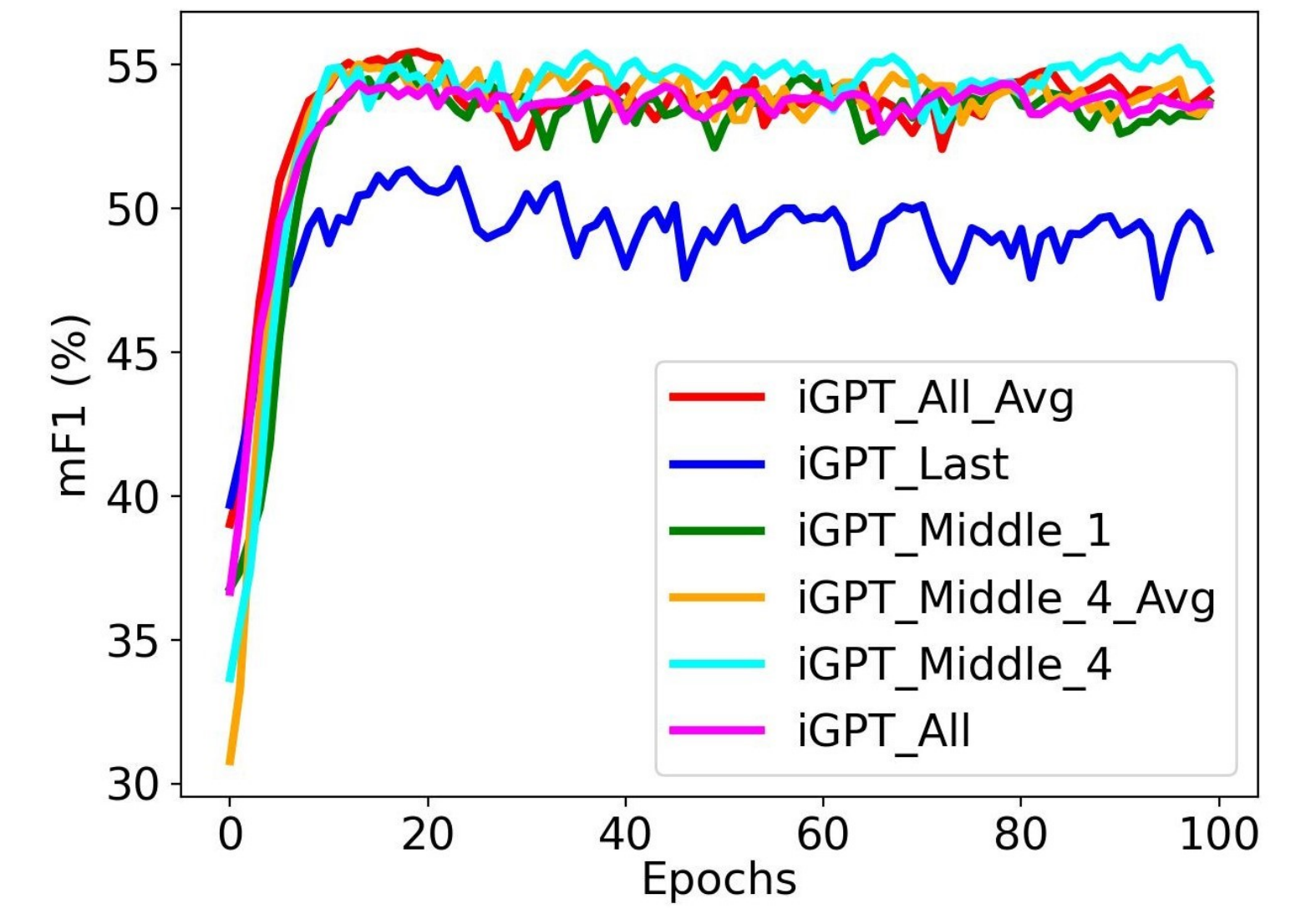}%
\label{Fig:hidden_feats_h}}

\subfloat[mIoU on CamVid]{\includegraphics[width=0.245\linewidth,height=0.15\linewidth]{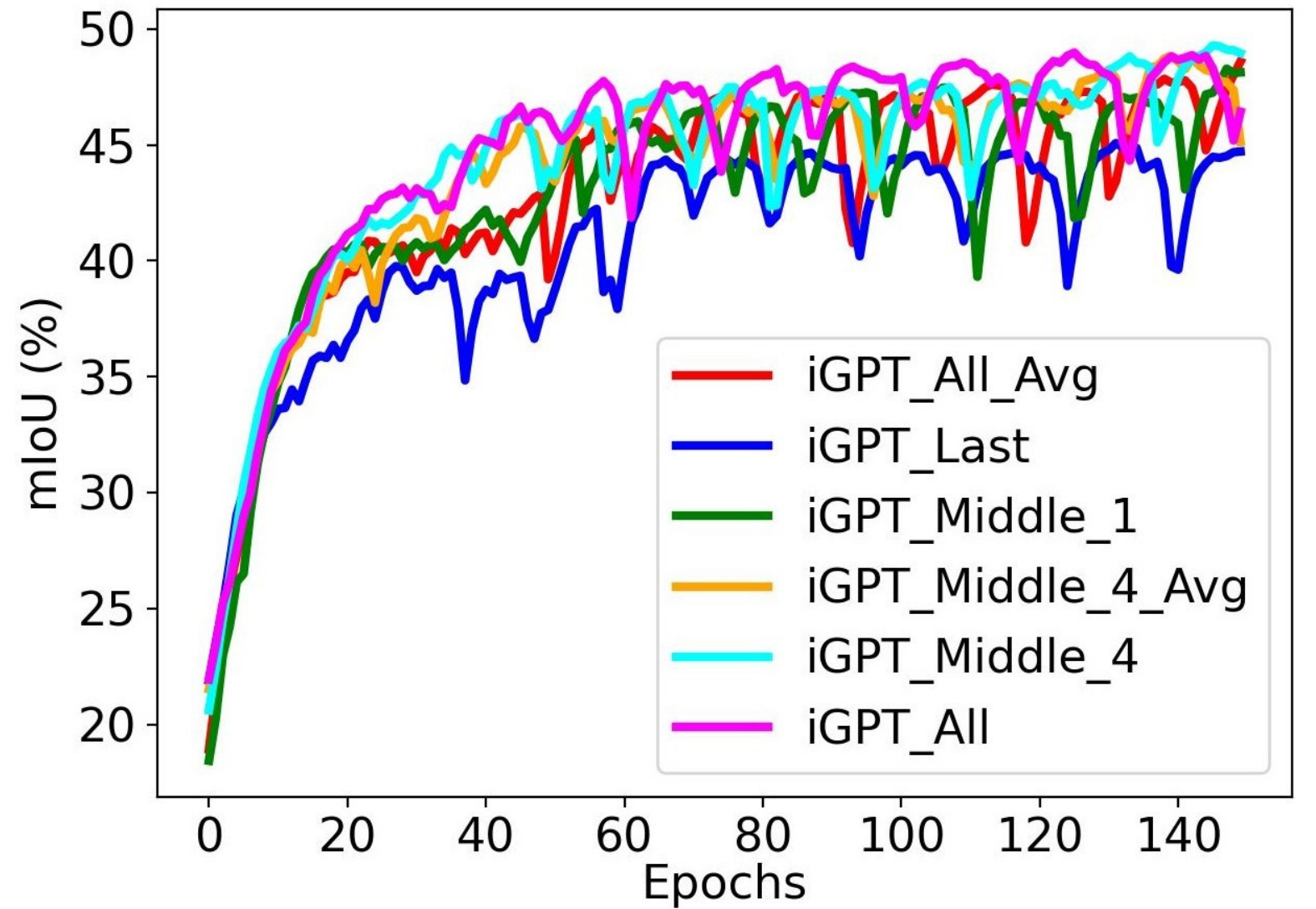}%
\label{Fig:hidden_feats_a}}
\subfloat[mPrecision on CamVid]{\includegraphics[width=0.245\linewidth,height=0.15\linewidth]{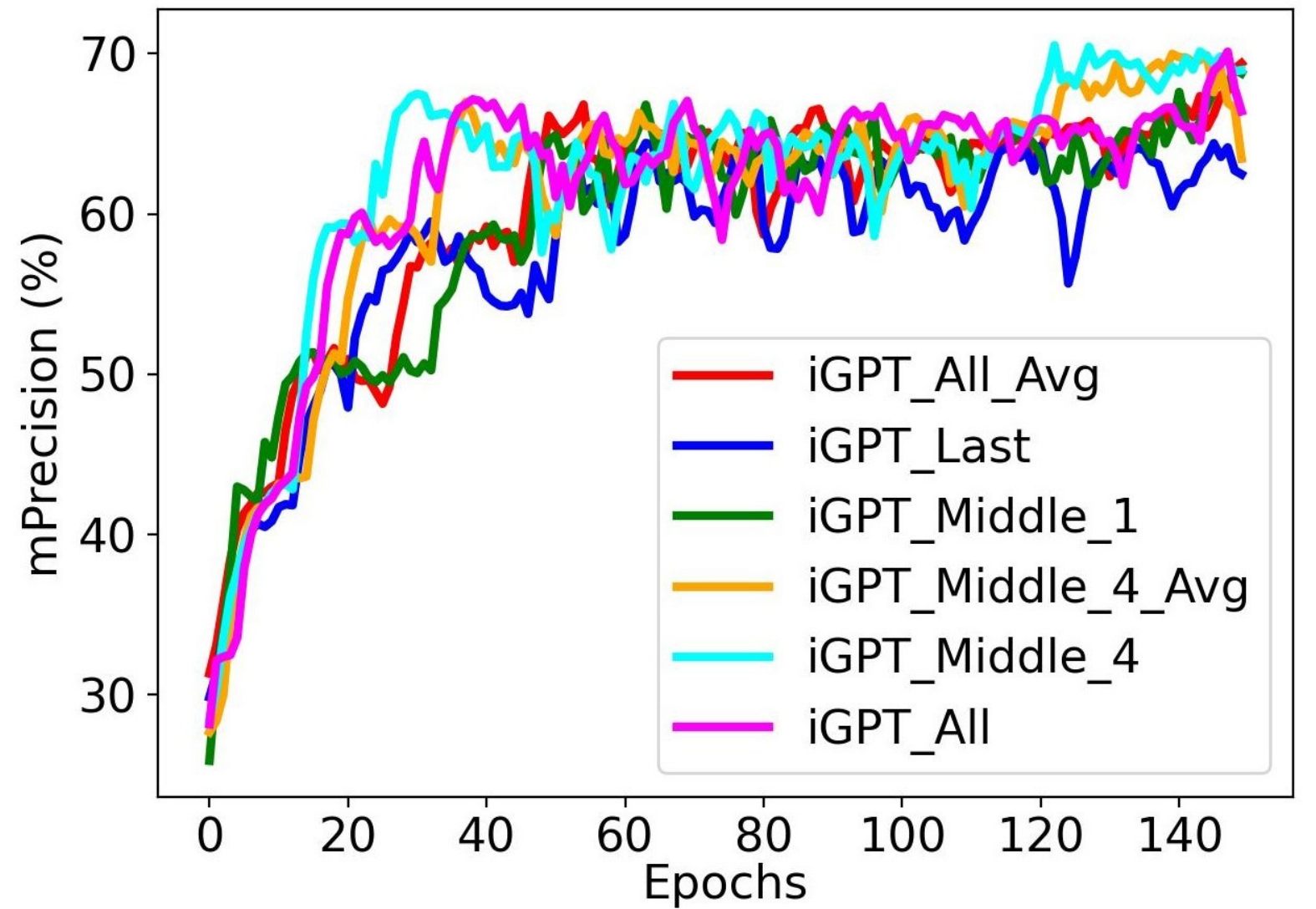}%
\label{Fig:hidden_feats_b}}
\subfloat[mRecall on CamVid]{\includegraphics[width=0.245\linewidth,height=0.15\linewidth]{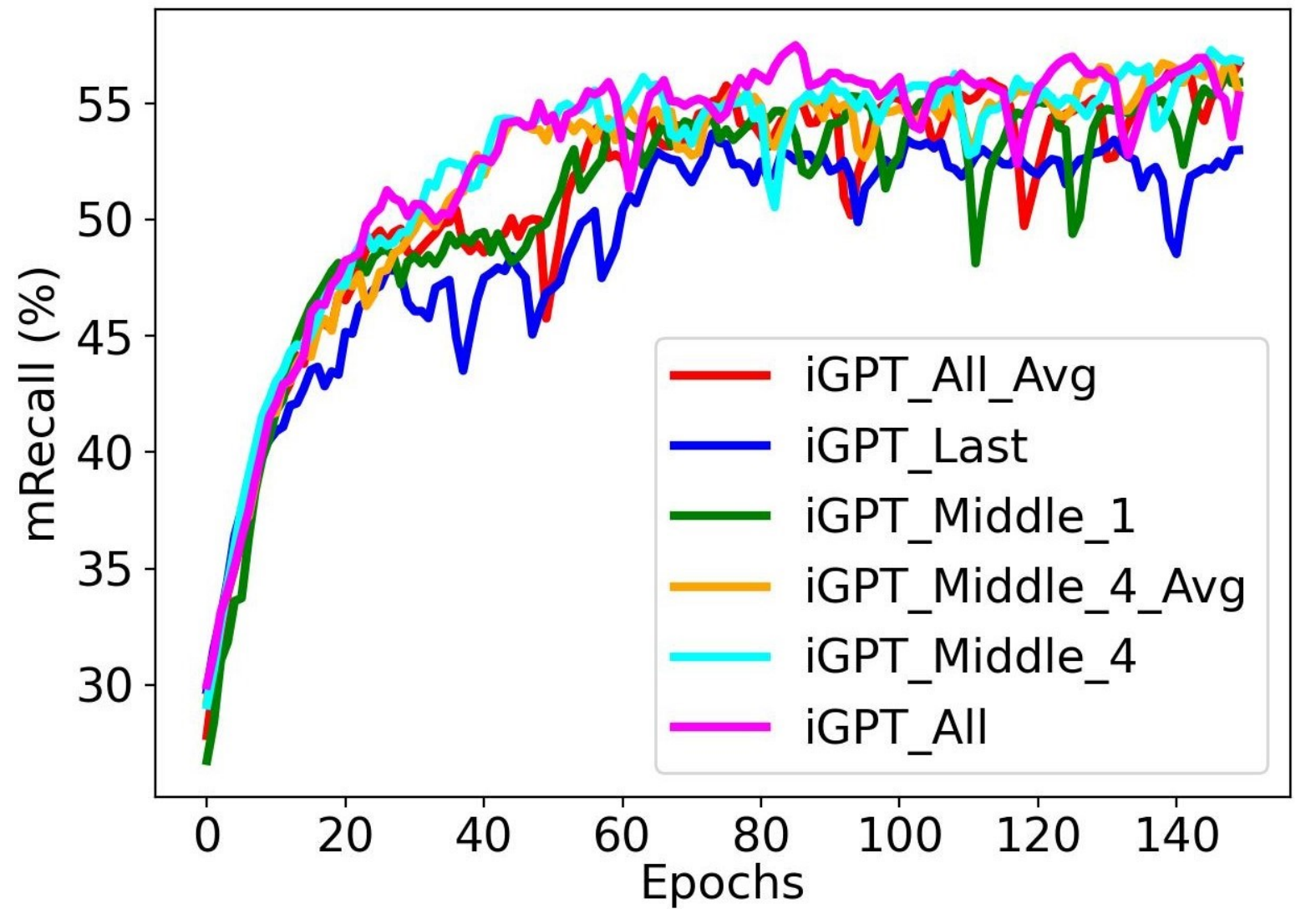}%
\label{Fig:hidden_feats_c}}
\subfloat[mF1 on CamVid]{\includegraphics[width=0.245\linewidth,height=0.15\linewidth]{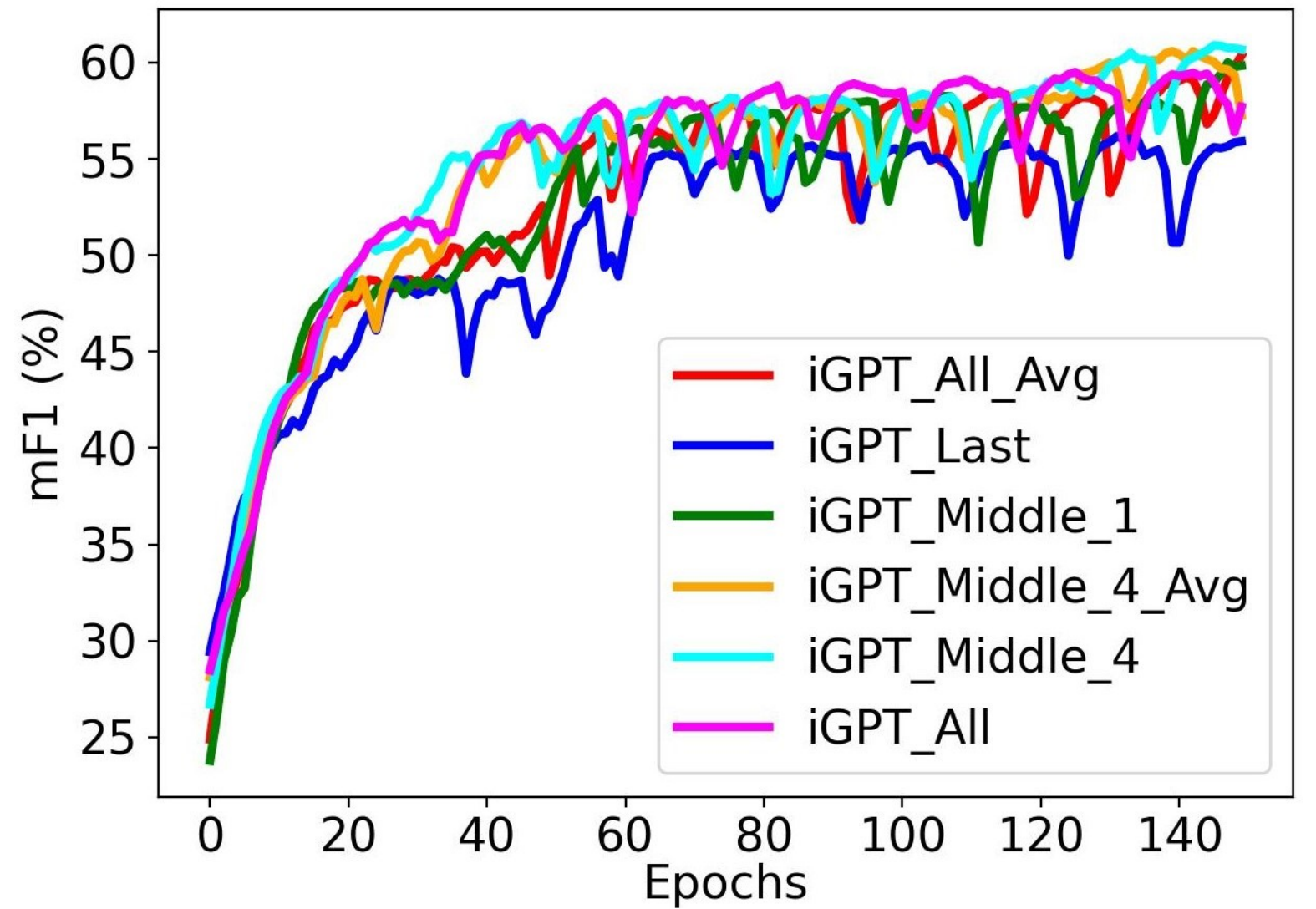}%
\label{Fig:hidden_feats_d}}
\caption{Comparison of using different hidden features on Cityscapes and CamVid datasets.}
\label{Fig:hidden_feats}
\end{figure*}

\begin{table*}[tp]
\centering
\setlength{\tabcolsep}{12.0pt}
\caption{Performance comparison of considered iGPT hidden features on Cityscapes and CamVid datasets}
\begin{tabularx}{\linewidth}{ccccclcccc}
\hline
\multirow{2}{*}{Feature Layer(s)} & \multicolumn{4}{c}{Cityscapes Dataset (19 Semantic Classes) (\%)}          &  & \multicolumn{4}{c}{CamVid Dataset (11 Semantic Classes) (\%)}              \\ \cline{2-5} \cline{7-10} 
                                  & mIoU           & mF1            & mPrecision     & mRecall        &  & mIoU           & mF1            & mPrecision     & mRecall        \\ \hline
iGPT\_All\_Avg                    & 43.70          & 53.45          & 54.16          & 54.71          &  & 48.59          & 60.37          & 69.40          & 56.67          \\
iGPT\_Last                        & 39.15          & 49.37          & 51.94          & 49.92          &  & 45.07          & 56.18          & 64.57          & 53.69          \\
iGPT\_Middle\_1                   & 43.49          & 53.24          & 54.41          & 53.87          &  & 48.28          & 60.00          & 69.03          & 56.37          \\
iGPT\_Middle\_4\_Avg              & 43.22          & 53.02          & 55.14          & 54.16          &  & 48.84          & 60.59          & 69.99          & 56.74          \\
iGPT\_Middle\_4                   & \textbf{45.81} & \textbf{55.15} & \textbf{56.18} & 56.00 &  & \textbf{49.29} & \textbf{60.90} & \textbf{70.52} & 57.27          \\
iGPT\_All                         & 44.76          & 54.34          & 55.28          & \textbf{56.38}              &  & 48.98          & 59.51          & 70.12          & \textbf{57.47} \\ \hline
\end{tabularx}
\label{Tab:iGPT_hidden_feats}
\vspace{-0.4cm}
\end{table*}

\subsubsection{Hidden Feature Selection from iGPT}
iGPT mentioned earlier contains multiple hidden layers included in its output. Previous research \cite{chen2020generative} suggested that features in the middle of the output hidden layers perform the best for training a linear classification model. However, which layer (or layers) performs optimally for street scene semantic understanding task in the context of AD remains an open question.

In our experiments, we aim to identify the most effective features by comparing the following six different options: I) Using features from the last layer (termed as iGPT\_Last); II) Using features from the middle layer (termed as iGPT\_Middle\_1); III) Using the averaging features from all layers (termed as iGPT\_ALL\_Avg); IV) Using the averaging features from the middle four layers (termed as iGPT\_Middle\_4\_Avg); V) Using features from the middle four layers (termed as iGPT\_Middle\_4); VI) Using features from all the layers (termed as iGPT\_ALL). This comprehensive comparison will help us identify the optimal layer(s) configuration in the AD context.

The results of this experiment are presented in \Cref{Fig:hidden_feats}, with \Cref{Fig:hidden_feats_e,Fig:hidden_feats_f,Fig:hidden_feats_g,Fig:hidden_feats_h} illustrating four different metrics for the Cityscapes dataset, while \Cref{Fig:hidden_feats_a,Fig:hidden_feats_b,Fig:hidden_feats_c,Fig:hidden_feats_d} shows the corresponding results for CamVid dataset. It is noticed that all four metrics show similar conclusions. In particular, it is evident that multiple layers performs better than single layer. For instance, from \Cref{Fig:hidden_feats_e} to \Cref{Fig:hidden_feats_h}, iGPT\_All and iGPT\_Middle\_4 exhibit better mIoU values and fluctuate less compared to other options. Furthermore, the averaging of multiple layers, such as iGPT\_All\_Avg and iGPT\_Middle\_4\_Avg, can outperform single layer but underperform multiple layers, as they smooth out the details of features in different layers. Moreover, a comparison between iGPT\_Last and iGPT\_Middle\_1 reveals a notable distinction: features extracted from the middle layer exhibit better performance over those from the last layer in the context of downstream semantic segmentation tasks. This observation is in line with the results reported by \cite{chen2020generative}. Notably, the performance disparity between iGPT\_Last and the other considered options is markedly more pronounced, underscoring the conclusion that iGPT\_Last represents the least favorable option for tasks involving street scene semantic understanding. Analyzing the evaluation metrics on CamVid dataset in \Cref{Fig:hidden_feats_a,Fig:hidden_feats_b,Fig:hidden_feats_c,Fig:hidden_feats_d}, we observe that they follow similar patterns to that of Cityscapes dataset. It is worth noting that the gap between iGPT\_Last and the other options in CamVid is narrower than that of Cityscapes, which is likely due to the smaller data complexity of CamVid than that of Cityscapes dataset. 

For gaining insight of the aforementioned observations and analysis, \Cref{Tab:iGPT_hidden_feats} provides a more quantitative perspective. Specifically, for both Cityscapes and CamVid datasets, the multiple layers options (termed as iGPT\_All and iGPT\_Middle\_4) achieve the best performance in almost all evaluation metrics. It is noteworthy that iGPT\_All and iGPT\_Middle\_4 exhibit comparably high performance, with only a narrow difference across all evaluation metrics. For instance, for the Cityscapes dataset, iGPT\_Middle\_4 surpasses iGPT\_All with marginal improvements of 1.05 in mIoU, 0.79 in mF1, and 0.9 in mPrecision. However, it falls short by 0.38 in mRecall. In the case of the CamVid dataset, iGPT\_Middle\_4 again outperforms iGPT\_All with increments of 0.31 in mIoU, 1.39 in mF1, and 0.4 in mPrecision, yet it lags by 0.2 in mRecall. Based on these observations, the subsequent experiments adopt iGPT\_Middle\_4 for further exploration, which strikes a trade-off between predictive performance and computational resources for street scene semantic understanding task within the realm of AD. 

\begin{figure}[tp]
\vspace{-0.3cm}
\hspace{-0.2cm}
\subfloat[Cityscapes]{\includegraphics[width=0.5\linewidth,height=0.35\linewidth]{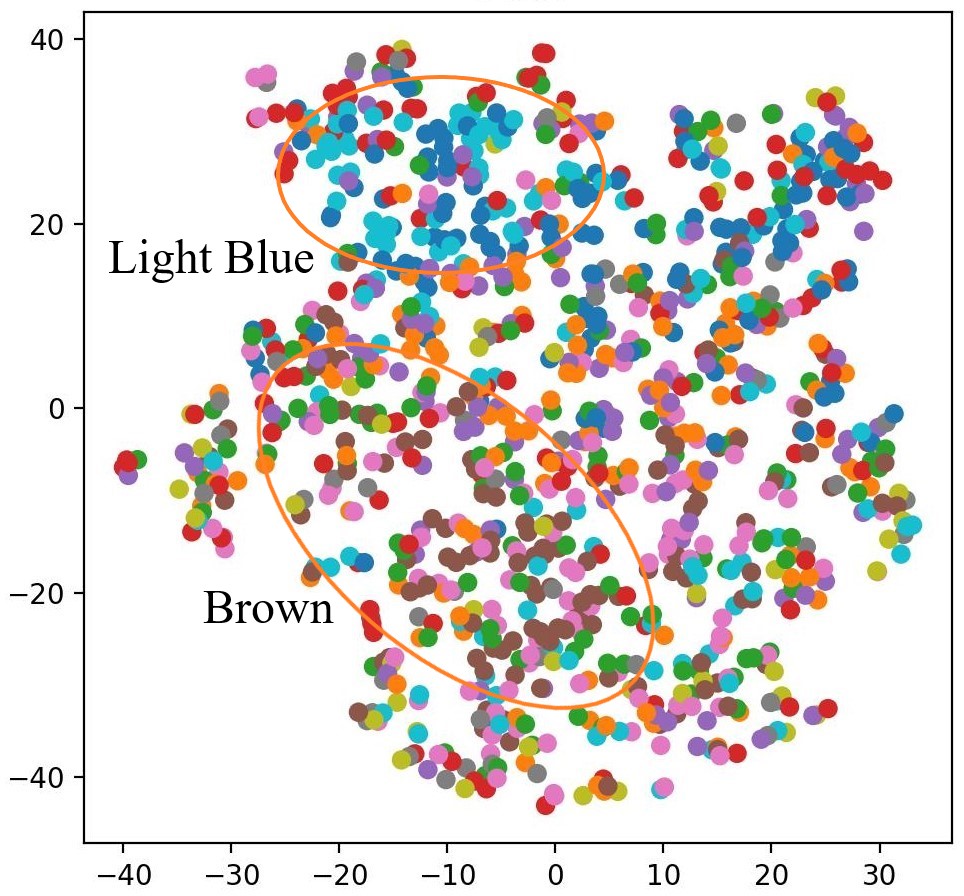}
\label{Fig:iGPT_variants_tsne_f}}
\subfloat[CamVid]{\includegraphics[width=0.5\linewidth,height=0.35\linewidth]{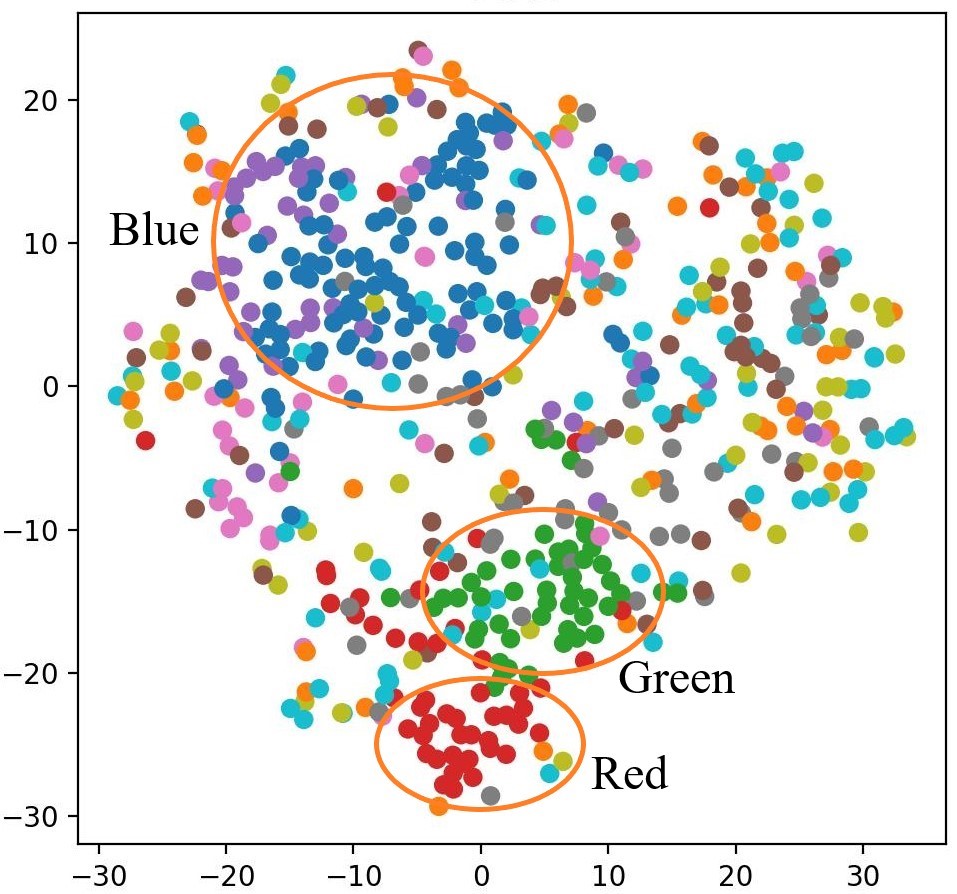}
\label{Fig:iGPT_variants_tsne_c}}
\caption{t-SNE visualization of hidden features of iGPT\_Middle\_4 on (a) Cityscapes test dataset and (b) CamVid test dataset. Colors represent semantic classes.}
\label{Fig:iGPT_tsne}
\vspace{0.1cm}
\end{figure}

\Cref{Fig:iGPT_tsne} presents a t-SNE visualization \cite{van2008visualizing} of the output of  iGPT\_Middle\_4 for Cityscapes test dataset and CamVid test dataset, respectively. This visualization highlights that certain semantic classes have been distinctly separated. Others, despite being intertwined, still exhibit a sufficient spread among data points. More specifically, \Cref{Fig:iGPT_variants_tsne_f} offers a visualization of the Cityscapes dataset, revealing that classes depicted in blue, light blue and brown tend to cluster together, while other classes are more dispersed, yet still maintain considerable separation. In a similar way, \Cref{Fig:iGPT_variants_tsne_c} for the CamVid dataset shows that classes identified by blue, green, and red are well separated from others. The remaining classes are not grouped closer together, yet they are sufficiently spaced apart to allow for distinct categorization.

\subsubsection{The proposed iGPT+Head vs Existing SOTA Models}
\label{iGPT_heads_vs_benchmarks}
Building upon our previous discussions, it was found that middle four layers (termed as iGPT\_Middle\_4) provides the best choice for street semantic understanding task in the context of AD. In this experiment, our goal is to examine whether iGPT+Head could outperform existing state-of-the-art models. Specifically, we will compare the performance of iGPT+Head with existing models: BiSeNetV2 \cite{yu2021bisenet} and SegNet \cite{badrinarayanan2017segnet}. 

\begin{figure*}[!t]
\centering
\subfloat[mIoU]{\includegraphics[width=0.25\linewidth,height=0.15\linewidth]{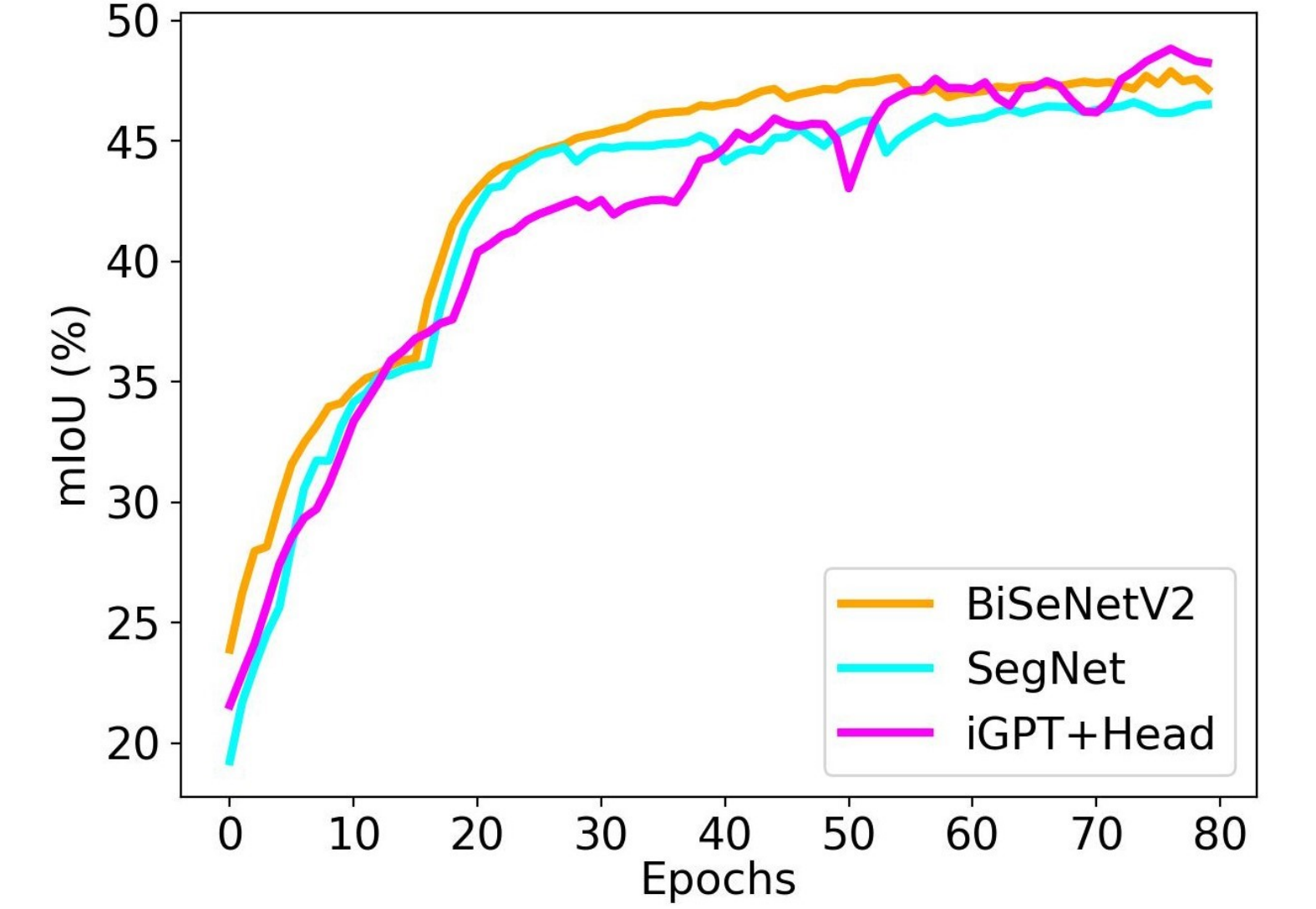}%
\label{Fig:iGPT_tradi_a}}
\subfloat[mPrecision]{\includegraphics[width=0.25\linewidth,height=0.15\linewidth]{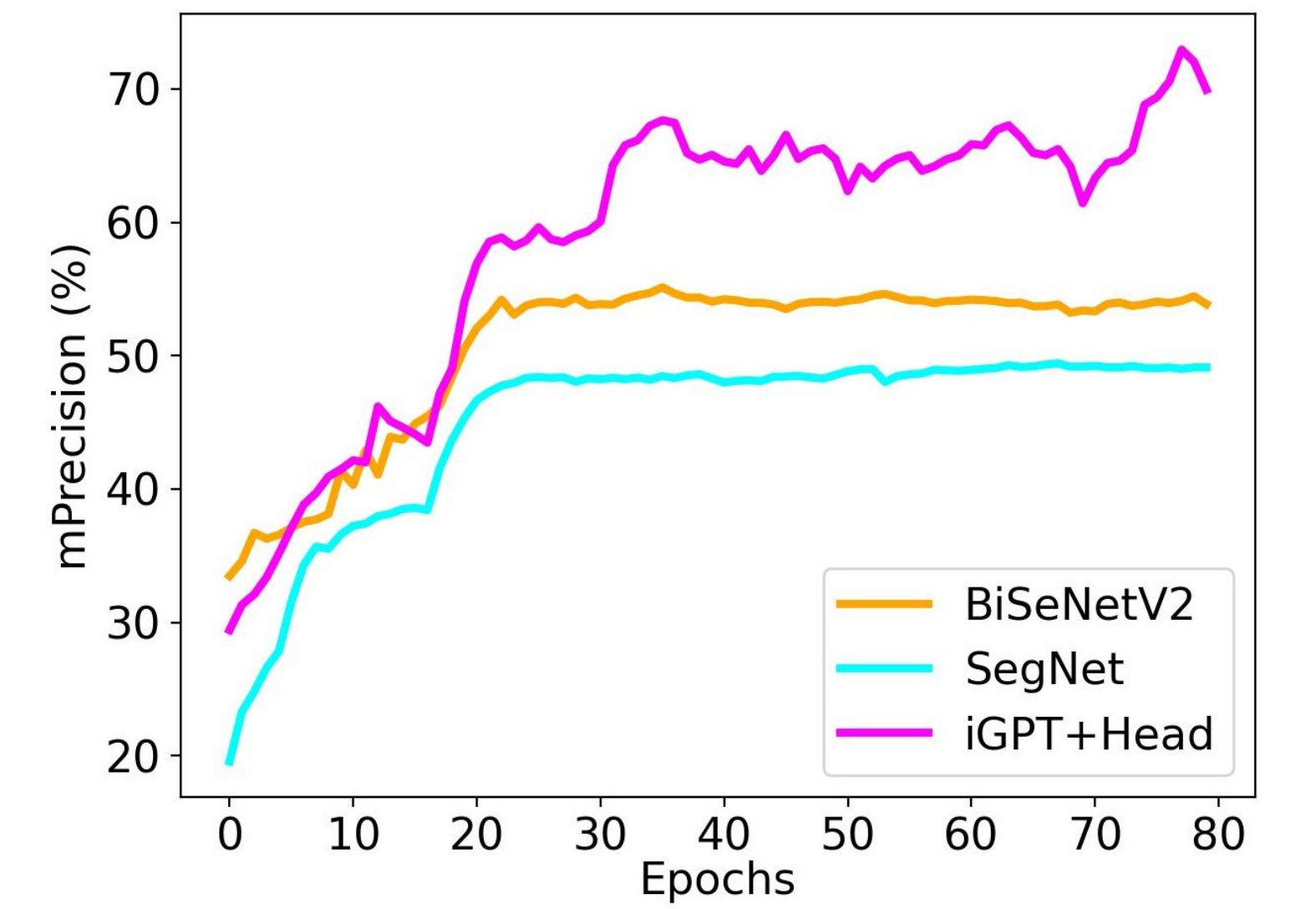}%
\label{Fig:iGPT_tradi_b}}
\subfloat[mRecall]{\includegraphics[width=0.25\linewidth,height=0.15\linewidth]{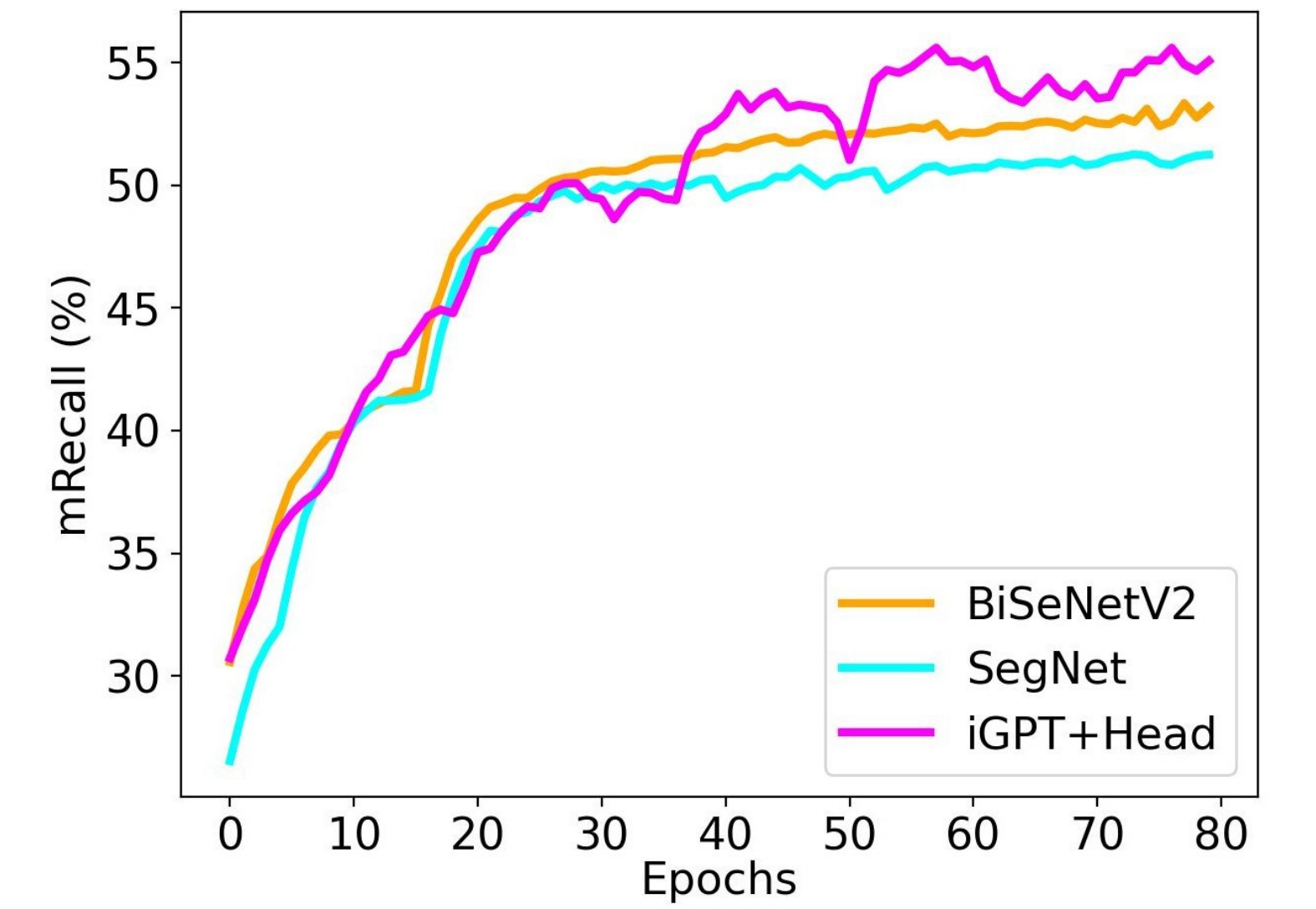}%
\label{Fig:iGPT_tradi_c}}
\subfloat[mF1]{\includegraphics[width=0.25\linewidth,height=0.15\linewidth]{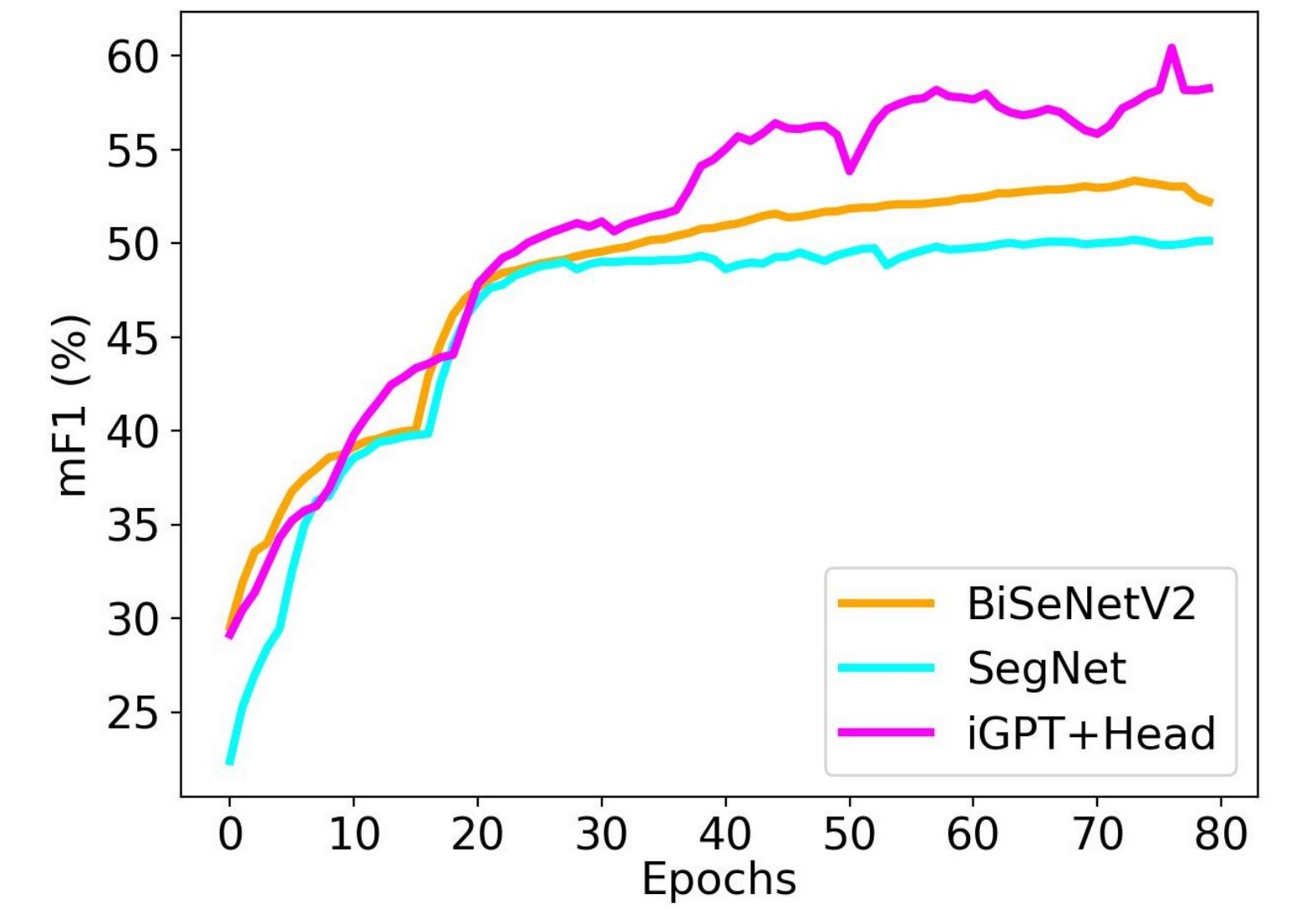}%
\label{Fig:iGPT_tradi_d}}
\caption{Performance of iGPT+Head against existing SOTA benchmarks on CamVid dataset.}
\label{Fig:iGPT_tradi}
\vspace{-0.2cm}
\end{figure*}

\Cref{Fig:iGPT_tradi_a,Fig:iGPT_tradi_b,Fig:iGPT_tradi_c,Fig:iGPT_tradi_d} present the performance across various metrics on CamVid dataset. There are two obvious observations: I) iGPT combined with SSeg head outperforms the current leading models BiSeNetV2 and SegNet in almost all metrics. II) iGPT+Head always converge faster than the compared SOTA benchmarks. This is because the pretrained LVM backbone has learned rich representations of various raw natural images.  
\Cref{Tab:iGPT_tradi_classes} compares the performance between iGPT+Head and its competitors quantitatively. The table indicates the following clear patterns: I) For classes that contain objects with large sizes, such as Sky, Building, Road, Sidewalk, Tree, and Car, almost all models demonstrate high performance, often exceeding 90\% across various metrics. II) The performance drops significantly for classes with narrow objects like Pole and Fence. For example, the models consistently show lower scores, with metrics for the Pole class approaching zero for all models. III) For the classes characterized by a high degree of shape variability, like Bicyclist, iGPT+Head surpasses the performance of BiSeNetV2 and SegNet. This better performance may be due to iGPT's exposure to a more diverse dataset during training, which likely includes a wide range of Bicyclist gestures, in contrast to the more constrained CamVid dataset used for BiSeNetV2 and SegNet. These findings highlight the importance of the scale and diversity of training data for iGPT, which contribute significantly to its robust generalization capabilities, particularly for complex classes with variable shapes.

\begin{table*}[tp]
\setlength{\tabcolsep}{6.0pt}
\caption{Class-wise inference performance comparison of all considered models on CamVid dataset}
\begin{tabularx}{\linewidth}{ccccccccccccc}
\hline
\multirow{2}{*}{Metrics}                   & \multirow{2}{*}{Models}      & \multicolumn{11}{c}{CamVid dataset (11 Semantic Classes) (\%)}                                                                     \\ \cline{3-13} 
                                          &                                  & Sky   & Building & Pole  & Road  & Sidewalk & Tree  & Signsymbol & Fence & Car   & Pedestrian & Bicyclist \\ \hline
\multicolumn{1}{c|}{\multirow{4}{*}{IoU}} & \multicolumn{1}{c|}{BiSeNetV2 \cite{yu2021bisenet}}   & 92.78 & 84.60    & 0.00 & 96.37 & 84.67    & 79.58 & 19.54      & 0.00  & 81.20 & 0.00      & 0.00      \\
\multicolumn{1}{c|}{}                     & \multicolumn{1}{c|}{SegNet \cite{badrinarayanan2017segnet}}      & 93.93 & 83.09    & 0.00  & 95.98 & 82.60    & 79.15 & 0.00       & 0.00  & 79.18 & 0.00       & 0.00      \\
\multicolumn{1}{c|}{}                     & \multicolumn{1}{c|}{iGPT+Head} & 85.31 & 74.60    & 0.05  & 89.15 & 63.81    & 64.60 & 23.80      & 34.30 & 50.80 & 4.36       & 44.80     \\ \hline
\multicolumn{1}{c|}{\multirow{4}{*}{F1}}  & \multicolumn{1}{c|}{BiSeNetV2 \cite{yu2021bisenet}}   & 96.25 & 91.66    & 0.00 & 98.15 & 91.70    & 88.63 & 31.73      & 0.00  & 89.62 & 0.00      & 0.00      \\
\multicolumn{1}{c|}{}                     & \multicolumn{1}{c|}{SegNet \cite{badrinarayanan2017segnet}}      & 96.87 & 90.77    & 0.00  & 97.95 & 90.46    & 88.36 & 0.00       & 0.00  & 88.37 & 0.00       & 0.00      \\
\multicolumn{1}{c|}{}                     & \multicolumn{1}{c|}{iGPT+Head} & 92.07 & 85.43    & 0.09  & 94.27 & 77.91    & 78.47 & 38.44      & 51.03 & 67.37 & 8.15       & 61.83     \\ \hline
\multicolumn{1}{c|}{\multirow{4}{*}{Precision}} & \multicolumn{1}{c|}{BiSeNetV2 \cite{yu2021bisenet}}   & 96.36 & 88.32    & 0.00 & 97.73 & 92.96    & 87.81 & 55.22      & 0.00  & 87.99 & 0.00      & 0.00      \\
\multicolumn{1}{c|}{}                     & \multicolumn{1}{c|}{SegNet \cite{badrinarayanan2017segnet}}      & 97.11 & 86.45    & 0.00  & 97.58 & 92.49    & 87.46 & 0.00       & 0.00  & 88.42 & 0.00       & 0.00      \\
\multicolumn{1}{c|}{}                     & \multicolumn{1}{c|}{iGPT+Head} & 93.25 & 83.41    & 10.94 & 93.75 & 84.20    & 85.07 & 96.84      & 77.53 & 76.84 & 73.18      & 88.30     \\ \hline
\multicolumn{1}{c|}{\multirow{4}{*}{Recall}} & \multicolumn{1}{c|}{BiSeNetV2 \cite{yu2021bisenet}}   & 97.21 & 96.86    & 0.00 & 98.84 & 92.05    & 92.10 & 23.49      & 0.00  & 94.25 & 0.00      & 0.00      \\
\multicolumn{1}{c|}{}                     & \multicolumn{1}{c|}{SegNet \cite{badrinarayanan2017segnet}}      & 98.02 & 96.95    & 0.00  & 98.79 & 91.59    & 91.61 & 0.00       & 0.00  & 90.96 & 0.00       & 0.00      \\
\multicolumn{1}{c|}{}                     & \multicolumn{1}{c|}{iGPT+Head} & 94.04 & 93.76    & 0.05  & 97.29 & 79.18    & 80.35 & 25.82      & 47.41 & 71.80 & 4.75       & 52.08     \\ \hline 
\end{tabularx}
\label{Tab:iGPT_tradi_classes}
\vspace{-0.5cm}
\end{table*}

\begin{figure*}[!t]
\centering
\subfloat[mIoU on Cityscapes]{\includegraphics[width=0.25\linewidth,height=0.15\linewidth]{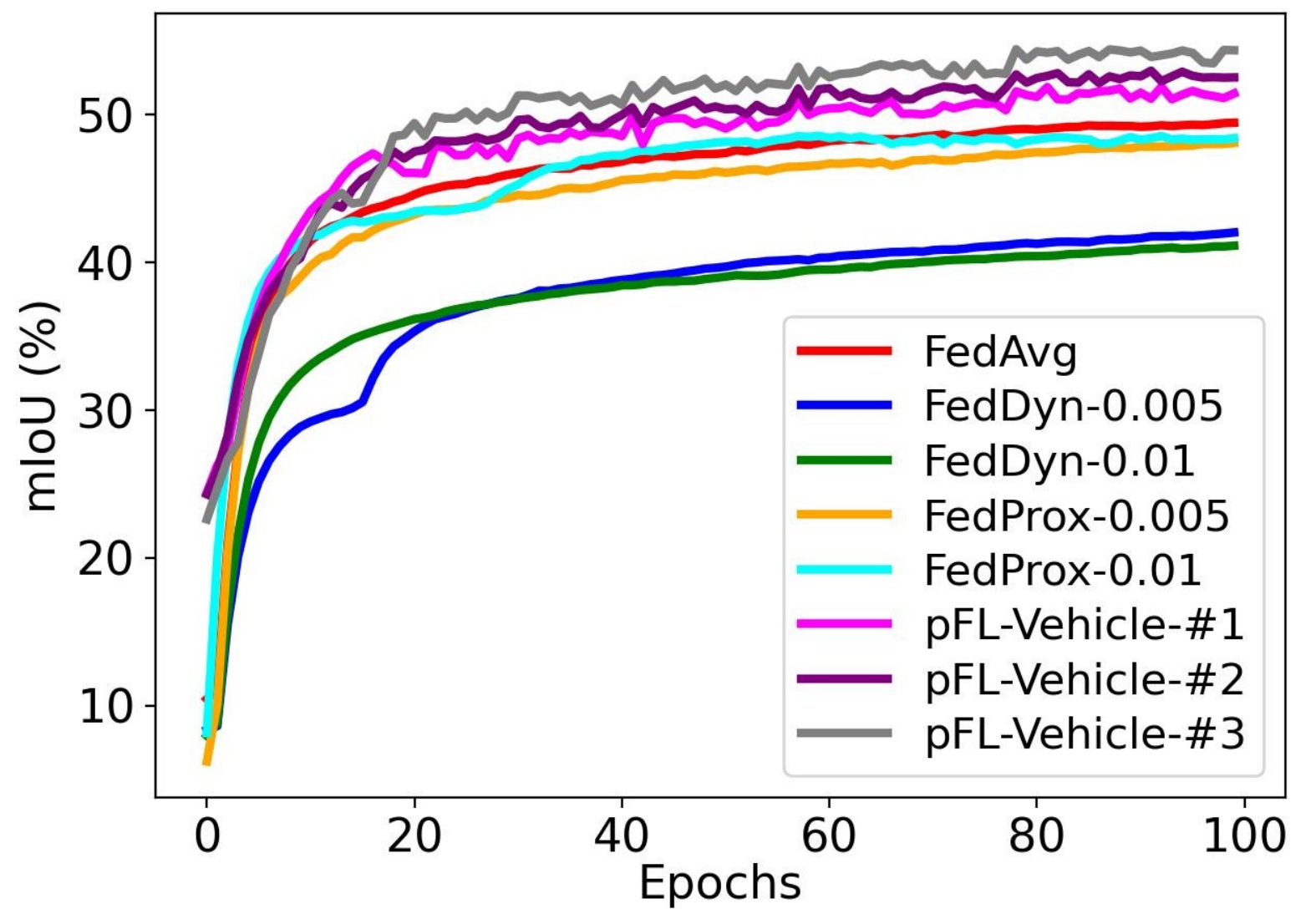}
\label{Fig:feat_pFL_e}}
\subfloat[mPrecision on Cityscapes]{\includegraphics[width=0.25\linewidth,height=0.15\linewidth]{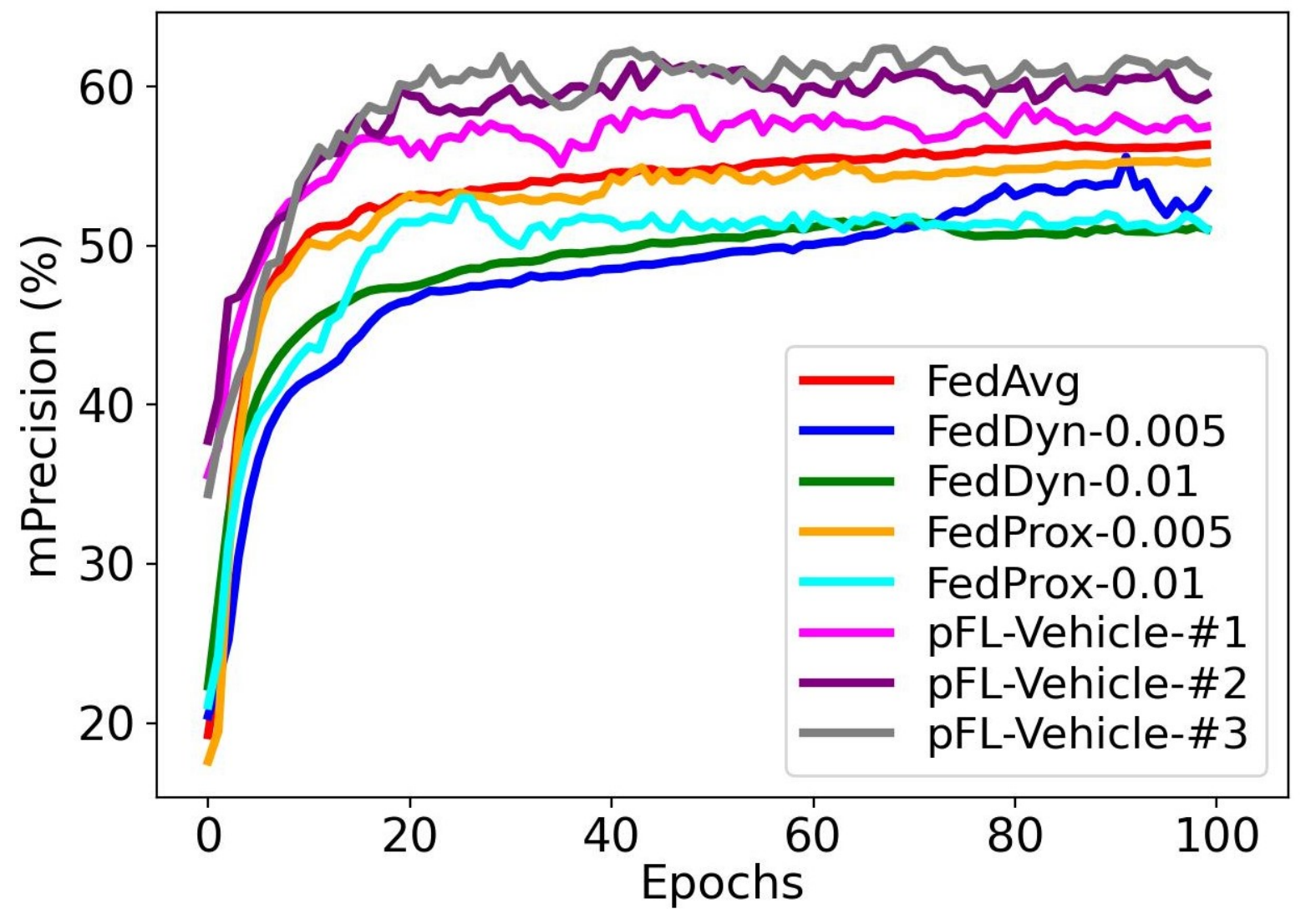}%
\label{Fig:feat_pFL_f}}
\subfloat[mRecall on Cityscapes]{\includegraphics[width=0.25\linewidth,height=0.15\linewidth]{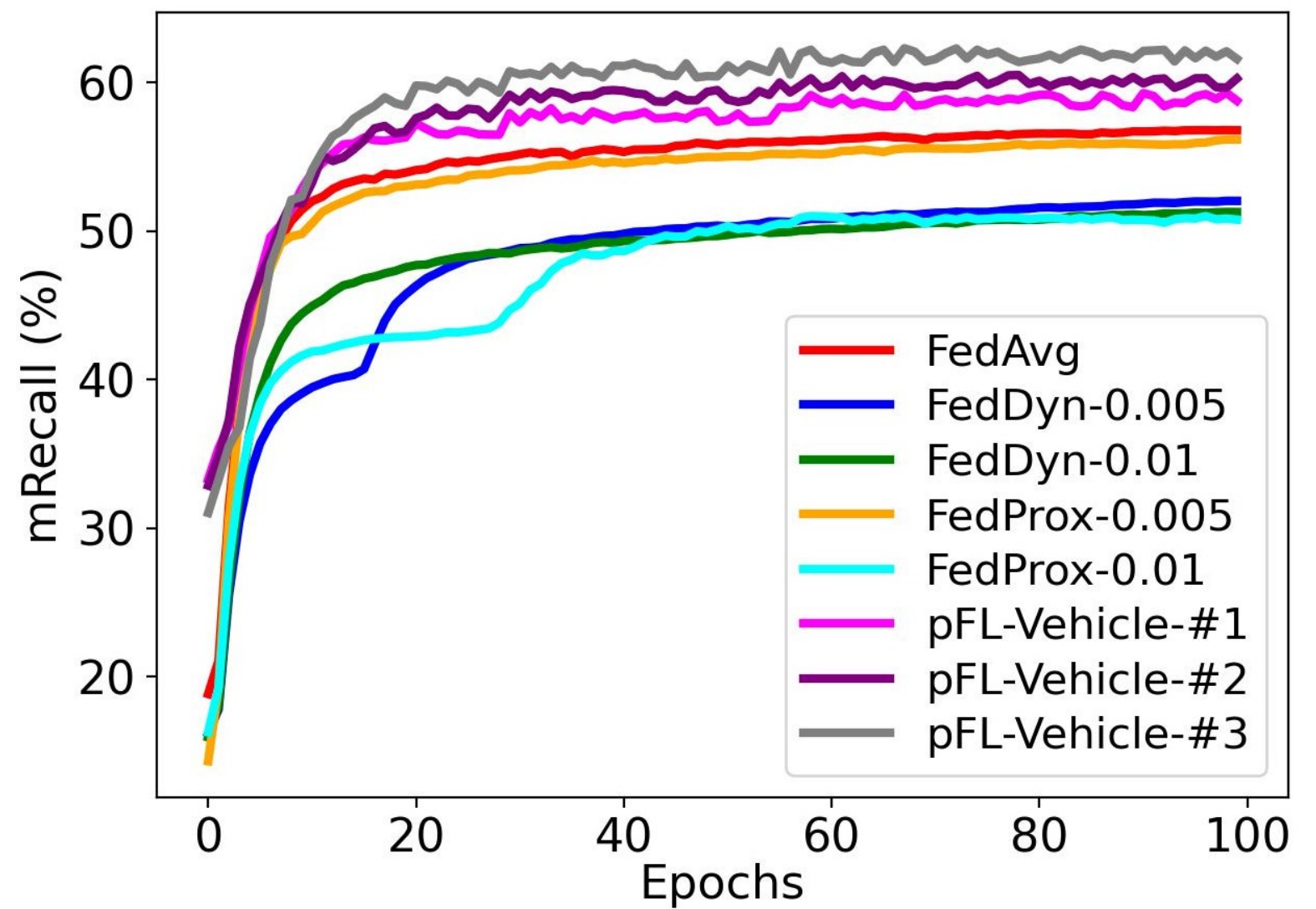}%
\label{Fig:feat_pFL_g}}
\subfloat[mF1 on Cityscapes]{\includegraphics[width=0.25\linewidth,height=0.15\linewidth]{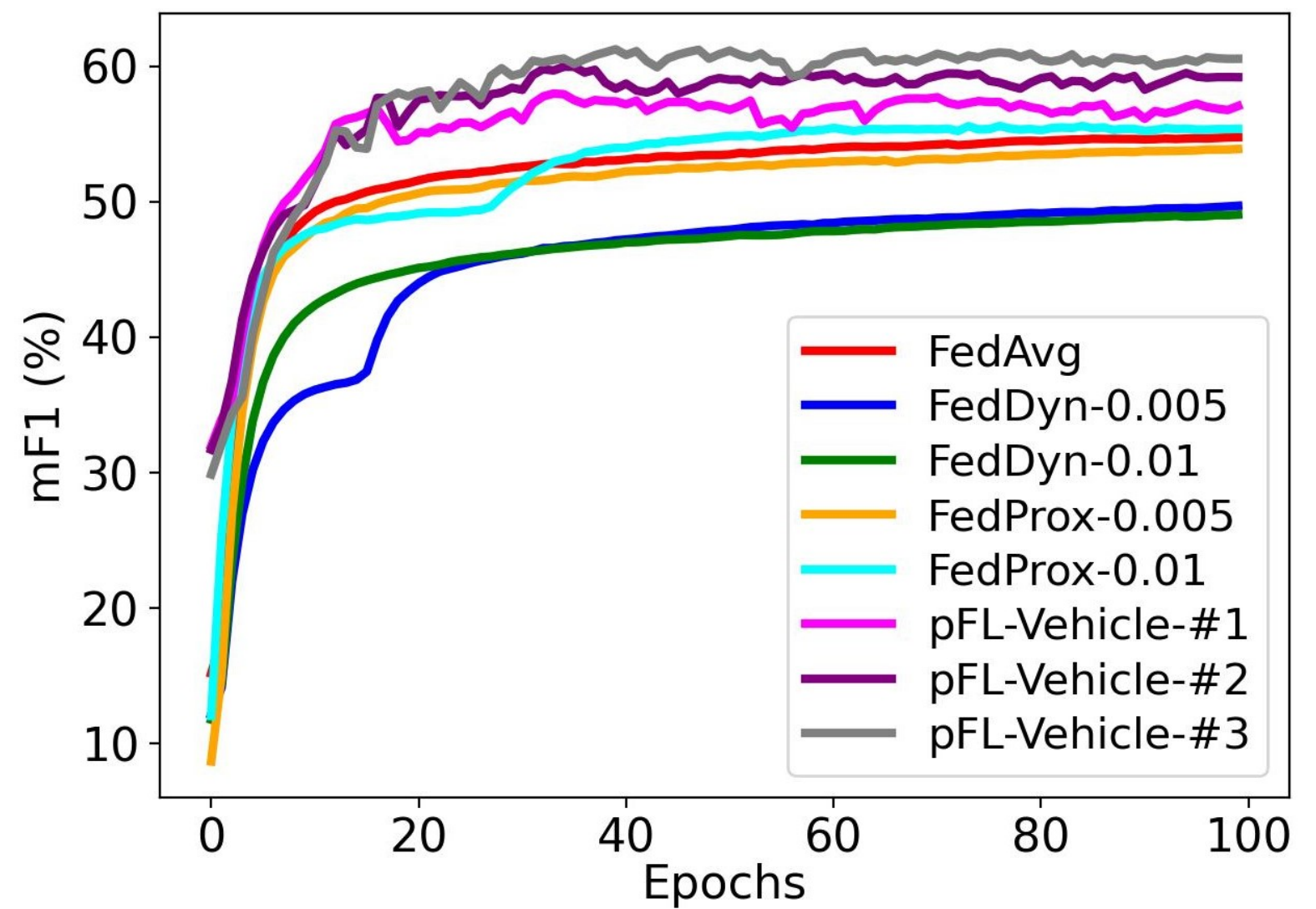}
\label{Fig:feat_pFL_h}}

\subfloat[mIoU on CamVid]{\includegraphics[width=0.25\linewidth,height=0.15\linewidth]{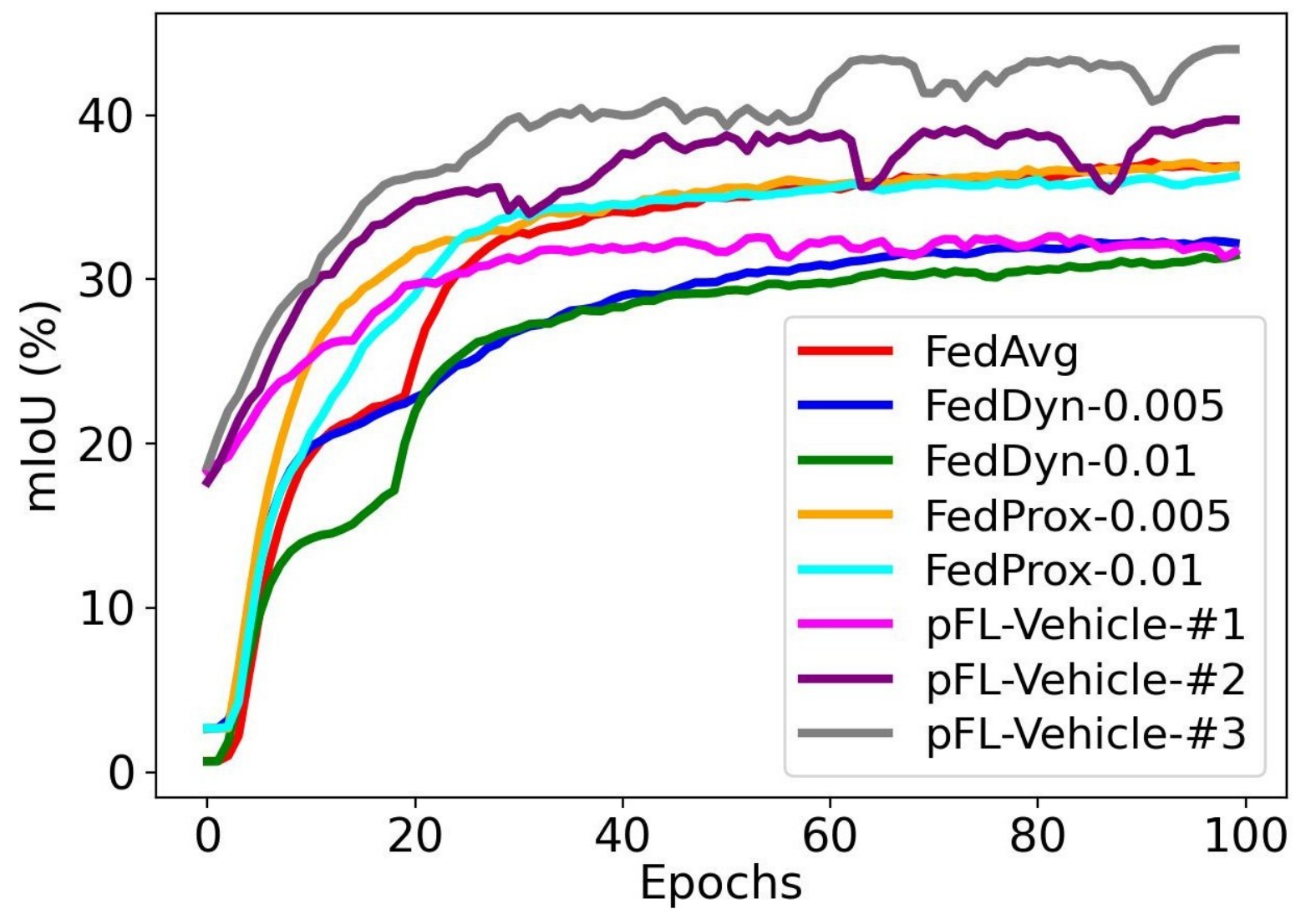}
\label{Fig:feat_pFL_a}}
\subfloat[mPrecision on CamVid]{\includegraphics[width=0.25\linewidth,height=0.15\linewidth]{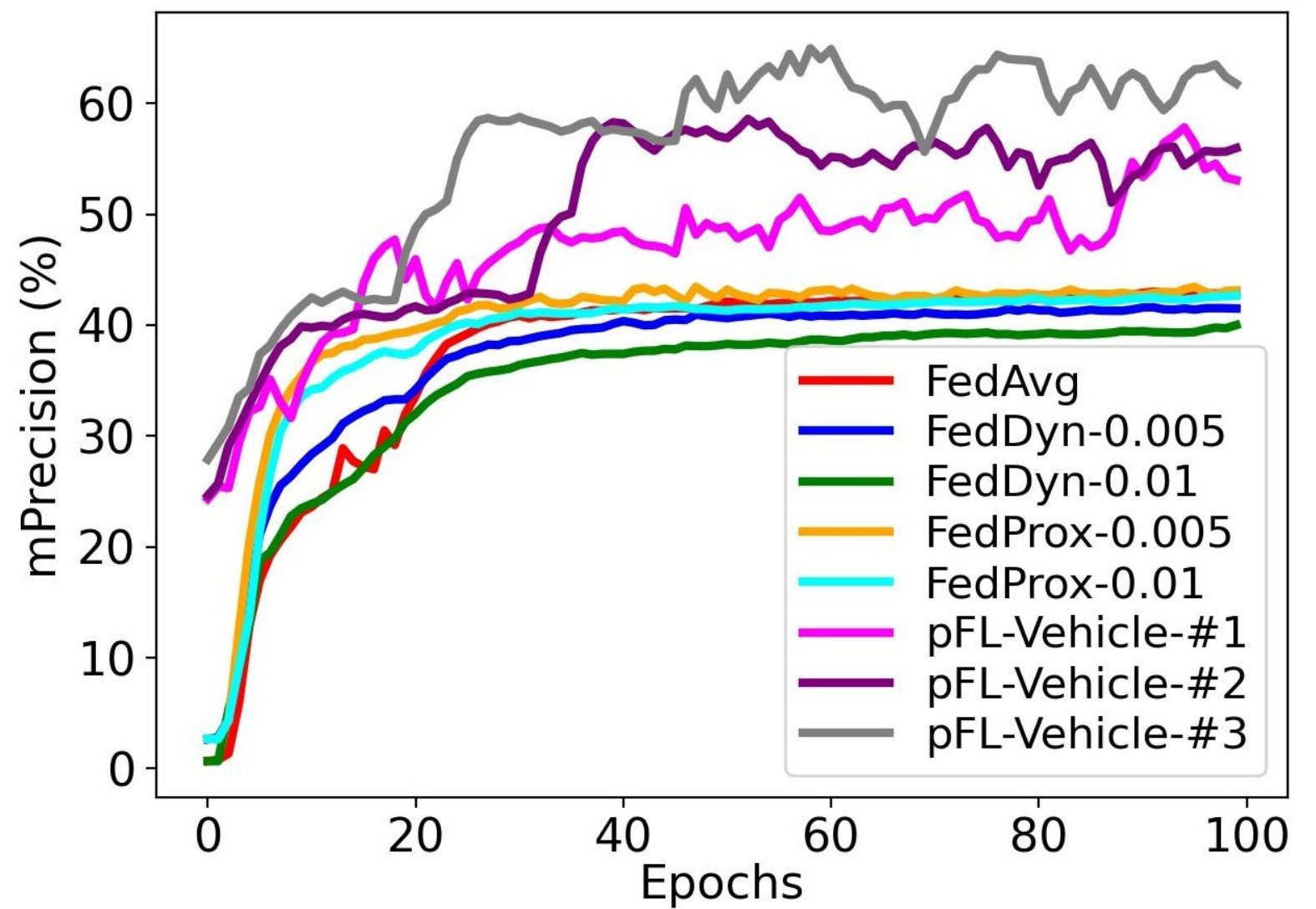}
\label{Fig:feat_pFL_b}}
\subfloat[mRecall on CamVid]{\includegraphics[width=0.243\linewidth,height=0.15\linewidth]{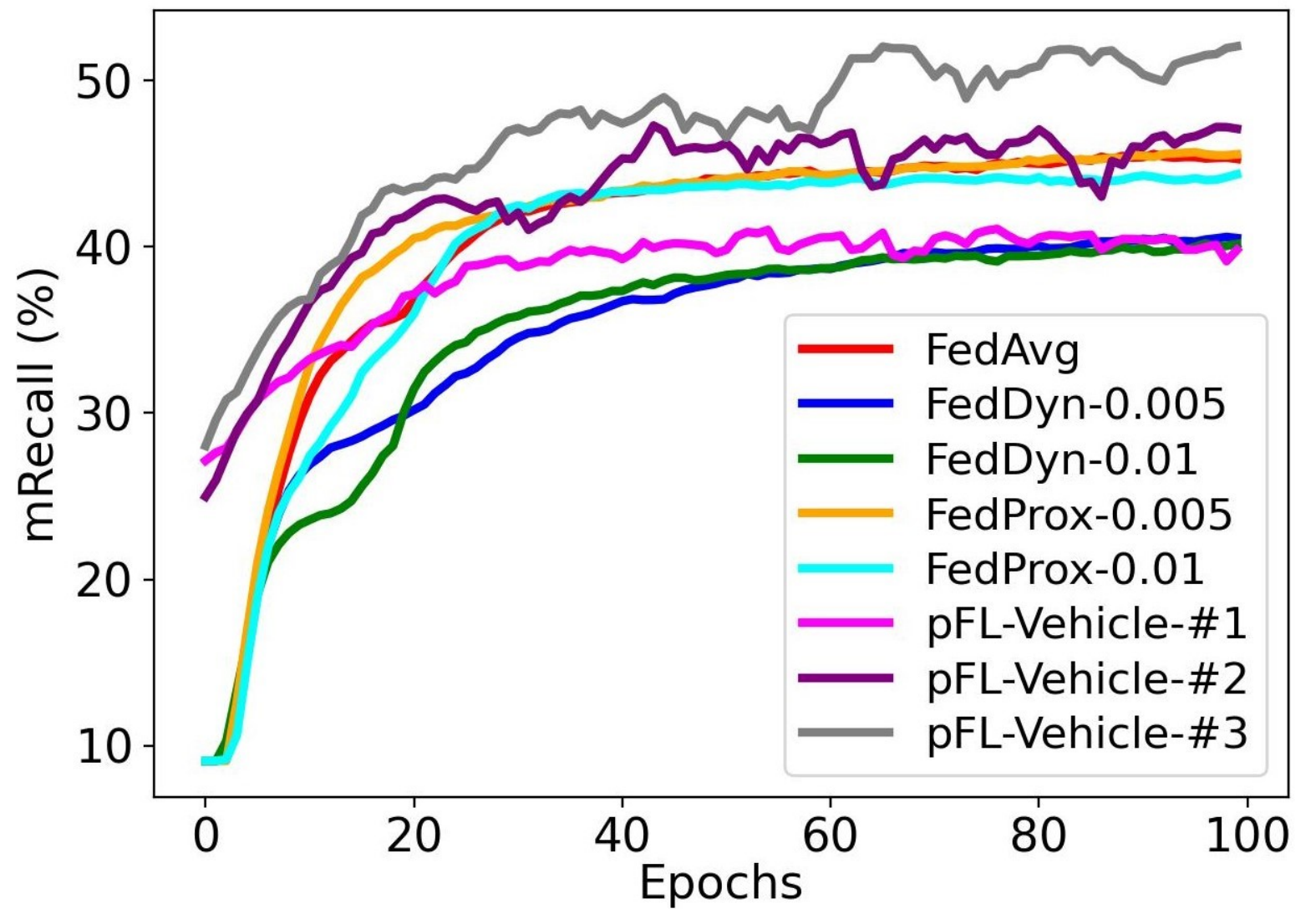}
\label{Fig:feat_pFL_c}}
\subfloat[mF1 on CamVid]{\includegraphics[width=0.243\linewidth,height=0.15\linewidth]{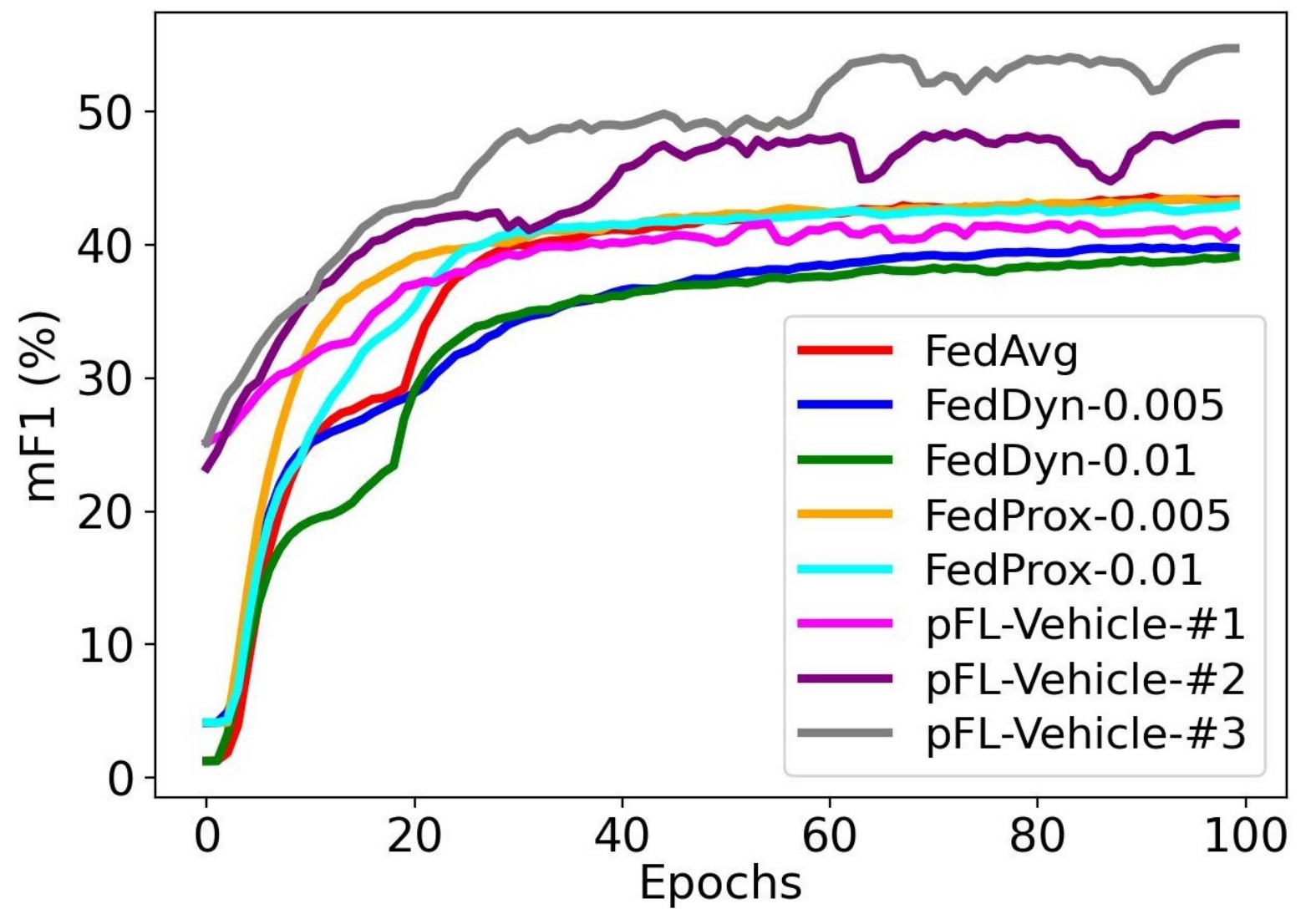}
\label{Fig:feat_pFL_d}}
\caption{Performance comparison of the proposed pFedLVM against other FL algorithms on Cityscapes and CamVid datasets.}
\label{Fig:feat_pFL}
\vspace{-0.5cm}
\end{figure*}

\subsection{Evaluation of the Proposed LVM-Driven Feature-based pFL (pFedLVM)}
\subsubsection{Performance Comparison between pFedLVM algorithm and existing SOTA FL Algorithms}
In this experiment, we compare the proposed feature-based personalized Federated Learning (pFL) approach with other FL algorithms: FedAvg \cite{mcmahan2017communication}, FedProx \cite{li2020federated}, and FedDyn \cite{acar2021federated}. Both FedDyn and FedProx include a hyperparameter that requires careful tuning. In addition, the SegNet model has been verified to achieve comparable performance to iGPT+Head on both Cityscapes and CamVid datasets, as detailed in the previous subsection. Therefore, the SegNet model is employed as the underlying architecture for all FL algorithms under consideration in this study. The hyperparameters for FedDyn and FedProx are set to 0.005 or 0.01. The notations FedDyn-0.005, FedDyn-0.01, FedProx-0.005, and FedProx-0.01 correspond to these models with the specified hyperparameters. Additionally, pFL-Vehicle-\#1, pFL-Vehicle-\#2, and pFL-Vehicle-\#3 represent the personalized models from the proposed framework for Vehicles $1$, $2$, and $3$, respectively.

\Cref{Fig:feat_pFL_e,Fig:feat_pFL_f,Fig:feat_pFL_g,Fig:feat_pFL_h} depict the inference performance of all considered models across all evaluation metrics on Cityscapes dataset. The results are clear: I) The personalized model tailored for Vehicle 1, Vehicle 2, and Vehicle 3 demonstrates better performance when compared to other benchmarks, underscoring the efficacy of the feature-based personalized learning approach. II) At the beginning, the metric scores for Vehicle 1, 2, and 3 are higher compared to the other benchmarks. This initial advantage for Vehicles 1, 2, and 3 is due to the utilization of a pretrained iGPT for feature extraction, in contrast to the other models that begin their training from scratch. III) Although Vehicle $1$, $2$, and $3$ surpass other benchmarks in overall performance, there are noticeable differences in their performance, with each vehicle showing a distinct accuracy. This variation in performance is attributed to the different sizes of datasets used for each vehicle (shown in \Cref{Tab:number_for_vehicles}). This suggests that larger datasets typically provide more comprehensive training, leading to better model performance, whereas smaller datasets may limit a model's ability to learn and generalize, resulting in a relatively poor performance. IV) The quantitative improvements of the pFL-Vehicle-\#3 compared to FedAvg are 10.05\%, 11.85\%, 10.71\% and 9.70\% in terms of mIoU, mF1, mPrecision and mRecall, respectively. \Cref{Fig:feat_pFL_a,Fig:feat_pFL_b,Fig:feat_pFL_c,Fig:feat_pFL_d} show the corresponding results of CamVid dataset, and the same conclusion can be drawn as in the Cityscapes dataset. In particular, the quantitative improvements are 18.47\%, 25.60\%, 51.03\% and 14.19\% in terms of mIoU, mF1, mPrecision and mRecall, respectively.

It is important to highlight that within the feature-based pFL mechanism, the personalized feature compressor (with parameters $\omega_{v, c}$) and the personalized downstream head (with parameters $\omega_{v, p}$) collectively play a pivotal role in achieving superior performance in comparison to other baselines. The experiments conducted in this study, therefore, serve to validate the effectiveness of both the personalized feature compressor and the personalized downstream head in enhancing model performance.

\subsubsection{t-SNE Visualization of pFedLVM against other existing SOTA FL algorithms}
\begin{figure*}[!t]
\centering
\subfloat[FedAvg]{\includegraphics[width=0.3\linewidth,height=0.16\linewidth]{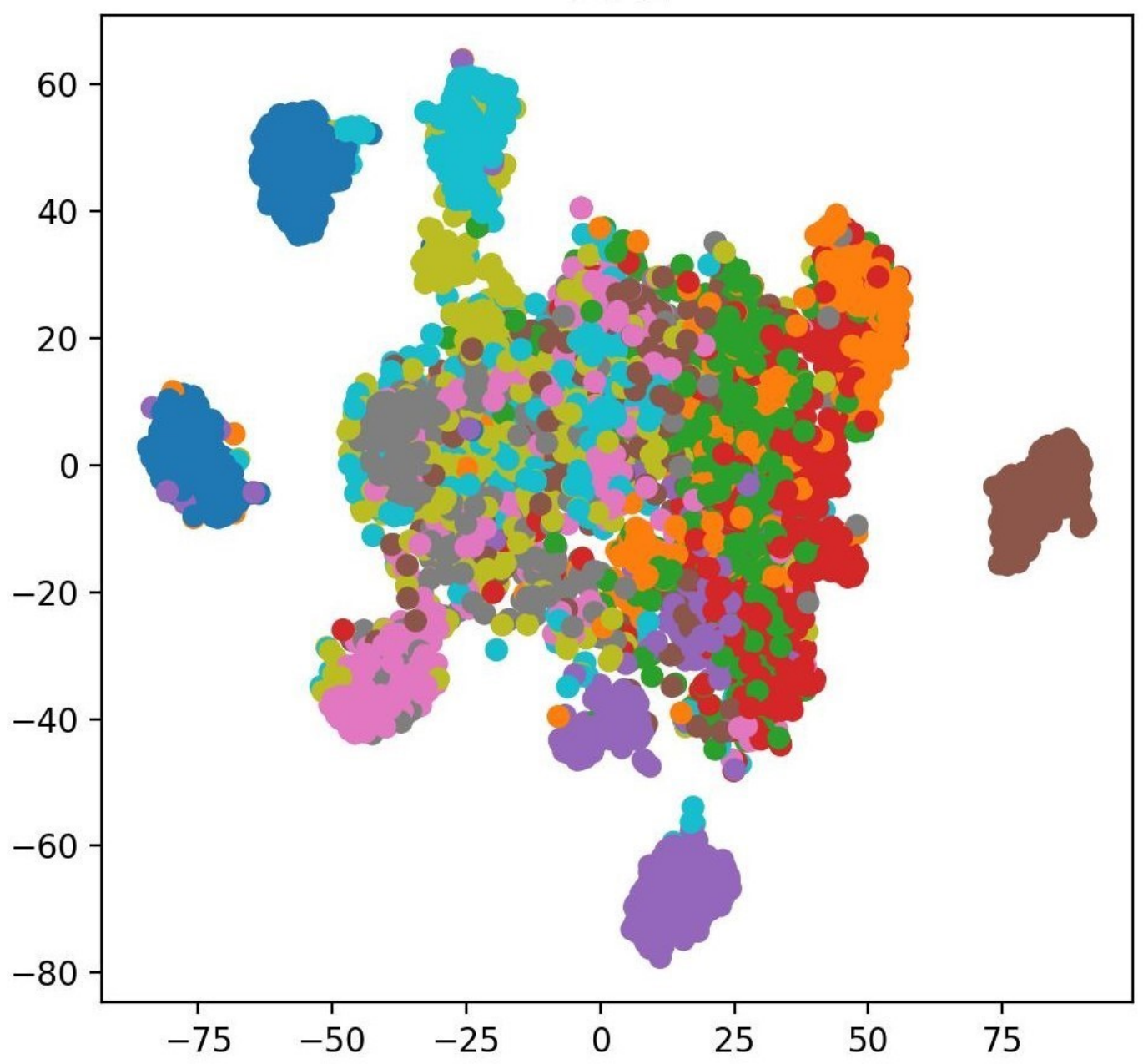}%
\label{Fig:pFL_visual_city_a}}
\subfloat[FedDyn-0.005]{\includegraphics[width=0.3\linewidth,height=0.16\linewidth]{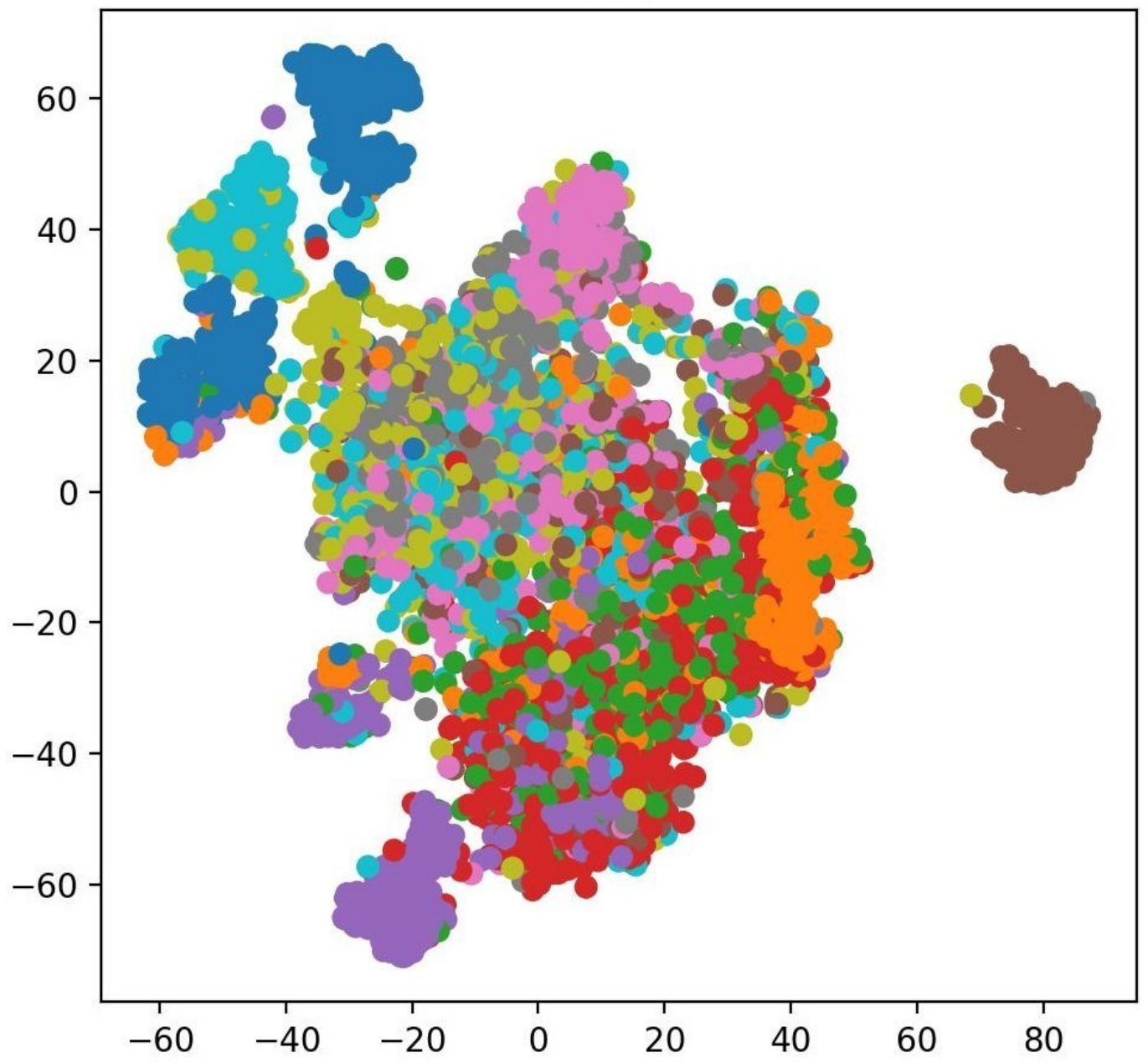}%
\label{Fig:pFL_visual_city_b}}
\subfloat[FedProx-0.005]{\includegraphics[width=0.3\linewidth,height=0.16\linewidth]{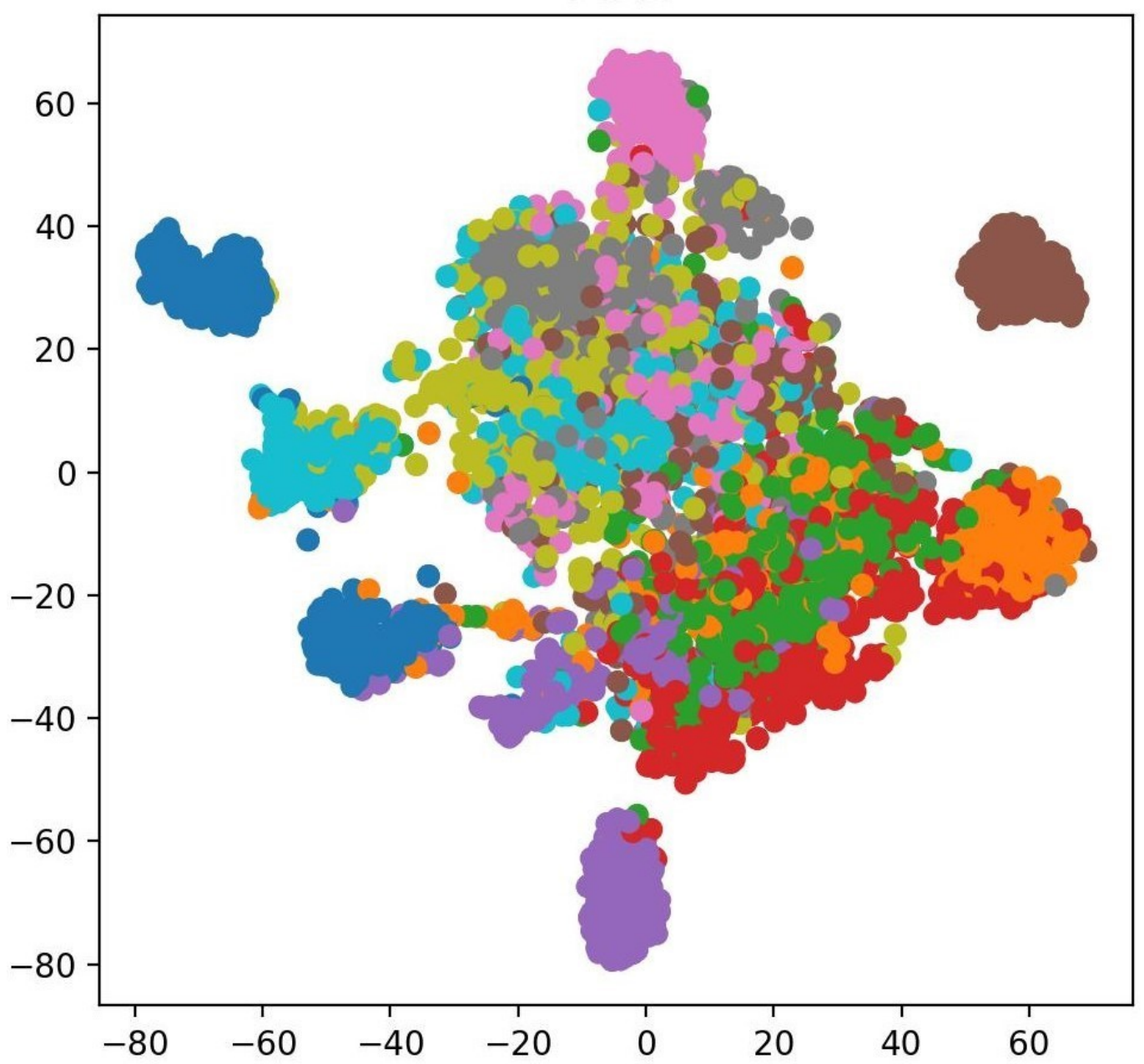}%
\label{Fig:pFL_visual_city_c}}

\subfloat[pFL-Vehicle-\#1]{\includegraphics[width=0.3\linewidth,height=0.16\linewidth]{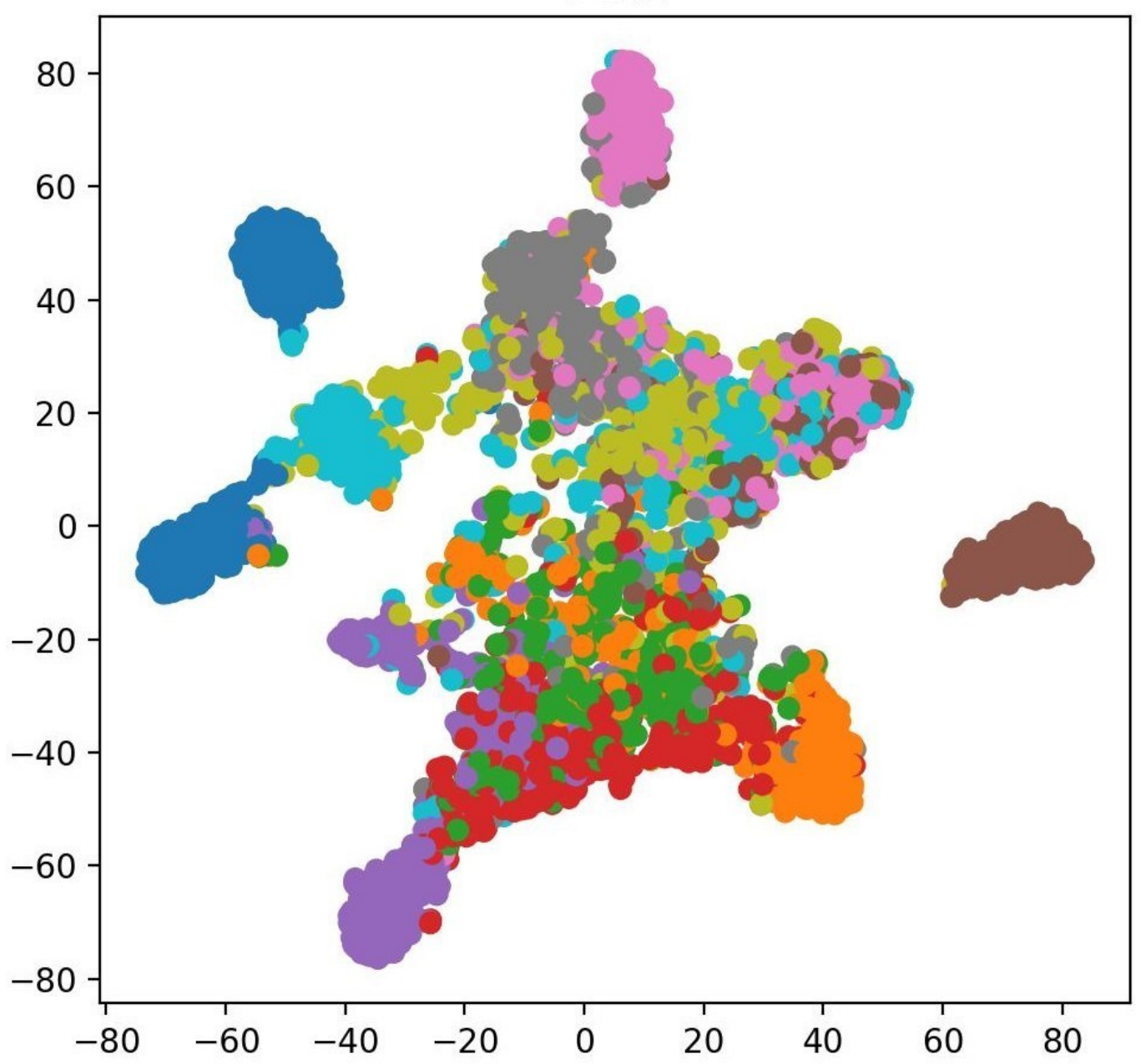}%
\label{Fig:pFL_visual_city_d}}
\subfloat[pFL-Vehicle-\#2]{\includegraphics[width=0.3\linewidth,height=0.16\linewidth]{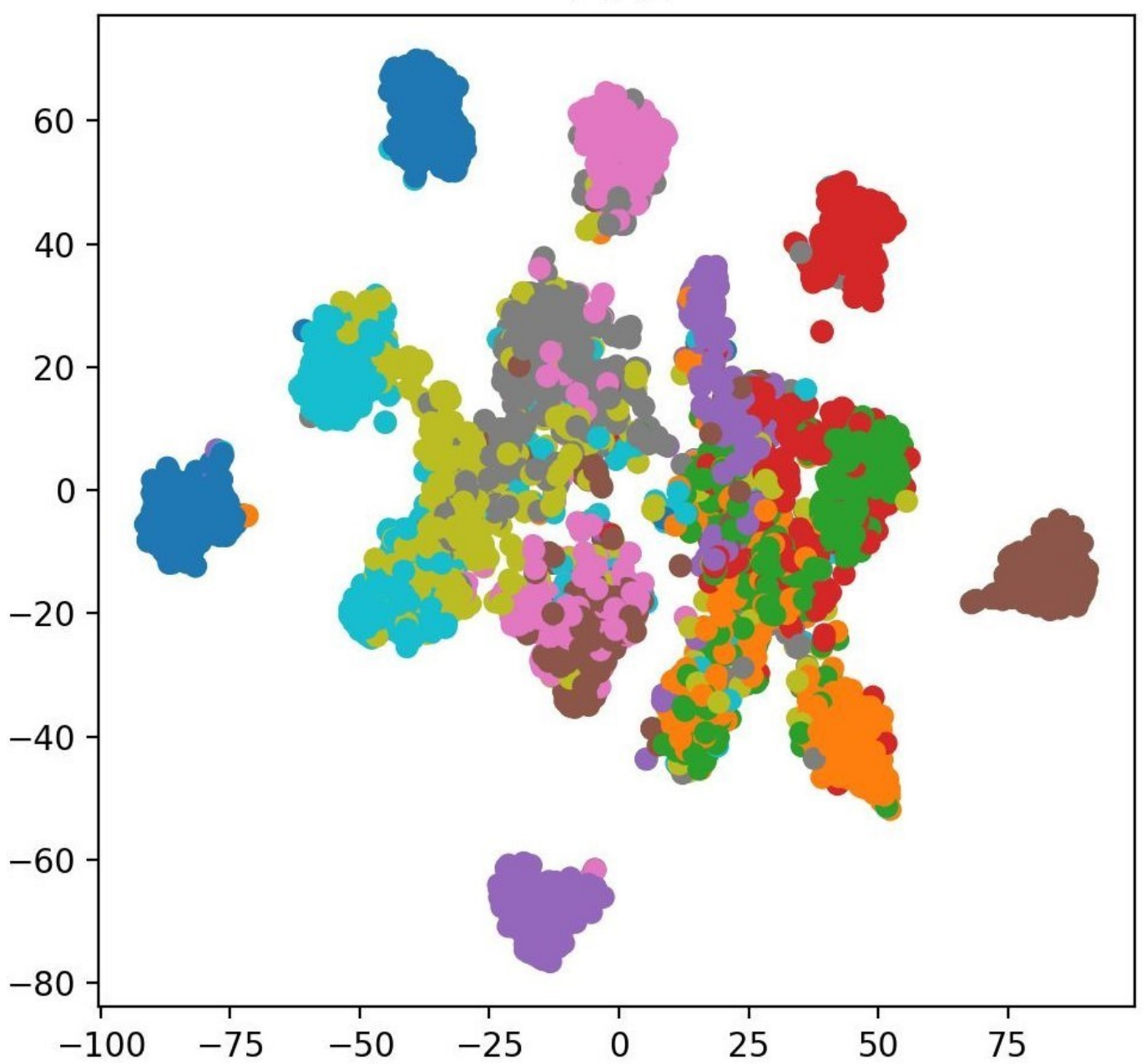}%
\label{Fig:pFL_visual_city_e}}
\subfloat[pFL-Vehicle-\#3]{\includegraphics[width=0.3\linewidth,height=0.16\linewidth]{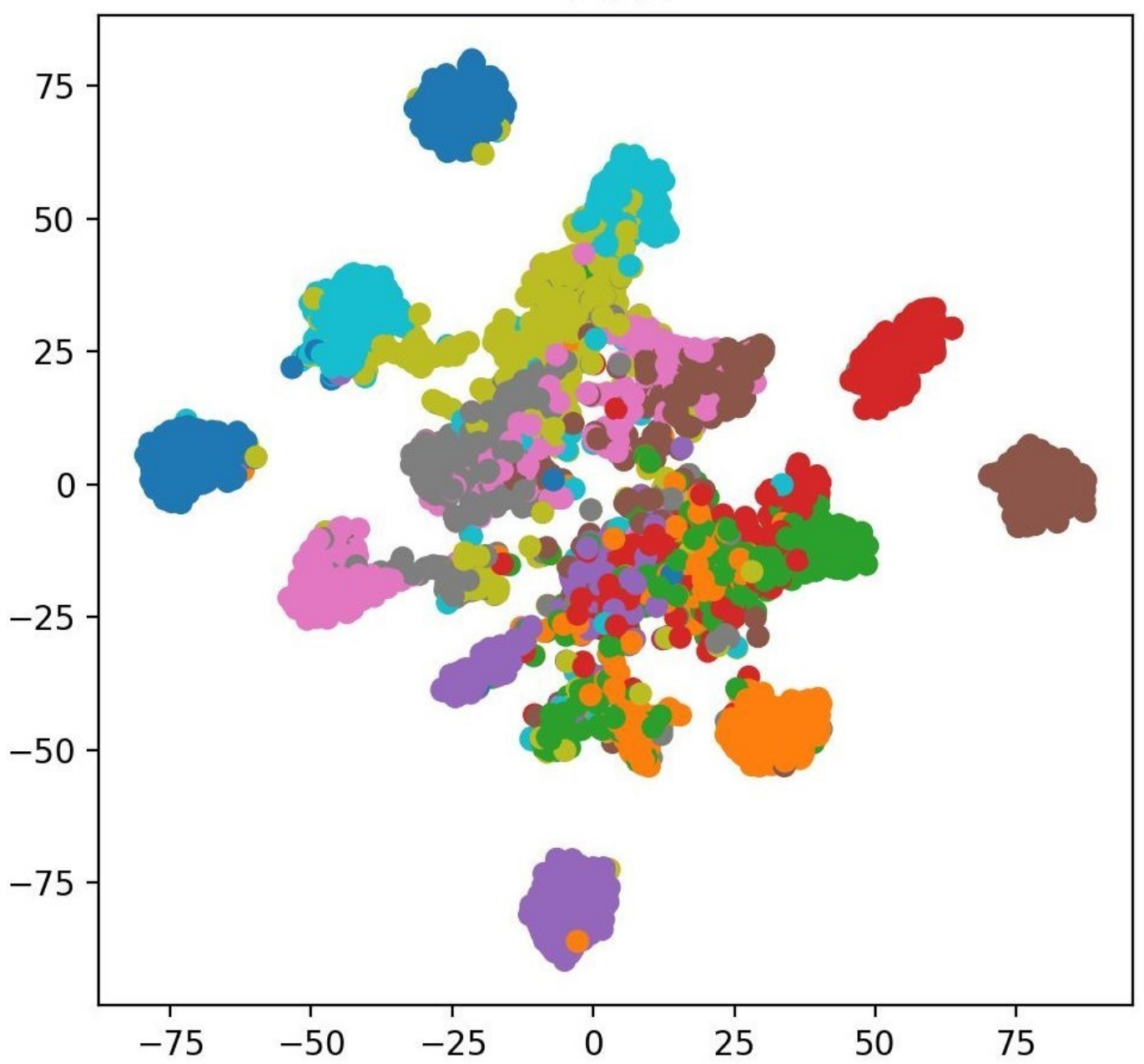}%
\label{Fig:pFL_visual_city_f}}
\caption{t-SNE visualization of pixel embedding of Cityscapes test dataset. Colors represent semantic classes.}
\label{Fig:pFL_visual_city}
\vspace{-0.4cm}
\end{figure*}

\Cref{Fig:pFL_visual_city} presents a t-SNE visualization \cite{van2008visualizing,miao2023fedseg} of the pixel embeddings for the Cityscapes test dataset, facilitating a comparative analysis of the models pFL-Vehicle-\#1, pFL-Vehicle-\#2, and pFL-Vehicle-\#3 against FedAvg, FedDyn-0.005, and FedProx-0.005. The visualization reveals that while FedAvg, FedDyn-0.005, and FedProx-0.005 models exhibit limited capability, distinguishing only some of the semantic classes with others remaining interspersed, the pFL-Vehicle-\#2 and pFL-Vehicle-\#3 models demonstrate a marked superiority in almost all the semantic classes. Conversely, pFL-Vehicle-\#1 lags somewhat behind in performance, a discrepancy that can be traced back to its smaller data volume relative to pFL-Vehicle-\#2 and pFL-Vehicle-\#3. These observations are in line with the performance metrics detailed in \Cref{Fig:feat_pFL_e,Fig:feat_pFL_f,Fig:feat_pFL_g,Fig:feat_pFL_h}.

\begin{figure*}[!t]
\centering
\subfloat[FedAvg]{\includegraphics[width=0.3\linewidth,height=0.16\linewidth]{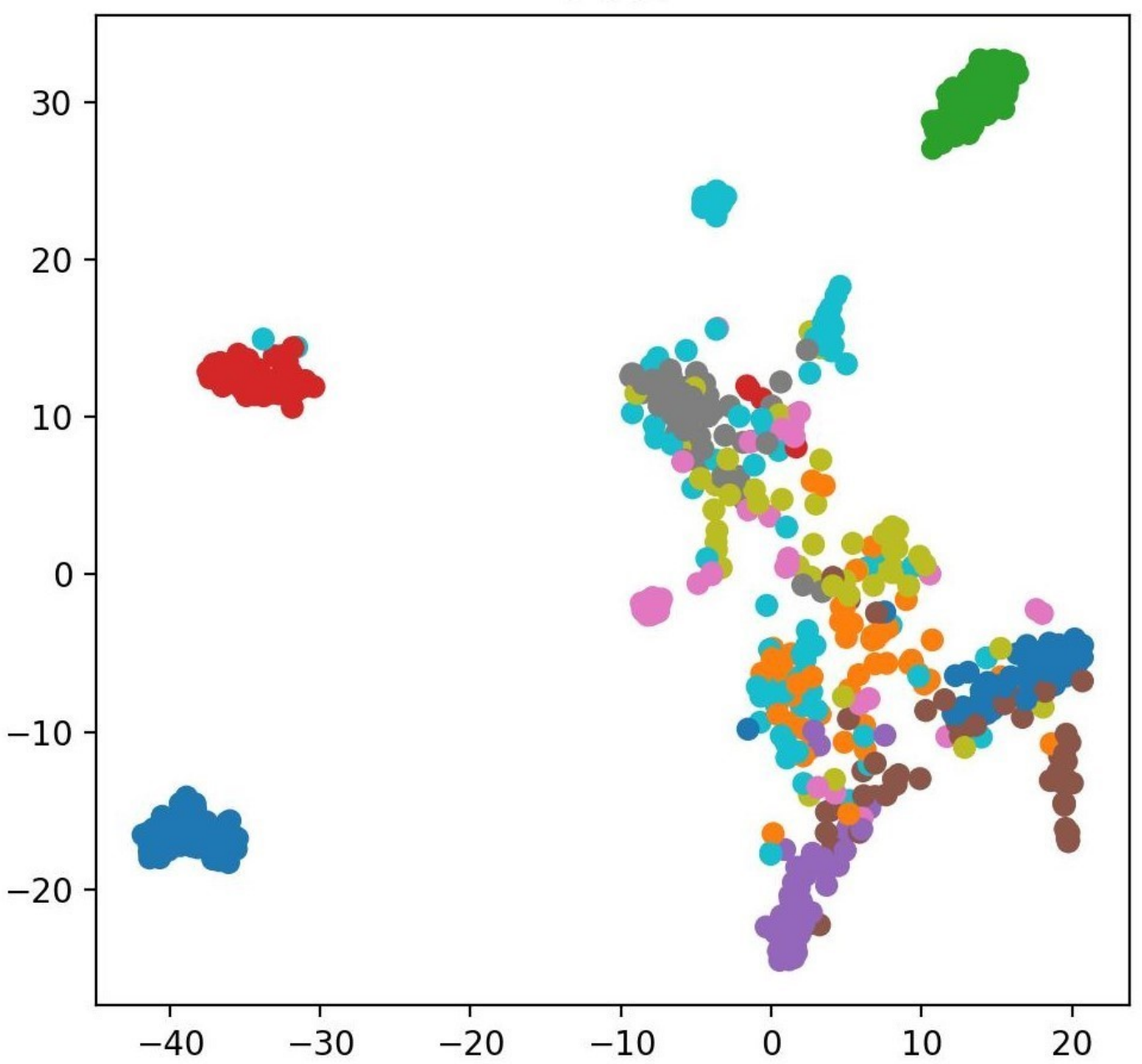}%
\label{Fig:pFL_visual_a}}
\subfloat[FedDyn-0.005]{\includegraphics[width=0.3\linewidth,height=0.16\linewidth]{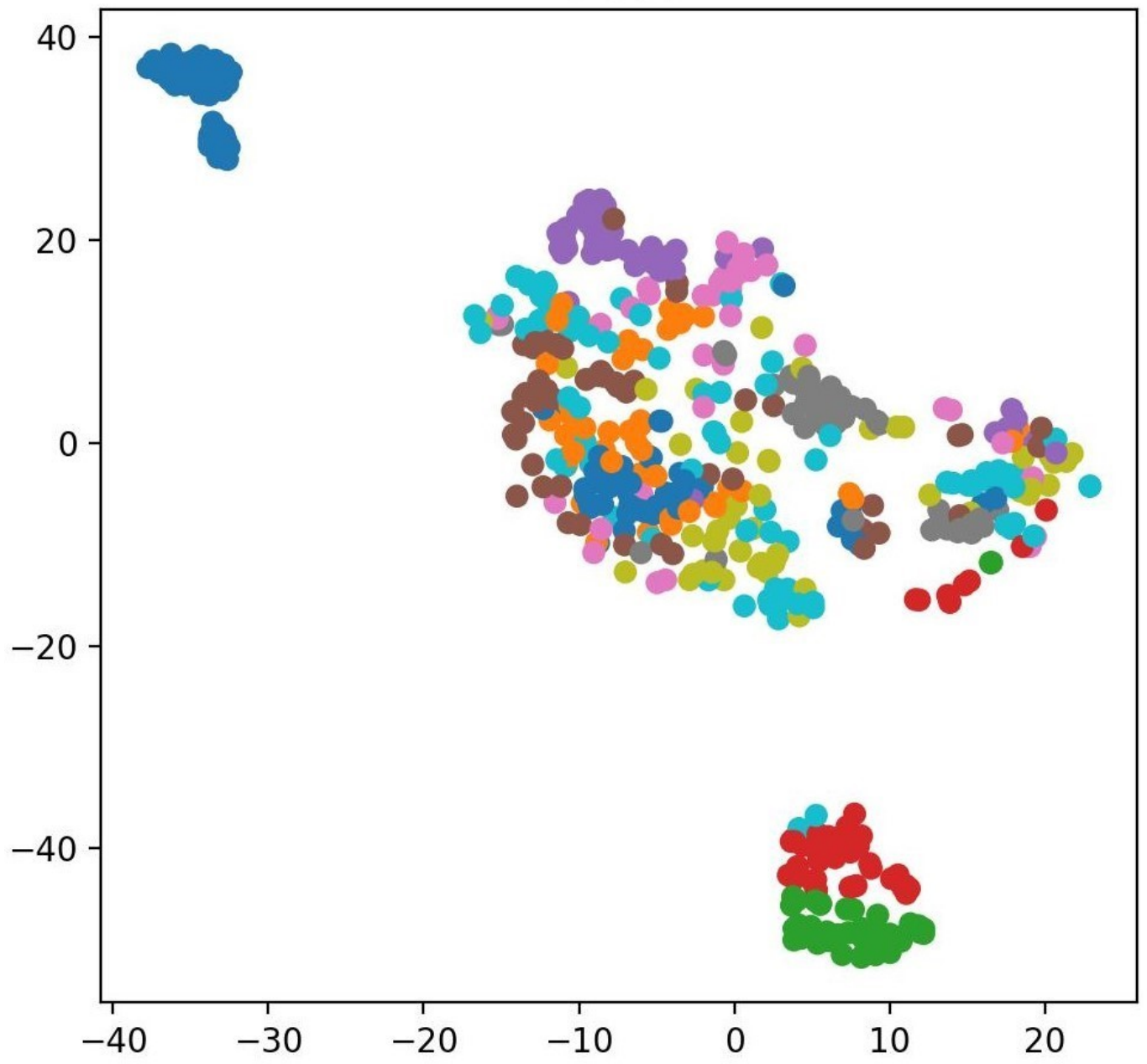}%
\label{Fig:pFL_visual_b}}
\subfloat[FedProx-0.005]{\includegraphics[width=0.3\linewidth,height=0.16\linewidth]{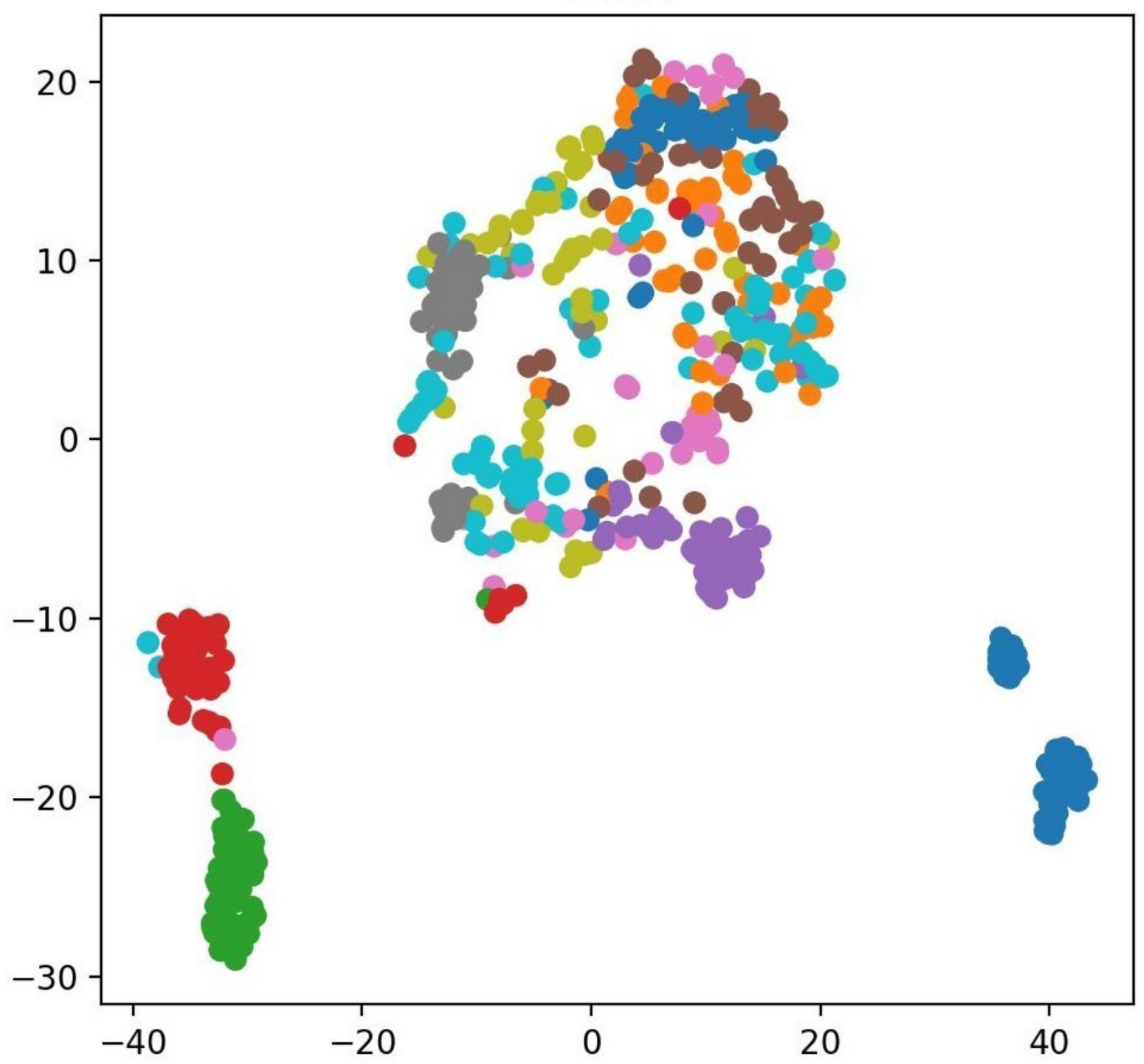}%
\label{Fig:pFL_visual_c}}

\subfloat[pFL-Vehicle-\#1]{\includegraphics[width=0.3\linewidth,height=0.16\linewidth]{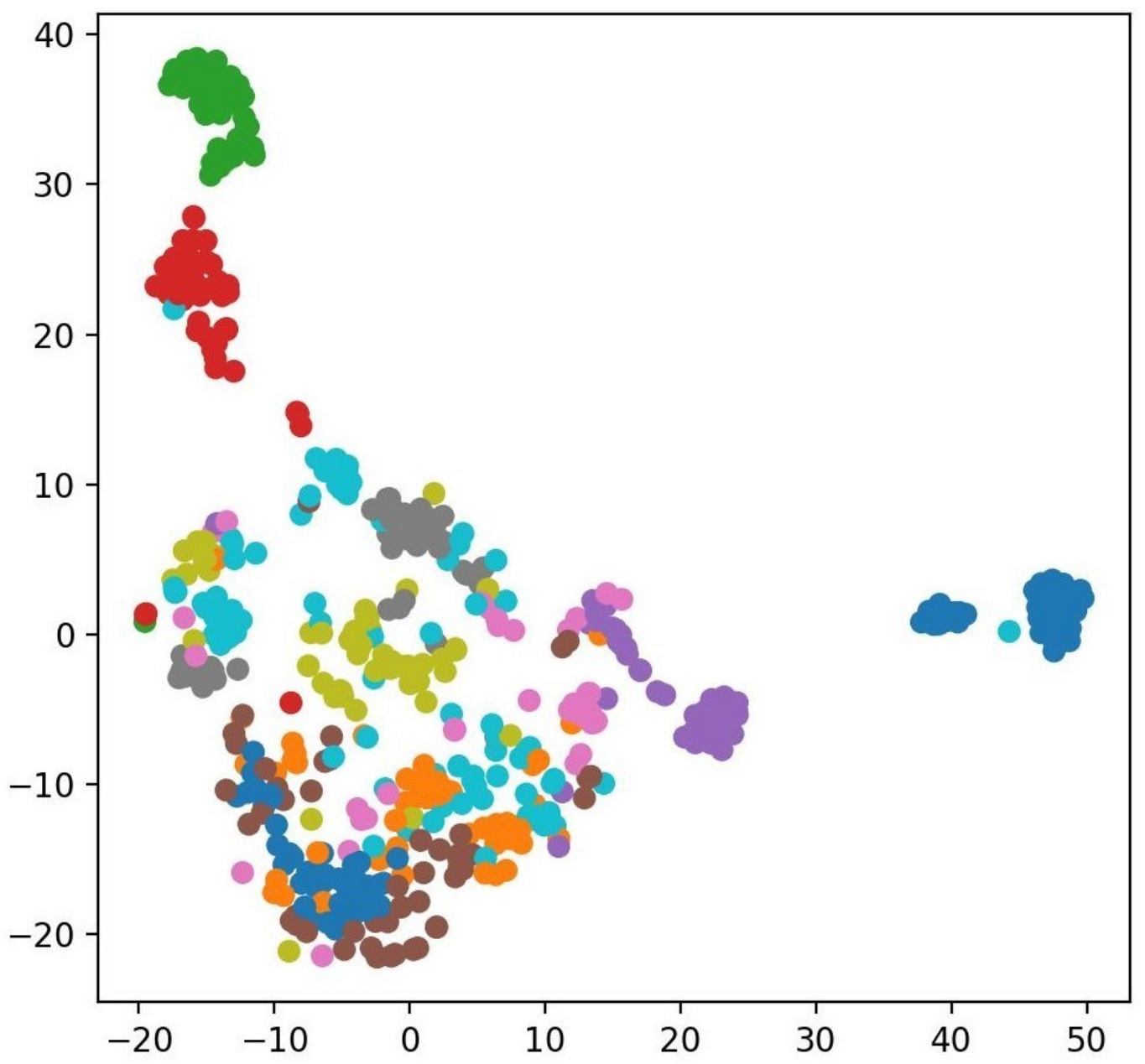}%
\label{Fig:pFL_visual_d}}
\subfloat[pFL-Vehicle-\#2]{\includegraphics[width=0.3\linewidth,height=0.16\linewidth]{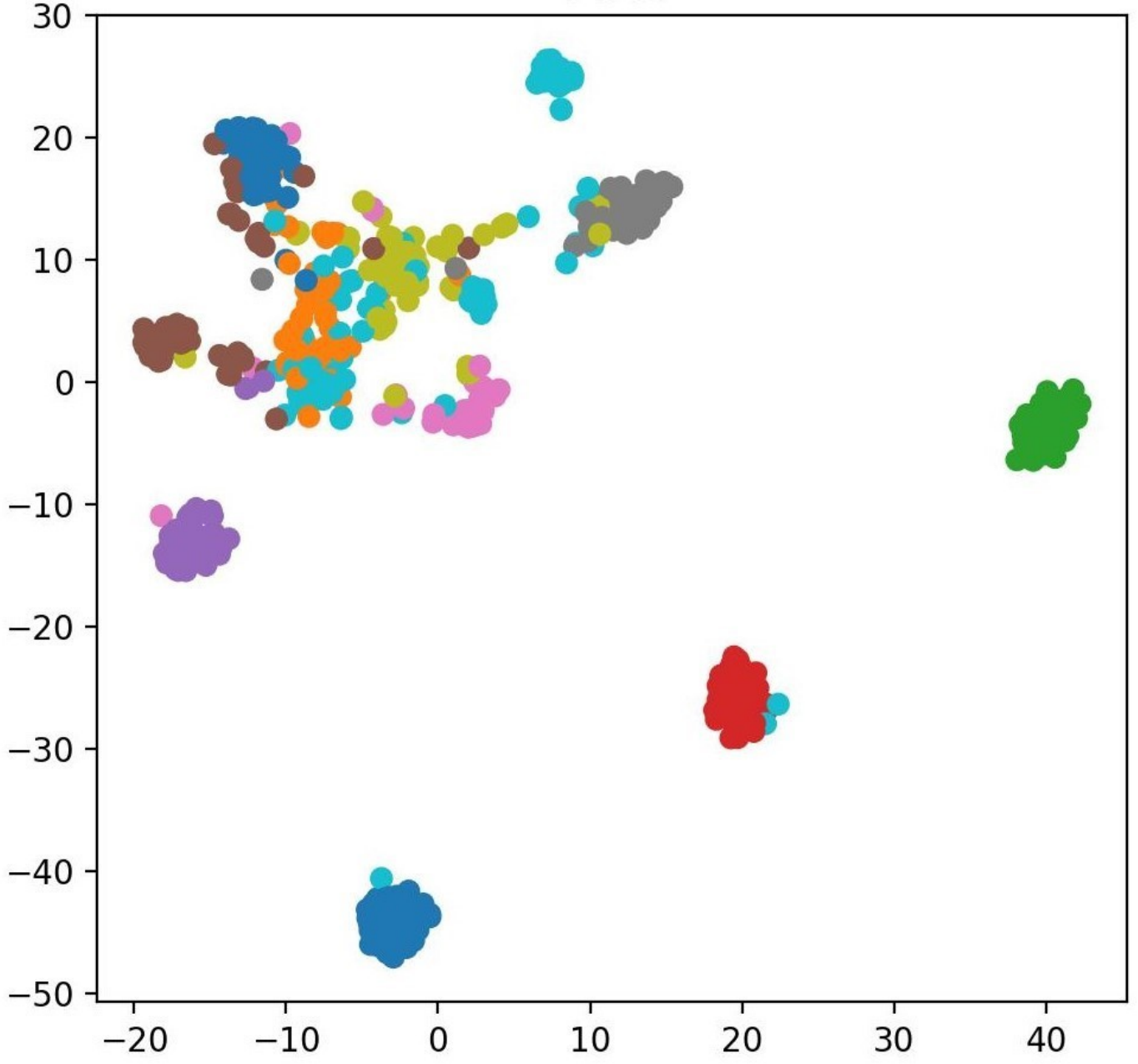}%
\label{Fig:pFL_visual_e}}
\subfloat[pFL-Vehicle-\#3]{\includegraphics[width=0.3\linewidth,height=0.16\linewidth]{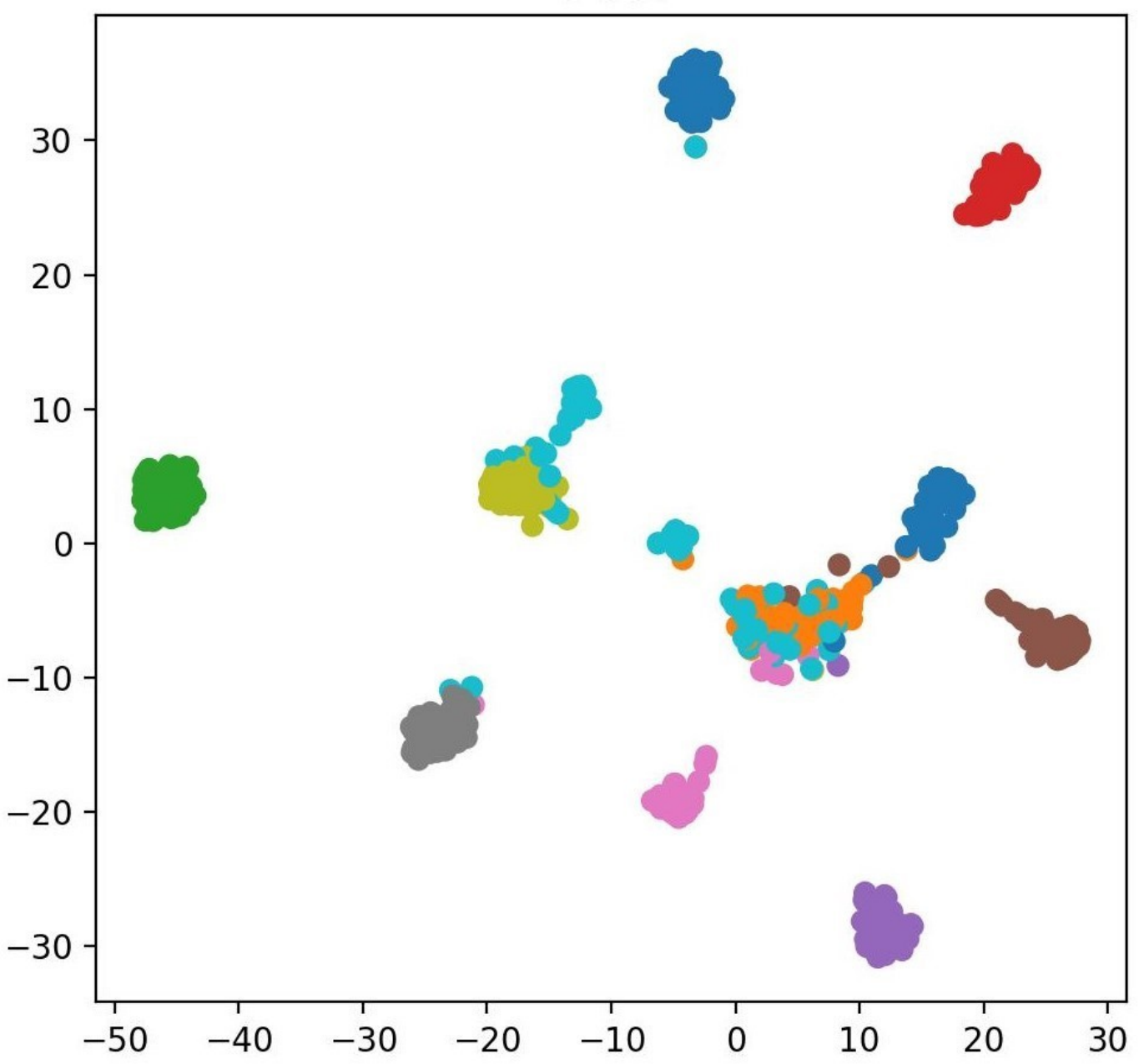}%
\label{Fig:pFL_visual_f}}
\caption{t-SNE visualization of pixel embedding of CamVid test dataset. Colors represent semantic classes.}
\label{Fig:pFL_visual}
\vspace{-0.5cm}
\end{figure*}

\Cref{Fig:pFL_visual} visualizes the pixel embeddings of the CamVid test dataset using t-SNE \cite{van2008visualizing,miao2023fedseg}. 
It can be seen that pFL-Vehicle\#2 and pFL-Vehicle\#3 models show the best separation of semantic classes, which is similar to that of Cityscapes dataset and aligns with the performance metrics presented in \Cref{Fig:feat_pFL_a,Fig:feat_pFL_b,Fig:feat_pFL_c,Fig:feat_pFL_d}.

\subsubsection{Evaluation of Communication Efficiency of the proposed pFedLVM algorithm}

\begin{figure*}[!t]
\vspace{-0.8cm}
\centering
\subfloat[Overhead Comparison on Cityscapes]{\includegraphics[width=0.35\linewidth]{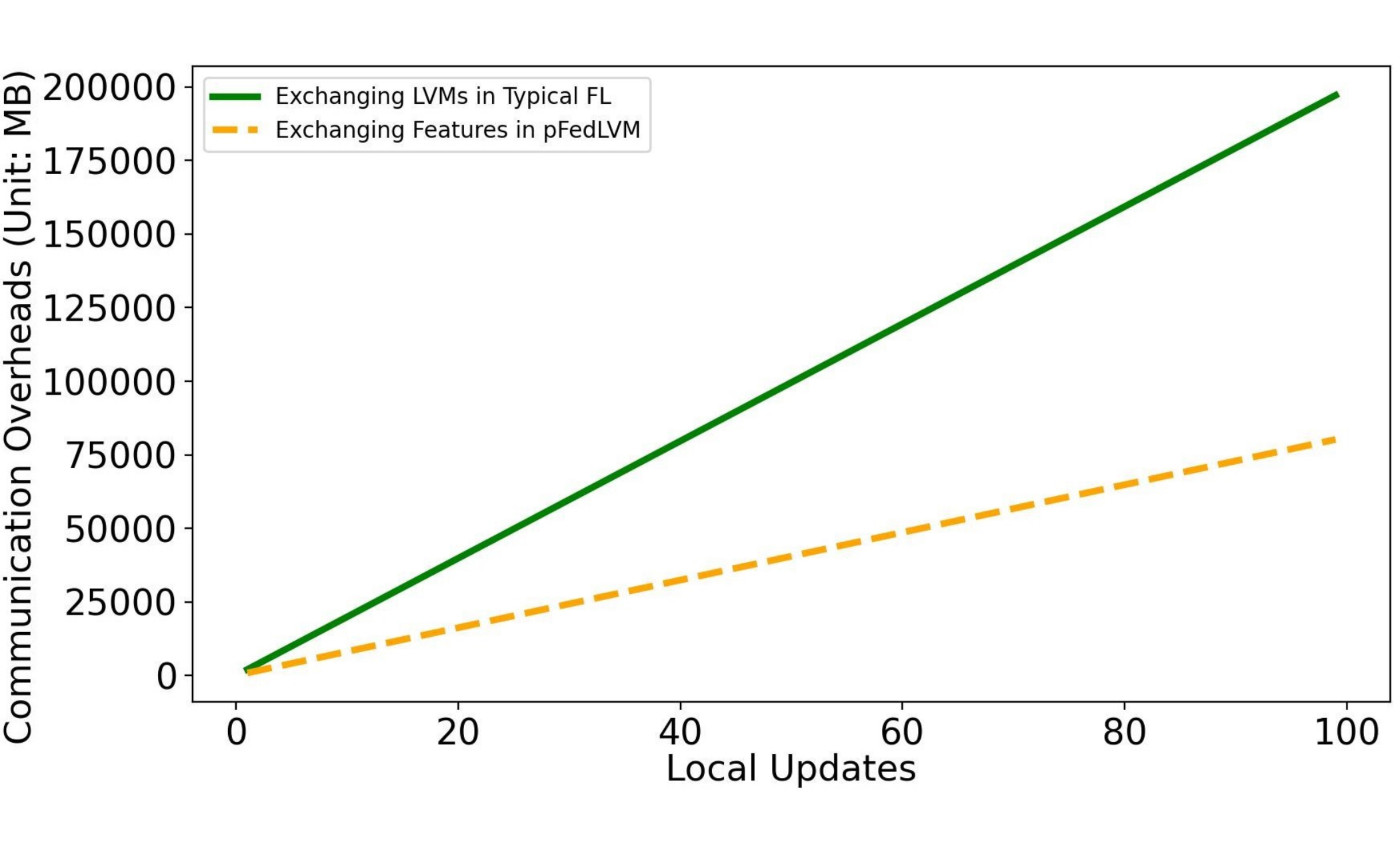}%
\label{Fig:overhead_comp_city}}
\hspace{0.5cm}
\subfloat[Overhead Comparison on CamVid]{\includegraphics[width=0.35\linewidth]{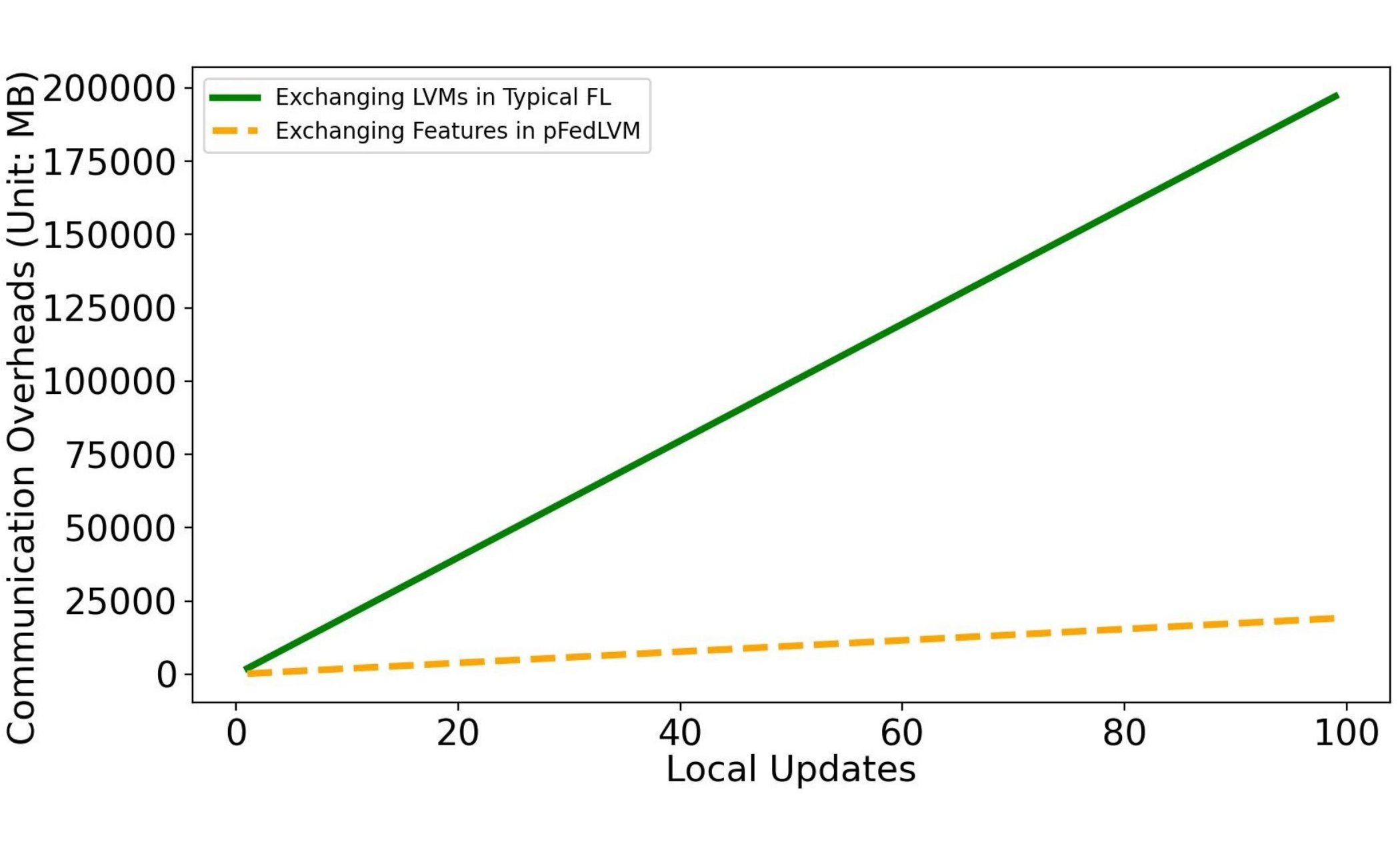}%
\label{Fig:overhead_comp_cam}}
\caption{Communication overheads of exchanging features in the proposed pFedLVM against exchanging LVMs in typical FL.}
\label{Fig:overheads_comparison}
\vspace{-0.5cm}
\end{figure*}

\Cref{Fig:overheads_comparison} shows the communication overheads of exchanging features in the proposed pFedLVM against the communication overheads of exchanging LVMs in typical FL. For both Cityscapes dataset (\Cref{Fig:overhead_comp_city}) and CamVid dataset (\Cref{Fig:overhead_comp_cam}), as the training progresses, the communication overheads of both exchanging schemes increase linearly yet with different growth rate, which suggests that the communication overhead saving by the proposed pFedLVM becomes more and more pronounced as the training progresses.

\begin{figure*}[!t]
\centering
\subfloat[Batch Size]{\includegraphics[width=0.28\linewidth, height=0.18\linewidth]{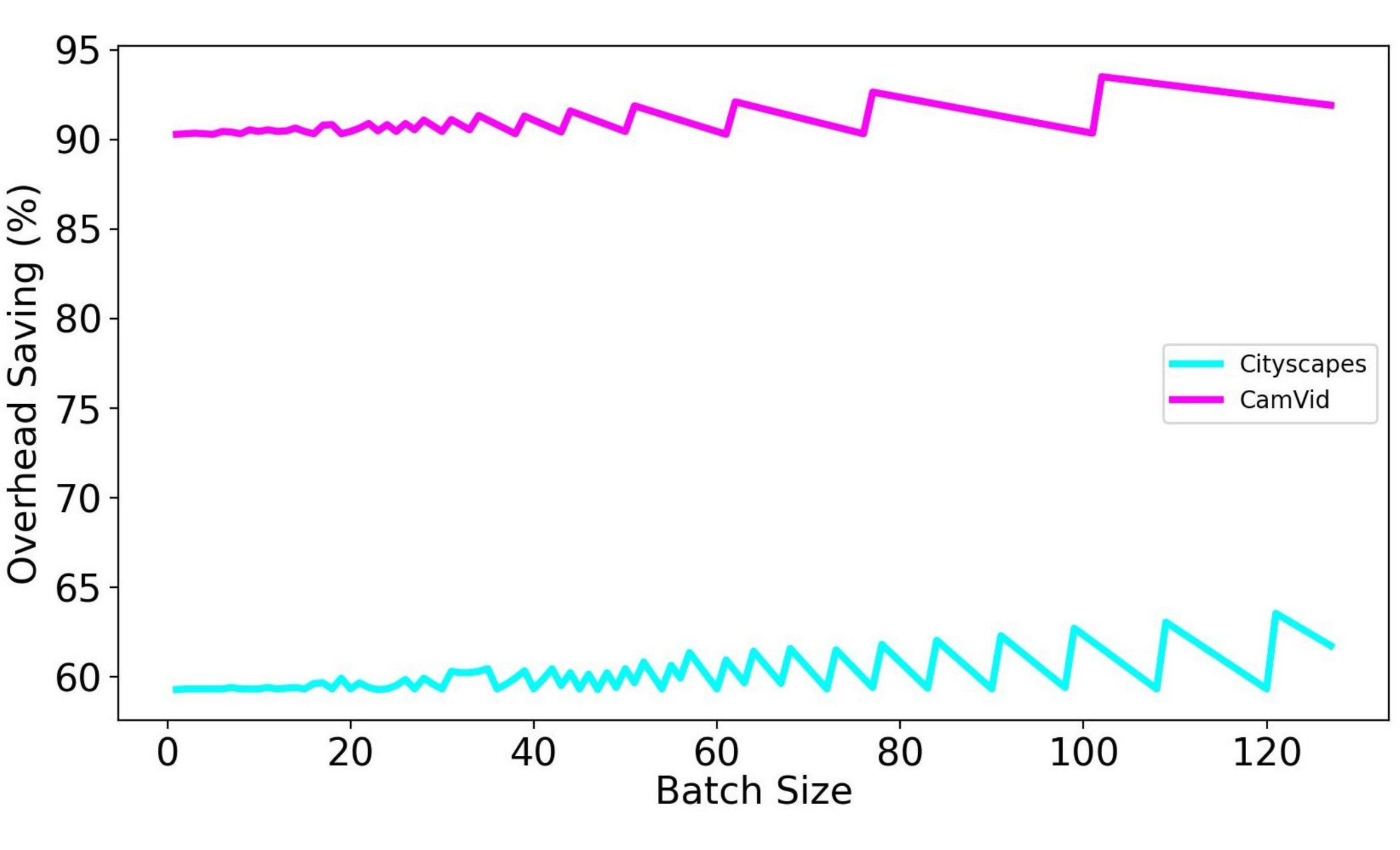}%
\label{Fig:reduction_wrt_bs}}
\hspace{0.3cm}
\subfloat[Feature Size]{\includegraphics[width=0.28\linewidth, height=0.18\linewidth]{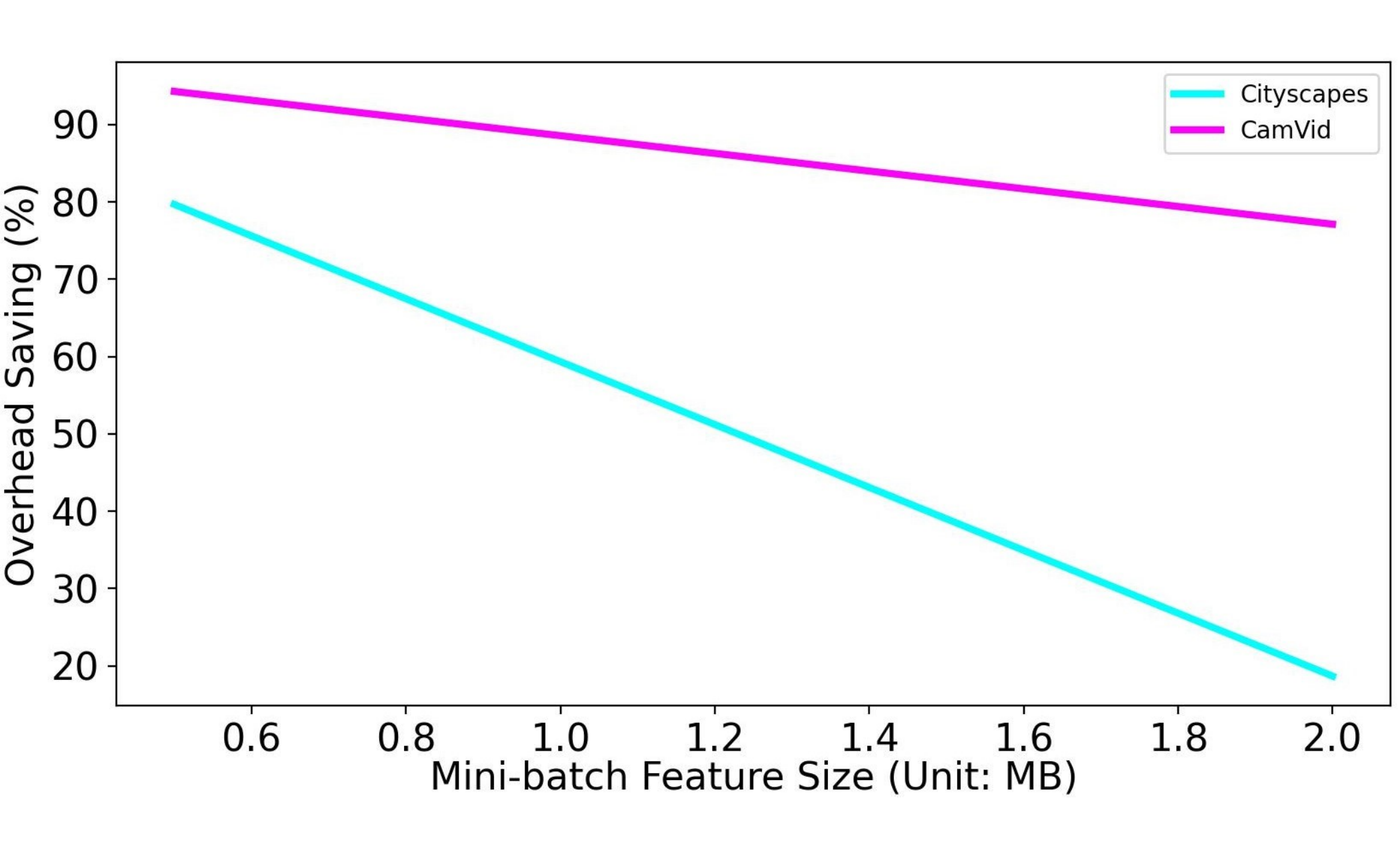}%
\label{Fig:reduction_wrt_fb}}
\hspace{0.3cm}
\subfloat[LVM Size]{\includegraphics[width=0.28\linewidth, height=0.18\linewidth]{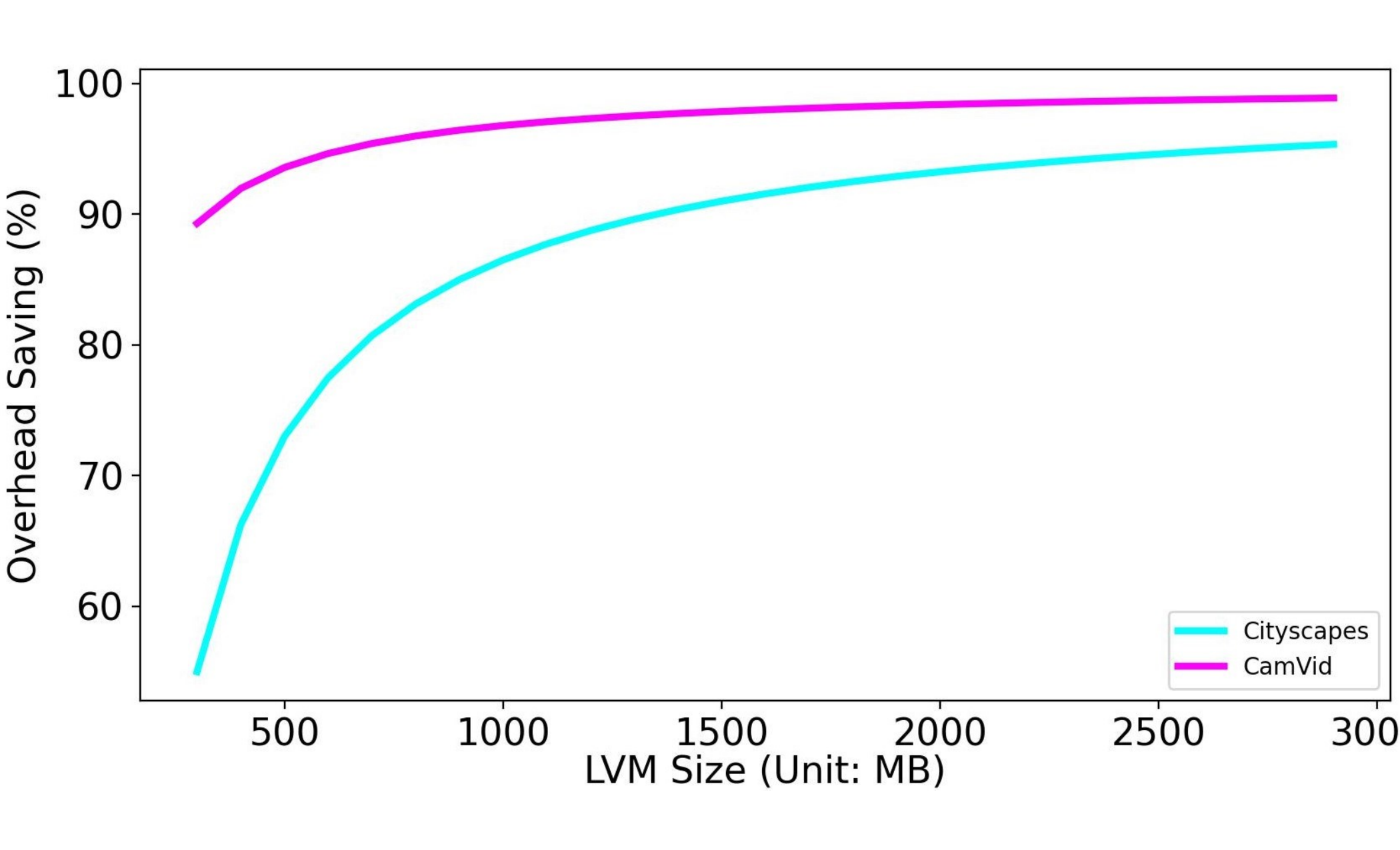}%
\label{Fig:reduction_wrt_mb}}
\caption{Illustration of how various factors contribute to the overhead saving on Cityscapes and CamVid datasets.}
\label{Fig:reductions_wrt_factors}
\vspace{-0.5cm}
\end{figure*}

To explore how mini-batch size, feature size and LVM size contribute to communication overhead saving achieved by the proposed pFedLVM against the typical FL, Fig. 13 shows the overhead saving versus these parameters. The observations and implications are summarized as follow: 
I) \Cref{Fig:reduction_wrt_bs} displays the result of how batch size contributes the communication overhead saving. We can observe that the communication overhead saving fluctuates in a small range as the batch size increases. As mini-batch size has little impact on improving the communication efficiency of pFedLVM, we can adjust it according to the performance need of the personalized models. II) \Cref{Fig:reduction_wrt_fb} tells us that the feature size has an important role in affecting communication overhead saving. For example, for the Cityscapes dataset, the communication reduction drops from approximately 60\% to 20\% when the feature size increases from 1.0MB to 2.0MB. As shrinking the feature size improves the communication efficiency substantially while it will decrease the personalized models’ performance, we should consider the trade-off carefully. III) LVM size also poses a significant effect on the communication overhead reduction, and this pattern can be viewed in \Cref{Fig:reduction_wrt_mb}. It is obvious that as the LVM size increases, the communication overhead reduction increases with a large margin as well. This suggests that the proposed framework is well suited for FL integrated with large models.

\section{Conclusion}
\label{conclusion}
This paper introduced pFedLVM framework, which integrates FL with LVM in autodriving context. The proposed framework deploys LVMs only on a central server to reduce computational burden of vehicles, and exchanges learned features instead of LVMs to reduce communication overheads. In addition, a personalized learning mechanism was incorporated, leading to superior performance than a global model trained using typical FL. The communication efficiency, space and time complexities of the proposed pFedLVM were also analyzed. Experimental results showed that pFedLVM outperforms currently existing SOTA approaches by large margins. Future work could incorporate multi-modal data, such as natural language, into pFedLVM. 

We acknowledge that the proposed pFedLVM framework has broad applicability across various domains. While we focus on AD due to its complex and highly heterogeneous data, the fundamental principles of pFedLVM can be adapted to other fields, such as finance, intelligent medicine, smart city, etc. This generalization highlights the potential widespread impact of the pFedLVM framework.


\begin{thebibliography}{10}
\bibitem{natan2022towards}
O.~Natan and J.~Miura, ``Towards compact autonomous driving perception with balanced learning and multi-sensor fusion,'' \emph{IEEE Transactions on Intelligent Transportation Systems}, vol.~23, no.~9, pp. 16\,249--16\,266, 2022.

\bibitem{kou2024fast}
W.-B. Kou, Q.~Lin, M.~Tang, R.~Ye, S.~Wang, G.~Zhu, and Y.-C. Wu, ``Fast-convergent and communication-alleviated heterogeneous hierarchical federated learning in autonomous driving,'' \emph{arXiv preprint arXiv:2409.19560}, 2024.

\bibitem{song2024robustness}
Z.~Song, L.~Liu, F.~Jia, Y.~Luo, C.~Jia, G.~Zhang, L.~Yang, and L.~Wang, ``Robustness-aware 3d object detection in autonomous driving: A review and outlook,'' \emph{IEEE Transactions on Intelligent Transportation Systems}, 2024.

\bibitem{karle2023multi}
P.~Karle, F.~Fent, S.~Huch, F.~Sauerbeck, and M.~Lienkamp, ``Multi-modal sensor fusion and object tracking for autonomous racing,'' \emph{IEEE Transactions on Intelligent Vehicles}, vol.~8, no.~7, pp. 3871--3883, 2023.

\bibitem{xiao2020multimodal}
Y.~Xiao, F.~Codevilla, A.~Gurram, O.~Urfalioglu, and A.~M. L{\'o}pez, ``Multimodal end-to-end autonomous driving,'' \emph{IEEE Transactions on Intelligent Transportation Systems}, vol.~23, no.~1, pp. 537--547, 2020.

\bibitem{10494721}
M.~Kang, S.~Wang, S.~Zhou, K.~Ye, J.~Jiang, and N.~Zheng, ``Ffinet: Future feedback interaction network for motion forecasting,'' \emph{IEEE Transactions on Intelligent Transportation Systems}, pp. 1--12, 2024.

\bibitem{10372140}
J.~Wu, J.~Ruenz, H.~Berkemeyer, L.~Dixon, and M.~Althoff, ``Goal-oriented pedestrian motion prediction,'' \emph{IEEE Transactions on Intelligent Transportation Systems}, vol.~25, no.~6, pp. 5282--5298, 2024.

\bibitem{yu2021bisenet}
C.~Yu, C.~Gao, J.~Wang, G.~Yu, C.~Shen, and N.~Sang, ``Bisenet v2: Bilateral network with guided aggregation for real-time semantic segmentation,'' \emph{International Journal of Computer Vision}, vol. 129, pp. 3051--3068, 2021.

\bibitem{10414408}
W.~Zheng, X.~Jiang, Z.~Fang, and Y.~Gao, ``Tv-net: A structure-level feature fusion network based on tensor voting for road crack segmentation,'' \emph{IEEE Transactions on Intelligent Transportation Systems}, vol.~25, no.~6, pp. 5743--5754, 2024.

\bibitem{hoyer2023domain}
L.~Hoyer, D.-X. Dai, and L.~Van~Gool, ``Domain adaptive and generalizable network architectures and training strategies for semantic image segmentation,'' \emph{IEEE Transactions on Pattern Analysis and Machine Intelligence}, 2023.

\bibitem{9917556}
C.~Yang, M.~Xu, Q.~Wang, Z.~Chen, K.~Huang, Y.~Ma, K.~Bian, G.~Huang, Y.~Liu, X.~Jin, and X.~Liu, ``Flash: Heterogeneity-aware federated learning at scale,'' \emph{IEEE Transactions on Mobile Computing}, vol.~23, no.~1, pp. 483--500, 2024.

\bibitem{mcmahan2017communication}
B.~McMahan, E.~Moore, D.~Ramage, S.~Hampson, and B.~A. y~Arcas, ``Communication-efficient learning of deep networks from decentralized data,'' in \emph{Artificial Intelligence and Statistics}.\hskip 1em plus 0.5em minus 0.4em\relax PMLR, 2017, pp. 1273--1282.

\bibitem{9272666}
G.~Zhu, Y.~Du, D.~Gündüz, and K.~Huang, ``One-bit over-the-air aggregation for communication-efficient federated edge learning: Design and convergence analysis,'' \emph{IEEE Transactions on Wireless Communications}, vol.~20, no.~3, pp. 2120--2135, 2021.

\bibitem{8970161}
G.~Zhu, D.~Liu, Y.~Du, C.~You, J.~Zhang, and K.~Huang, ``Toward an intelligent edge: Wireless communication meets machine learning,'' \emph{IEEE Communications Magazine}, vol.~58, no.~1, pp. 19--25, 2020.

\bibitem{zhu2023pushing}
G.~Zhu, Z.~Lyu, X.~Jiao, P.~Liu, M.~Chen, J.~Xu, S.~Cui, and P.~Zhang, ``Pushing ai to wireless network edge: An overview on integrated sensing, communication, and computation towards 6g,'' \emph{Science China Information Sciences}, vol.~66, no.~3, p. 130301, 2023.

\bibitem{kou2023communication}
W.-B. Kou, S.~Wang, G.~Zhu, B.~Luo, Y.~Chen, D.~W.~K. Ng, and Y.-C. Wu, ``Communication resources constrained hierarchical federated learning for end-to-end autonomous driving,'' in \emph{2023 IEEE/RSJ International Conference on Intelligent Robots and Systems (IROS)}.\hskip 1em plus 0.5em minus 0.4em\relax IEEE, 2023, pp. 9383--9390.

\bibitem{wu2024hierarchical}
H.-T. Wu, H.~Li, H.-L. Chi, W.-B. Kou, Y.-C. Wu, and S.~Wang, ``A hierarchical federated learning framework for collaborative quality defect inspection in construction,'' \emph{Engineering Applications of Artificial Intelligence}, vol. 133, p. 108218, 2024.

\bibitem{10346209}
Y.~Hui, J.~Hu, N.~Cheng, G.~Zhao, R.~Chen, T.~H. Luan, and K.~Aldubaikhy, ``Rcfl: Redundancy-aware collaborative federated learning in vehicular networks,'' \emph{IEEE Transactions on Intelligent Transportation Systems}, vol.~25, no.~6, pp. 5539--5553, 2024.

\bibitem{9831009}
B.~Li, Y.~Jiang, Q.~Pei, T.~Li, L.~Liu, and R.~Lu, ``Feel: Federated end-to-end learning with non-iid data for vehicular ad hoc networks,'' \emph{IEEE Transactions on Intelligent Transportation Systems}, vol.~23, no.~9, pp. 16\,728--16\,740, 2022.

\bibitem{nakkiran2021deep}
P.~Nakkiran, G.~Kaplun, Y.~Bansal, T.~Yang, B.~Barak, and I.~Sutskever, ``Deep double descent: Where bigger models and more data hurt,'' \emph{Journal of Statistical Mechanics: Theory and Experiment}, vol. 2021, no.~12, p. 124003, 2021.

\bibitem{vaswani2017attention}
A.~Vaswani, ``Attention is all you need,'' \emph{Advances in Neural Information Processing Systems}, 2017.

\bibitem{kaplan2020scaling}
J.~Kaplan, S.~McCandlish, T.~Henighan, T.~B. Brown, B.~Chess, R.~Child, S.~Gray, A.~Radford, J.~Wu, and D.~Amodei, ``Scaling laws for neural language models,'' \emph{arXiv preprint arXiv:2001.08361}, 2020.

\bibitem{Zhu2023mini}
D.~Zhu, J.~Chen, X.~Shen, X.~Li, and M.~Elhoseiny, ``Minigpt-4: Enhancing vision-language understanding with advanced large language models,'' \emph{arXiv preprint arXiv:2304.10592}, 2023.

\bibitem{ouyang2022training}
L.~Ouyang, J.~Wu, X.~Jiang, D.~Almeida, C.~Wainwright, P.~Mishkin, C.~Zhang, S.~Agarwal, K.~Slama, A.~Ray \emph{et~al.}, ``Training language models to follow instructions with human feedback,'' \emph{Advances in neural information processing systems}, vol.~35, pp. 27\,730--27\,744, 2022.

\bibitem{jiang2023structgpt}
J.~Jiang, K.~Zhou, Z.~Dong, K.~Ye, W.~X. Zhao, and J.-R. Wen, ``Structgpt: A general framework for large language model to reason over structured data,'' 2023.

\bibitem{wei2022chain}
J.~Wei, X.~Wang, D.~Schuurmans, M.~Bosma, F.~Xia, E.~Chi, Q.~V. Le, D.~Zhou \emph{et~al.}, ``Chain-of-thought prompting elicits reasoning in large language models,'' \emph{Advances in neural information processing systems}, vol.~35, pp. 24\,824--24\,837, 2022.

\bibitem{wang2023videomae}
L.~Wang, B.~Huang, Z.~Zhao, Z.~Tong, Y.~He, Y.~Wang, Y.~Wang, and Y.~Qiao, ``Videomae v2: Scaling video masked autoencoders with dual masking,'' in \emph{Proceedings of the IEEE/CVF Conference on Computer Vision and Pattern Recognition}, 2023, pp. 14\,549--14\,560.

\bibitem{ralethe2022generic}
S.~Ralethe and J.~Buys, ``Generic overgeneralization in pre-trained language models,'' 2022.

\bibitem{collacciani2023interpretation}
C.~Collacciani and G.~Rambelli, ``Interpretation of generalization in masked language models: An investigation straddling quantifiers and generics,'' 2023.

\bibitem{10388394}
Y.~Yang, Z.~Zhou, J.~Wu, Y.~Wang, and R.~Xiong, ``Class semantics modulation for open-set instance segmentation,'' \emph{IEEE Robotics and Automation Letters}, vol.~9, no.~3, pp. 2240--2247, 2024.

\bibitem{10161421}
Q.~Yan, S.~Li, C.~Liu, M.~Liu, and Q.~Chen, ``Fdlnet: Boosting real-time semantic segmentation by image-size convolution via frequency domain learning,'' in \emph{2023 IEEE International Conference on Robotics and Automation (ICRA)}, 2023.

\bibitem{yang2022deaot}
Z.~Yang and Y.~Yang, ``Decoupling features in hierarchical propagation for video object segmentation,'' in \emph{Advances in Neural Information Processing Systems (NeurIPS)}, 2022.

\bibitem{zhou2022rethinking}
T.~Zhou, W.~Wang, E.~Konukoglu, and L.~Van~Gool, ``Rethinking semantic segmentation: A prototype view,'' in \emph{Proceedings of the IEEE/CVF Conference on Computer Vision and Pattern Recognition}, 2022, pp. 2582--2593.

\bibitem{xie2021segformer}
E.~Xie, W.~Wang, Z.~Yu, A.~Anandkumar, J.~M. Alvarez, and P.~Luo, ``Segformer: Simple and efficient design for semantic segmentation with transformers,'' in \emph{Neural Information Processing Systems (NeurIPS)}, 2021.

\bibitem{9697426}
Y.~B. Can, A.~Liniger, O.~Unal, D.~Paudel, and L.~Van~Gool, ``Understanding bird’s-eye view of road semantics using an onboard camera,'' \emph{IEEE Robotics and Automation Letters}, vol.~7, no.~2, pp. 3302--3309, 2022.

\bibitem{10342134}
W.-B. Kou, S.~Wang, G.~Zhu, B.~Luo, Y.~Chen, D.~W. Kwan~Ng, and Y.-C. Wu, ``Communication resources constrained hierarchical federated learning for end-to-end autonomous driving,'' in \emph{2023 IEEE/RSJ International Conference on Intelligent Robots and Systems (IROS)}, 2023, pp. 9383--9390.

\bibitem{sun2016high}
Z.~Sun, Z.~Huang, Q.~Zhu, X.~Li, and D.~Liu, ``High-precision motion control method and practice for autonomous driving in complex off-road environments,'' in \emph{2016 IEEE Intelligent Vehicles Symposium (IV)}.\hskip 1em plus 0.5em minus 0.4em\relax IEEE, 2016, pp. 767--773.

\bibitem{jiang2019multi}
Y.~Jiang, H.~Yedidsion, S.~Zhang, G.~Sharon, and P.~Stone, ``Multi-robot planning with conflicts and synergies,'' \emph{Autonomous Robots}, vol.~43, pp. 2011--2032, 2019.

\bibitem{nguyen2022deep}
A.~Nguyen, T.~Do, M.~Tran, B.~X. Nguyen, C.~Duong, T.~Phan, E.~Tjiputra, and Q.~D. Tran, ``Deep federated learning for autonomous driving,'' in \emph{2022 IEEE Intelligent Vehicles Symposium (IV)}.\hskip 1em plus 0.5em minus 0.4em\relax IEEE, 2022, pp. 1824--1830.

\bibitem{9165167}
Y.~Xiao, F.~Codevilla, A.~Gurram, O.~Urfalioglu, and A.~M. López, ``Multimodal end-to-end autonomous driving,'' \emph{IEEE Transactions on Intelligent Transportation Systems}, vol.~23, no.~1, pp. 537--547, 2022.

\bibitem{10324362}
S.~Zhang, J.~Li, L.~Shi, M.~Ding, D.~C. Nguyen, W.~Tan, J.~Weng, and Z.~Han, ``Federated learning in intelligent transportation systems: Recent applications and open problems,'' \emph{IEEE Transactions on Intelligent Transportation Systems}, vol.~25, no.~5, pp. 3259--3285, 2024.

\bibitem{9244132}
H.~Liao, Z.~Zhou, W.~Kong, Y.~Chen, X.~Wang, Z.~Wang, and S.~Al~Otaibi, ``Learning-based intent-aware task offloading for air-ground integrated vehicular edge computing,'' \emph{IEEE Transactions on Intelligent Transportation Systems}, vol.~22, no.~8, pp. 5127--5139, 2021.

\bibitem{10033088}
Y.~Xiao, R.~Xia, Y.~Li, G.~Shi, D.~N. Nguyen, D.~T. Hoang, D.~Niyato, and M.~Krunz, ``Distributed traffic synthesis and classification in edge networks: A federated self-supervised learning approach,'' \emph{IEEE Transactions on Mobile Computing}, vol.~23, no.~2, pp. 1815--1829, 2024.

\bibitem{liu2020federated}
B.~Liu, L.~Wang, M.~Liu, and C.-Z. Xu, ``Federated imitation learning: A novel framework for cloud robotic systems with heterogeneous sensor data,'' \emph{IEEE Robotics and Automation Letters}, vol.~5, no.~2, pp. 3509--3516, 2020.

\bibitem{10101681}
S.~S. Shinde and D.~Tarchi, ``Joint air-ground distributed federated learning for intelligent transportation systems,'' \emph{IEEE Transactions on Intelligent Transportation Systems}, vol.~24, no.~9, pp. 9996--10\,011, 2023.

\bibitem{10279509}
Q.~Lin, Y.~Li, W.-B. Kou, T.-H. Chang, and Y.-C. Wu, ``Communication-efficient joint signal compression and activity detection in cell-free massive mimo,'' in \emph{ICC 2023 - IEEE International Conference on Communications}, 2023, pp. 5030--5035.

\bibitem{9505307}
S.~Liu, J.~Yu, X.~Deng, and S.~Wan, ``Fedcpf: An efficient-communication federated learning approach for vehicular edge computing in 6g communication networks,'' \emph{IEEE Transactions on Intelligent Transportation Systems}, vol.~23, no.~2, pp. 1616--1629, 2022.

\bibitem{10529194}
Q.~Lin, Y.~Li, W.-B. Kou, T.-H. Chang, and Y.-C. Wu, ``Communication-efficient activity detection for cell-free massive mimo: An augmented model-driven end-to-end learning framework,'' \emph{IEEE Transactions on Wireless Communications}, pp. 1--1, 2024.

\bibitem{pillutla2022federated}
K.~Pillutla, K.~Malik, A.-R. Mohamed, M.~Rabbat, M.~Sanjabi, and L.~Xiao, ``Federated learning with partial model personalization,'' in \emph{International Conference on Machine Learning}.\hskip 1em plus 0.5em minus 0.4em\relax PMLR, 2022, pp. 17\,716--17\,758.

\bibitem{tan2022towards}
A.~Z. Tan, H.~Yu, L.~Cui, and Q.~Yang, ``Towards personalized federated learning,'' \emph{IEEE Transactions on Neural Networks and Learning Systems}, 2022, {DOI}:~10.1109/TNNLS.2022.3160699.

\bibitem{zhang2021parameterized}
J.~Zhang, S.~Guo, X.~Ma, H.~Wang, W.~Xu, and F.~Wu, ``Parameterized knowledge transfer for personalized federated learning,'' \emph{Advances in Neural Information Processing Systems}, vol.~34, pp. 10\,092--10\,104, 2021.

\bibitem{huang2021personalized}
Y.~Huang, L.~Chu, Z.~Zhou, L.~Wang, J.~Liu, J.~Pei, and Y.~Zhang, ``Personalized cross-silo federated learning on non-iid data,'' in \emph{Proceedings of the AAAI Conference on Artificial Intelligence}, vol.~35, no.~9, 2021, pp. 7865--7873, {DOI}:~10.1609/aaai.v35i9.16960.

\bibitem{bui2019federated}
D.~Bui \emph{et~al.}, ``Federated user representation learning,'' \emph{arXiv preprint arXiv:1909.12535}, 2019.

\bibitem{NEURIPS2020_24389bfe}
A.~Fallah, A.~Mokhtari, and A.~Ozdaglar, ``{Personalized Federated Learning with Theoretical Guarantees: A Model-Agnostic Meta-Learning Approach},'' in \emph{NeurIPS}, 2020.

\bibitem{collins2021exploiting}
L.~Collins, H.~Hassani, A.~Mokhtari, and S.~Shakkottai, ``Exploiting shared representations for personalized federated learning,'' in \emph{International Conference on Machine Learning}.\hskip 1em plus 0.5em minus 0.4em\relax PMLR, Jul. 2021, pp. 2089--2099.

\bibitem{fu2022complexity}
Y.~Fu, H.~Peng, A.~Sabharwal, P.~Clark, and T.~Khot, ``Complexity-based prompting for multi-step reasoning,'' in \emph{The Eleventh International Conference on Learning Representations}, 2022.

\bibitem{ren2023rejuvenating}
S.~Ren, Z.~Wang, H.~Zhu, J.~Xiao, A.~Yuille, and C.~Xie, ``Rejuvenating image-gpt as strong visual representation learners,'' 2023.

\bibitem{wang2023all}
J.~Wang, Y.~Ge, R.~Yan, Y.~Ge, K.~Q. Lin, S.~Tsutsui, X.~Lin, G.~Cai, J.~Wu, Y.~Shan \emph{et~al.}, ``All in one: Exploring unified video-language pre-training,'' in \emph{Proceedings of the IEEE/CVF Conference on Computer Vision and Pattern Recognition}, 2023, pp. 6598--6608.

\bibitem{chen2020generative}
M.~Chen, A.~Radford, R.~Child, J.~Wu, H.~Jun, D.~Luan, and I.~Sutskever, ``Generative pretraining from pixels,'' in \emph{International conference on machine learning}.\hskip 1em plus 0.5em minus 0.4em\relax PMLR, 2020, pp. 1691--1703.

\bibitem{chen2017deeplab}
L.-C. Chen, G.~Papandreou, I.~Kokkinos, K.~Murphy, and A.~L. Yuille, ``Deeplab: Semantic image segmentation with deep convolutional nets, atrous convolution, and fully connected crfs,'' \emph{IEEE transactions on pattern analysis and machine intelligence}, vol.~40, no.~4, pp. 834--848, 2017.

\bibitem{li2020federated}
T.~Li, S.~Hu, A.~Beirami, and V.~Smith, ``Federated multi-task learning for competing constraints,'' \emph{arXiv preprint arXiv:2012.04221}, 2020.

\bibitem{acar2021federated}
D.~A.~E. Acar, Y.~Zhao, R.~Matas, M.~Mattina, P.~Whatmough, and V.~Saligrama, ``Federated learning based on dynamic regularization,'' in \emph{International Conference on Learning Representations}, 2021.

\bibitem{Cordts2016Cityscapes}
M.~Cordts, M.~Omran, S.~Ramos, T.~Rehfeld, M.~Enzweiler, R.~Benenson, U.~Franke, S.~Roth, and B.~Schiele, ``The cityscapes dataset for semantic urban scene understanding,'' in \emph{Proc. of the IEEE Conference on Computer Vision and Pattern Recognition (CVPR)}, 2016.

\bibitem{brostow2008segmentation}
G.~J. Brostow, J.~Shotton, J.~Fauqueur, and R.~Cipolla, ``Segmentation and recognition using structure from motion point clouds,'' in \emph{Computer Vision--ECCV 2008: 10th European Conference on Computer Vision, Marseille, France, October 12-18, 2008, Proceedings, Part I 10}.\hskip 1em plus 0.5em minus 0.4em\relax Springer, 2008, pp. 44--57.

\bibitem{badrinarayanan2017segnet}
V.~Badrinarayanan, A.~Kendall, and R.~Cipolla, ``Segnet: A deep convolutional encoder-decoder architecture for image segmentation,'' \emph{IEEE transactions on pattern analysis and machine intelligence}, vol.~39, no.~12, pp. 2481--2495, 2017.

\bibitem{van2008visualizing}
L.~Van~der Maaten and G.~Hinton, ``Visualizing data using t-sne.'' \emph{Journal of machine learning research}, vol.~9, no.~11, 2008.

\bibitem{miao2023fedseg}
J.~Miao, Z.~Yang, L.~Fan, and Y.~Yang, ``Fedseg: Class-heterogeneous federated learning for semantic segmentation,'' in \emph{Proceedings of the IEEE/CVF Conference on Computer Vision and Pattern Recognition}, 2023, pp. 8042--8052.

\end{thebibliography}
\end{document}